\documentstyle[avm, rep12, lingmacros, buthesis]{report}
\include{psfig}
\newcounter{treecount}
\newcounter{branchcount}
\newsavebox{\parentbox}
\newsavebox{\treebox}
\newsavebox{\treeboxone}
\newsavebox{\treeboxtwo}
\newsavebox{\treeboxthree}
\newsavebox{\treeboxfour}
\newsavebox{\treeboxfive}
\newsavebox{\treeboxsix}
\newsavebox{\treeboxseven}
\newsavebox{\treeboxeight}
\newsavebox{\treeboxnine}
\newsavebox{\treeboxten}
\newsavebox{\treeboxeleven}
\newsavebox{\treeboxtwelve}
\newsavebox{\treeboxthirteen}
\newsavebox{\treeboxfourteen}
\newsavebox{\treeboxfifteen}
\newsavebox{\treeboxsixteen}
\newsavebox{\treeboxseventeen}
\newsavebox{\treeboxeighteen}
\newsavebox{\treeboxnineteen}
\newsavebox{\treeboxtwenty}
\newlength{\treeoffsetone}
\newlength{\treeoffsettwo}
\newlength{\treeoffsetthree}
\newlength{\treeoffsetfour}
\newlength{\treeoffsetfive}
\newlength{\treeoffsetsix}
\newlength{\treeoffsetseven}
\newlength{\treeoffseteight}
\newlength{\treeoffsetnine}
\newlength{\treeoffsetten}
\newlength{\treeoffseteleven}
\newlength{\treeoffsettwelve}
\newlength{\treeoffsetthirteen}
\newlength{\treeoffsetfourteen}
\newlength{\treeoffsetfifteen}
\newlength{\treeoffsetsixteen}
\newlength{\treeoffsetseventeen}
\newlength{\treeoffseteighteen}
\newlength{\treeoffsetnineteen}
\newlength{\treeoffsettwenty}

\newlength{\treeshiftone}
\newlength{\treeshifttwo}
\newlength{\treeshiftthree}
\newlength{\treeshiftfour}
\newlength{\treeshiftfive}
\newlength{\treeshiftsix}
\newlength{\treeshiftseven}
\newlength{\treeshifteight}
\newlength{\treeshiftnine}
\newlength{\treeshiftten}
\newlength{\treeshifteleven}
\newlength{\treeshifttwelve}
\newlength{\treeshiftthirteen}
\newlength{\treeshiftfourteen}
\newlength{\treeshiftfifteen}
\newlength{\treeshiftsixteen}
\newlength{\treeshiftseventeen}
\newlength{\treeshifteighteen}
\newlength{\treeshiftnineteen}
\newlength{\treeshifttwenty}
\newlength{\treewidthone}
\newlength{\treewidthtwo}
\newlength{\treewidththree}
\newlength{\treewidthfour}
\newlength{\treewidthfive}
\newlength{\treewidthsix}
\newlength{\treewidthseven}
\newlength{\treewidtheight}
\newlength{\treewidthnine}
\newlength{\treewidthten}
\newlength{\treewidtheleven}
\newlength{\treewidthtwelve}
\newlength{\treewidththirteen}
\newlength{\treewidthfourteen}
\newlength{\treewidthfifteen}
\newlength{\treewidthsixteen}
\newlength{\treewidthseventeen}
\newlength{\treewidtheighteen}
\newlength{\treewidthnineteen}
\newlength{\treewidthtwenty}
\newlength{\daughteroffsetone}
\newlength{\daughteroffsettwo}
\newlength{\daughteroffsetthree}
\newlength{\daughteroffsetfour}
\newlength{\branchwidthone}
\newlength{\branchwidthtwo}
\newlength{\branchwidththree}
\newlength{\branchwidthfour}
\newlength{\parentoffset}
\newlength{\treeoffset}
\newlength{\daughteroffset}
\newlength{\branchwidth}
\newlength{\parentwidth}
\newlength{\treewidth}
\newcommand{\ontop}[1]{\begin{tabular}{c}#1\end{tabular}}
\newcommand{\poptree}{%
\ifnum\value{treecount}=0\typeout{QobiTeX warning---Tree stack underflow}\fi%
\addtocounter{treecount}{-1}%
\setlength{\treeoffsettwo}{\treeoffsetthree}%
\setlength{\treeoffsetthree}{\treeoffsetfour}%
\setlength{\treeoffsetfour}{\treeoffsetfive}%
\setlength{\treeoffsetfive}{\treeoffsetsix}%
\setlength{\treeoffsetsix}{\treeoffsetseven}%
\setlength{\treeoffsetseven}{\treeoffseteight}%
\setlength{\treeoffseteight}{\treeoffsetnine}%
\setlength{\treeoffsetnine}{\treeoffsetten}%
\setlength{\treeoffsetten}{\treeoffseteleven}%
\setlength{\treeoffseteleven}{\treeoffsettwelve}%
\setlength{\treeoffsettwelve}{\treeoffsetthirteen}%
\setlength{\treeoffsetthirteen}{\treeoffsetfourteen}%
\setlength{\treeoffsetfourteen}{\treeoffsetfifteen}%
\setlength{\treeoffsetfifteen}{\treeoffsetsixteen}%
\setlength{\treeoffsetsixteen}{\treeoffsetseventeen}%
\setlength{\treeoffsetseventeen}{\treeoffseteighteen}%
\setlength{\treeoffseteighteen}{\treeoffsetnineteen}%
\setlength{\treeoffsetnineteen}{\treeoffsettwenty}%
\setlength{\treeshifttwo}{\treeshiftthree}%
\setlength{\treeshiftthree}{\treeshiftfour}%
\setlength{\treeshiftfour}{\treeshiftfive}%
\setlength{\treeshiftfive}{\treeshiftsix}%
\setlength{\treeshiftsix}{\treeshiftseven}%
\setlength{\treeshiftseven}{\treeshifteight}%
\setlength{\treeshifteight}{\treeshiftnine}%
\setlength{\treeshiftnine}{\treeshiftten}%
\setlength{\treeshiftten}{\treeshifteleven}%
\setlength{\treeshifteleven}{\treeshifttwelve}%
\setlength{\treeshifttwelve}{\treeshiftthirteen}%
\setlength{\treeshiftthirteen}{\treeshiftfourteen}%
\setlength{\treeshiftfourteen}{\treeshiftfifteen}%
\setlength{\treeshiftfifteen}{\treeshiftsixteen}%
\setlength{\treeshiftsixteen}{\treeshiftseventeen}%
\setlength{\treeshiftseventeen}{\treeshifteighteen}%
\setlength{\treeshifteighteen}{\treeshiftnineteen}%
\setlength{\treeshiftnineteen}{\treeshifttwenty}%
\setlength{\treewidthtwo}{\treewidththree}%
\setlength{\treewidththree}{\treewidthfour}%
\setlength{\treewidthfour}{\treewidthfive}%
\setlength{\treewidthfive}{\treewidthsix}%
\setlength{\treewidthsix}{\treewidthseven}%
\setlength{\treewidthseven}{\treewidtheight}%
\setlength{\treewidtheight}{\treewidthnine}%
\setlength{\treewidthnine}{\treewidthten}%
\setlength{\treewidthten}{\treewidtheleven}%
\setlength{\treewidtheleven}{\treewidthtwelve}%
\setlength{\treewidthtwelve}{\treewidththirteen}%
\setlength{\treewidththirteen}{\treewidthfourteen}%
\setlength{\treewidthfourteen}{\treewidthfifteen}%
\setlength{\treewidthfifteen}{\treewidthsixteen}%
\setlength{\treewidthsixteen}{\treewidthseventeen}%
\setlength{\treewidthseventeen}{\treewidtheighteen}%
\setlength{\treewidtheighteen}{\treewidthnineteen}%
\setlength{\treewidthnineteen}{\treewidthtwenty}%
\sbox{\treeboxtwo}{\usebox{\treeboxthree}}%
\sbox{\treeboxthree}{\usebox{\treeboxfour}}%
\sbox{\treeboxfour}{\usebox{\treeboxfive}}%
\sbox{\treeboxfive}{\usebox{\treeboxsix}}%
\sbox{\treeboxsix}{\usebox{\treeboxseven}}%
\sbox{\treeboxseven}{\usebox{\treeboxeight}}%
\sbox{\treeboxeight}{\usebox{\treeboxnine}}%
\sbox{\treeboxnine}{\usebox{\treeboxten}}%
\sbox{\treeboxten}{\usebox{\treeboxeleven}}%
\sbox{\treeboxeleven}{\usebox{\treeboxtwelve}}%
\sbox{\treeboxtwelve}{\usebox{\treeboxthirteen}}%
\sbox{\treeboxthirteen}{\usebox{\treeboxfourteen}}%
\sbox{\treeboxfourteen}{\usebox{\treeboxfifteen}}%
\sbox{\treeboxfifteen}{\usebox{\treeboxsixteen}}%
\sbox{\treeboxsixteen}{\usebox{\treeboxseventeen}}%
\sbox{\treeboxseventeen}{\usebox{\treeboxeighteen}}%
\sbox{\treeboxeighteen}{\usebox{\treeboxnineteen}}%
\sbox{\treeboxnineteen}{\usebox{\treeboxtwenty}}}
\newcommand{\leaf}[1]{%
\ifnum\value{treecount}=20\typeout{QobiTeX warning---Tree stack overflow}\fi%
\addtocounter{treecount}{1}%
\sbox{\treeboxtwenty}{\usebox{\treeboxnineteen}}%
\sbox{\treeboxnineteen}{\usebox{\treeboxeighteen}}%
\sbox{\treeboxeighteen}{\usebox{\treeboxseventeen}}%
\sbox{\treeboxseventeen}{\usebox{\treeboxsixteen}}%
\sbox{\treeboxsixteen}{\usebox{\treeboxfifteen}}%
\sbox{\treeboxfifteen}{\usebox{\treeboxfourteen}}%
\sbox{\treeboxfourteen}{\usebox{\treeboxthirteen}}%
\sbox{\treeboxthirteen}{\usebox{\treeboxtwelve}}%
\sbox{\treeboxtwelve}{\usebox{\treeboxeleven}}%
\sbox{\treeboxeleven}{\usebox{\treeboxten}}%
\sbox{\treeboxten}{\usebox{\treeboxnine}}%
\sbox{\treeboxnine}{\usebox{\treeboxeight}}%
\sbox{\treeboxeight}{\usebox{\treeboxseven}}%
\sbox{\treeboxseven}{\usebox{\treeboxsix}}%
\sbox{\treeboxsix}{\usebox{\treeboxfive}}%
\sbox{\treeboxfive}{\usebox{\treeboxfour}}%
\sbox{\treeboxfour}{\usebox{\treeboxthree}}%
\sbox{\treeboxthree}{\usebox{\treeboxtwo}}%
\sbox{\treeboxtwo}{\usebox{\treeboxone}}%
\sbox{\treeboxone}{\ontop{#1}}%
\sbox{\treeboxone}{\raisebox{-\ht\treeboxone}{\usebox{\treeboxone}}}%
\setlength{\treeoffsettwenty}{\treeoffsetnineteen}%
\setlength{\treeoffsetnineteen}{\treeoffseteighteen}%
\setlength{\treeoffseteighteen}{\treeoffsetseventeen}%
\setlength{\treeoffsetseventeen}{\treeoffsetsixteen}%
\setlength{\treeoffsetsixteen}{\treeoffsetfifteen}%
\setlength{\treeoffsetfifteen}{\treeoffsetfourteen}%
\setlength{\treeoffsetfourteen}{\treeoffsetthirteen}%
\setlength{\treeoffsetthirteen}{\treeoffsettwelve}%
\setlength{\treeoffsettwelve}{\treeoffseteleven}%
\setlength{\treeoffseteleven}{\treeoffsetten}%
\setlength{\treeoffsetten}{\treeoffsetnine}%
\setlength{\treeoffsetnine}{\treeoffseteight}%
\setlength{\treeoffseteight}{\treeoffsetseven}%
\setlength{\treeoffsetseven}{\treeoffsetsix}%
\setlength{\treeoffsetsix}{\treeoffsetfive}%
\setlength{\treeoffsetfive}{\treeoffsetfour}%
\setlength{\treeoffsetfour}{\treeoffsetthree}%
\setlength{\treeoffsetthree}{\treeoffsettwo}%
\setlength{\treeoffsettwo}{\treeoffsetone}%
\setlength{\treeoffsetone}{0.5\wd\treeboxone}%
\setlength{\treeshifttwenty}{\treeshiftnineteen}%
\setlength{\treeshiftnineteen}{\treeshifteighteen}%
\setlength{\treeshifteighteen}{\treeshiftseventeen}%
\setlength{\treeshiftseventeen}{\treeshiftsixteen}%
\setlength{\treeshiftsixteen}{\treeshiftfifteen}%
\setlength{\treeshiftfifteen}{\treeshiftfourteen}%
\setlength{\treeshiftfourteen}{\treeshiftthirteen}%
\setlength{\treeshiftthirteen}{\treeshifttwelve}%
\setlength{\treeshifttwelve}{\treeshifteleven}%
\setlength{\treeshifteleven}{\treeshiftten}%
\setlength{\treeshiftten}{\treeshiftnine}%
\setlength{\treeshiftnine}{\treeshifteight}%
\setlength{\treeshifteight}{\treeshiftseven}%
\setlength{\treeshiftseven}{\treeshiftsix}%
\setlength{\treeshiftsix}{\treeshiftfive}%
\setlength{\treeshiftfive}{\treeshiftfour}%
\setlength{\treeshiftfour}{\treeshiftthree}%
\setlength{\treeshiftthree}{\treeshifttwo}%
\setlength{\treeshifttwo}{\treeshiftone}%
\setlength{\treeshiftone}{0pt}%
\setlength{\treewidthtwenty}{\treewidthnineteen}%
\setlength{\treewidthnineteen}{\treewidtheighteen}%
\setlength{\treewidtheighteen}{\treewidthseventeen}%
\setlength{\treewidthseventeen}{\treewidthsixteen}%
\setlength{\treewidthsixteen}{\treewidthfifteen}%
\setlength{\treewidthfifteen}{\treewidthfourteen}%
\setlength{\treewidthfourteen}{\treewidththirteen}%
\setlength{\treewidththirteen}{\treewidthtwelve}%
\setlength{\treewidthtwelve}{\treewidtheleven}%
\setlength{\treewidtheleven}{\treewidthten}%
\setlength{\treewidthten}{\treewidthnine}%
\setlength{\treewidthnine}{\treewidtheight}%
\setlength{\treewidtheight}{\treewidthseven}%
\setlength{\treewidthseven}{\treewidthsix}%
\setlength{\treewidthsix}{\treewidthfive}%
\setlength{\treewidthfive}{\treewidthfour}%
\setlength{\treewidthfour}{\treewidththree}%
\setlength{\treewidththree}{\treewidthtwo}%
\setlength{\treewidthtwo}{\treewidthone}%
\setlength{\treewidthone}{\wd\treeboxone}}
\newcommand{\branch}[2]{%
\setcounter{branchcount}{#1}%
\ifnum\value{branchcount}=1\sbox{\parentbox}{\ontop{#2}}%
\setlength{\parentoffset}{\treeoffsetone}%
\addtolength{\parentoffset}{-0.5\wd\parentbox}%
\setlength{\daughteroffset}{0in}%
\ifdim\parentoffset<0in%
\setlength{\daughteroffset}{-\parentoffset}%
\setlength{\parentoffset}{0in}\fi%
\setlength{\parentwidth}{\parentoffset}%
\addtolength{\parentwidth}{\wd\parentbox}%
\setlength{\treeoffset}{\daughteroffset}%
\addtolength{\treeoffset}{\treeoffsetone}%
\setlength{\treewidth}{\wd\treeboxone}%
\addtolength{\treewidth}{\daughteroffset}%
\ifdim\treewidth<\parentwidth\setlength{\treewidth}{\parentwidth}\fi%
\sbox{\treebox}{\begin{minipage}{\treewidth}%
\begin{flushleft}%
\hspace*{\parentoffset}\usebox{\parentbox}\\
{\setlength{\unitlength}{2ex}%
\hspace*{\treeoffset}\begin{picture}(0,1)%
\put(0,0){\line(0,1){1}}%
\end{picture}}\\
\vspace{-\baselineskip}
\hspace*{\daughteroffset}%
\raisebox{-\ht\treeboxone}{\usebox{\treeboxone}}%
\end{flushleft}%
\end{minipage}}%
\setlength{\treeoffsetone}{\parentoffset}%
\addtolength{\treeoffsetone}{0.5\wd\parentbox}%
\setlength{\treeshiftone}{0pt}%
\setlength{\treewidthone}{\treewidth}%
\sbox{\treeboxone}{\usebox{\treebox}}%
\else\ifnum\value{branchcount}=2\sbox{\parentbox}{\ontop{#2}}%
\setlength{\branchwidthone}{\treewidthtwo}%
\addtolength{\branchwidthone}{\treeoffsetone}%
\addtolength{\branchwidthone}{-\treeshiftone}%
\addtolength{\branchwidthone}{-\treeoffsettwo}%
\setlength{\branchwidth}{\branchwidthone}%
\setlength{\daughteroffsetone}{\branchwidth}%
\addtolength{\daughteroffsetone}{-\branchwidthone}%
\addtolength{\daughteroffsetone}{-\treeshiftone}%
\setlength{\parentoffset}{-0.5\wd\parentbox}%
\addtolength{\parentoffset}{\treeoffsettwo}%
\addtolength{\parentoffset}{0.5\branchwidth}%
\setlength{\daughteroffset}{0in}%
\ifdim\parentoffset<0in%
\setlength{\daughteroffset}{-\parentoffset}%
\setlength{\parentoffset}{0in}\fi%
\setlength{\parentwidth}{\parentoffset}%
\addtolength{\parentwidth}{\wd\parentbox}%
\setlength{\treeoffset}{\daughteroffset}%
\addtolength{\treeoffset}{\treeoffsettwo}%
\setlength{\treewidth}{\wd\treeboxone}%
\addtolength{\treewidth}{\daughteroffsetone}%
\addtolength{\treewidth}{\treewidthtwo}%
\addtolength{\treewidth}{\daughteroffset}%
\ifdim\treewidth<\parentwidth\setlength{\treewidth}{\parentwidth}\fi%
\sbox{\treebox}{\begin{minipage}{\treewidth}%
\begin{flushleft}%
\hspace*{\parentoffset}\usebox{\parentbox}\\
{\setlength{\unitlength}{0.5\branchwidth}%
\hspace*{\treeoffset}\begin{picture}(2,0.5)%
\put(0,0){\line(2,1){1}}%
\put(2,0){\line(-2,1){1}}%
\end{picture}}\\
\vspace{-\baselineskip}
\hspace*{\daughteroffset}%
\makebox[\treewidthtwo][l]%
{\raisebox{-\ht\treeboxtwo}{\usebox{\treeboxtwo}}}%
\hspace*{\daughteroffsetone}%
\raisebox{-\ht\treeboxone}{\usebox{\treeboxone}}%
\end{flushleft}%
\end{minipage}}%
\setlength{\treeoffsetone}{\parentoffset}%
\addtolength{\treeoffsetone}{0.5\wd\parentbox}%
\setlength{\treeshiftone}{0pt}%
\setlength{\treewidthone}{\treewidth}%
\sbox{\treeboxone}{\usebox{\treebox}}\poptree%
\else\ifnum\value{branchcount}=3\sbox{\parentbox}{\ontop{#2}}%
\setlength{\branchwidthone}{\treewidthtwo}%
\addtolength{\branchwidthone}{\treeoffsetone}%
\addtolength{\branchwidthone}{-\treeshiftone}%
\addtolength{\branchwidthone}{-\treeoffsettwo}%
\setlength{\branchwidthtwo}{\treewidththree}%
\addtolength{\branchwidthtwo}{\treeoffsettwo}%
\addtolength{\branchwidthtwo}{-\treeshifttwo}%
\addtolength{\branchwidthtwo}{-\treeoffsetthree}%
\setlength{\branchwidth}{\branchwidthone}%
\ifdim\branchwidthtwo>\branchwidth%
\setlength{\branchwidth}{\branchwidthtwo}\fi%
\setlength{\daughteroffsetone}{\branchwidth}%
\addtolength{\daughteroffsetone}{-\branchwidthone}%
\addtolength{\daughteroffsetone}{-\treeshiftone}%
\setlength{\daughteroffsettwo}{\branchwidth}%
\addtolength{\daughteroffsettwo}{-\branchwidthtwo}%
\addtolength{\daughteroffsettwo}{-\treeshifttwo}%
\setlength{\parentoffset}{-0.5\wd\parentbox}%
\addtolength{\parentoffset}{\treeoffsetthree}%
\addtolength{\parentoffset}{\branchwidth}%
\setlength{\daughteroffset}{0in}%
\ifdim\parentoffset<0in%
\setlength{\daughteroffset}{-\parentoffset}%
\setlength{\parentoffset}{0in}\fi%
\setlength{\parentwidth}{\parentoffset}%
\addtolength{\parentwidth}{\wd\parentbox}%
\setlength{\treeoffset}{\daughteroffset}%
\addtolength{\treeoffset}{\treeoffsetthree}%
\setlength{\treewidth}{\wd\treeboxone}%
\addtolength{\treewidth}{\daughteroffsetone}%
\addtolength{\treewidth}{\treewidthtwo}%
\addtolength{\treewidth}{\daughteroffsettwo}%
\addtolength{\treewidth}{\treewidththree}%
\addtolength{\treewidth}{\daughteroffset}%
\ifdim\treewidth<\parentwidth\setlength{\treewidth}{\parentwidth}\fi%
\sbox{\treebox}{\begin{minipage}{\treewidth}%
\begin{flushleft}%
\hspace*{\parentoffset}\usebox{\parentbox}\\
{\setlength{\unitlength}{0.5\branchwidth}%
\hspace*{\treeoffset}\begin{picture}(4,1)%
\put(0,0){\line(2,1){2}}%
\put(2,0){\line(0,1){1}}%
\put(4,0){\line(-2,1){2}}%
\end{picture}}\\
\vspace{-\baselineskip}
\hspace*{\daughteroffset}%
\makebox[\treewidththree][l]%
{\raisebox{-\ht\treeboxthree}{\usebox{\treeboxthree}}}%
\hspace*{\daughteroffsettwo}%
\makebox[\treewidthtwo][l]%
{\raisebox{-\ht\treeboxtwo}{\usebox{\treeboxtwo}}}%
\hspace*{\daughteroffsetone}%
\raisebox{-\ht\treeboxone}{\usebox{\treeboxone}}%
\end{flushleft}%
\end{minipage}}%
\setlength{\treeoffsetone}{\parentoffset}%
\addtolength{\treeoffsetone}{0.5\wd\parentbox}%
\setlength{\treeshiftone}{0pt}%
\setlength{\treewidthone}{\treewidth}%
\sbox{\treeboxone}{\usebox{\treebox}}\poptree\poptree%
\else\ifnum\value{branchcount}=4\sbox{\parentbox}{\ontop{#2}}%
\setlength{\branchwidthone}{\treewidthtwo}%
\addtolength{\branchwidthone}{\treeoffsetone}%
\addtolength{\branchwidthone}{-\treeshiftone}%
\addtolength{\branchwidthone}{-\treeoffsettwo}%
\setlength{\branchwidthtwo}{\treewidththree}%
\addtolength{\branchwidthtwo}{\treeoffsettwo}%
\addtolength{\branchwidthtwo}{-\treeshifttwo}%
\addtolength{\branchwidthtwo}{-\treeoffsetthree}%
\setlength{\branchwidththree}{\treewidthfour}%
\addtolength{\branchwidththree}{\treeoffsetthree}%
\addtolength{\branchwidththree}{-\treeshiftthree}%
\addtolength{\branchwidththree}{-\treeoffsetfour}%
\setlength{\branchwidth}{\branchwidthone}%
\ifdim\branchwidthtwo>\branchwidth%
\setlength{\branchwidth}{\branchwidthtwo}\fi%
\ifdim\branchwidththree>\branchwidth%
\setlength{\branchwidth}{\branchwidththree}\fi%
\setlength{\daughteroffsetone}{\branchwidth}%
\addtolength{\daughteroffsetone}{-\branchwidthone}%
\addtolength{\daughteroffsetone}{-\treeshiftone}%
\setlength{\daughteroffsettwo}{\branchwidth}%
\addtolength{\daughteroffsettwo}{-\branchwidthtwo}%
\addtolength{\daughteroffsettwo}{-\treeshifttwo}%
\setlength{\daughteroffsetthree}{\branchwidth}%
\addtolength{\daughteroffsetthree}{-\branchwidththree}%
\addtolength{\daughteroffsetthree}{-\treeshiftthree}%
\setlength{\parentoffset}{-0.5\wd\parentbox}%
\addtolength{\parentoffset}{\treeoffsetfour}%
\addtolength{\parentoffset}{1.5\branchwidth}%
\setlength{\daughteroffset}{0in}%
\ifdim\parentoffset<0in%
\setlength{\daughteroffset}{-\parentoffset}%
\setlength{\parentoffset}{0in}\fi%
\setlength{\parentwidth}{\parentoffset}%
\addtolength{\parentwidth}{\wd\parentbox}%
\setlength{\treeoffset}{\daughteroffset}%
\addtolength{\treeoffset}{\treeoffsetfour}%
\setlength{\treewidth}{\wd\treeboxone}%
\addtolength{\treewidth}{\daughteroffsetone}%
\addtolength{\treewidth}{\treewidthtwo}%
\addtolength{\treewidth}{\daughteroffsettwo}%
\addtolength{\treewidth}{\treewidththree}%
\addtolength{\treewidth}{\daughteroffsetthree}%
\addtolength{\treewidth}{\treewidthfour}%
\addtolength{\treewidth}{\daughteroffset}%
\ifdim\treewidth<\parentwidth\setlength{\treewidth}{\parentwidth}\fi%
\sbox{\treebox}{\begin{minipage}{\treewidth}%
\begin{flushleft}%
\hspace*{\parentoffset}\usebox{\parentbox}\\
{\setlength{\unitlength}{0.5\branchwidth}%
\hspace*{\treeoffset}\begin{picture}(6,1)%
\put(0,0){\line(3,1){3}}%
\put(2,0){\line(1,1){1}}%
\put(4,0){\line(-1,1){1}}%
\put(6,0){\line(-3,1){3}}%
\end{picture}}\\
\vspace{-\baselineskip}
\hspace*{\daughteroffset}%
\makebox[\treewidthfour][l]%
{\raisebox{-\ht\treeboxfour}{\usebox{\treeboxfour}}}%
\hspace*{\daughteroffsetthree}%
\makebox[\treewidththree][l]%
{\raisebox{-\ht\treeboxthree}{\usebox{\treeboxthree}}}%
\hspace*{\daughteroffsettwo}%
\makebox[\treewidthtwo][l]%
{\raisebox{-\ht\treeboxtwo}{\usebox{\treeboxtwo}}}%
\hspace*{\daughteroffsetone}%
\raisebox{-\ht\treeboxone}{\usebox{\treeboxone}}%
\end{flushleft}%
\end{minipage}}%
\setlength{\treeoffsetone}{\parentoffset}%
\addtolength{\treeoffsetone}{0.5\wd\parentbox}%
\setlength{\treeshiftone}{0pt}%
\setlength{\treewidthone}{\treewidth}%
\sbox{\treeboxone}{\usebox{\treebox}}\poptree\poptree\poptree%
\else\ifnum\value{branchcount}=5\sbox{\parentbox}{\ontop{#2}}%
\setlength{\branchwidthone}{\treewidthtwo}%
\addtolength{\branchwidthone}{\treeoffsetone}%
\addtolength{\branchwidthone}{-\treeshiftone}%
\addtolength{\branchwidthone}{-\treeoffsettwo}%
\setlength{\branchwidthtwo}{\treewidththree}%
\addtolength{\branchwidthtwo}{\treeoffsettwo}%
\addtolength{\branchwidthtwo}{-\treeshifttwo}%
\addtolength{\branchwidthtwo}{-\treeoffsetthree}%
\setlength{\branchwidththree}{\treewidthfour}%
\addtolength{\branchwidththree}{\treeoffsetthree}%
\addtolength{\branchwidththree}{-\treeshiftthree}%
\addtolength{\branchwidththree}{-\treeoffsetfour}%
\setlength{\branchwidthfour}{\treewidthfive}%
\addtolength{\branchwidthfour}{\treeoffsetfour}%
\addtolength{\branchwidthfour}{-\treeshiftfour}%
\addtolength{\branchwidthfour}{-\treeoffsetfive}%
\setlength{\branchwidth}{\branchwidthone}%
\ifdim\branchwidthtwo>\branchwidth%
\setlength{\branchwidth}{\branchwidthtwo}\fi%
\ifdim\branchwidththree>\branchwidth%
\setlength{\branchwidth}{\branchwidththree}\fi%
\ifdim\branchwidthfour>\branchwidth%
\setlength{\branchwidth}{\branchwidthfour}\fi%
\setlength{\daughteroffsetone}{\branchwidth}%
\addtolength{\daughteroffsetone}{-\branchwidthone}%
\addtolength{\daughteroffsetone}{-\treeshiftone}%
\setlength{\daughteroffsettwo}{\branchwidth}%
\addtolength{\daughteroffsettwo}{-\branchwidthtwo}%
\addtolength{\daughteroffsettwo}{-\treeshifttwo}%
\setlength{\daughteroffsetthree}{\branchwidth}%
\addtolength{\daughteroffsetthree}{-\branchwidththree}%
\addtolength{\daughteroffsetthree}{-\treeshiftthree}%
\setlength{\daughteroffsetfour}{\branchwidth}%
\addtolength{\daughteroffsetfour}{-\branchwidthfour}%
\addtolength{\daughteroffsetfour}{-\treeshiftfour}%
\setlength{\parentoffset}{-0.5\wd\parentbox}%
\addtolength{\parentoffset}{\treeoffsetfive}%
\addtolength{\parentoffset}{2\branchwidth}%
\setlength{\daughteroffset}{0in}%
\ifdim\parentoffset<0in%
\setlength{\daughteroffset}{-\parentoffset}%
\setlength{\parentoffset}{0in}\fi%
\setlength{\parentwidth}{\parentoffset}%
\addtolength{\parentwidth}{\wd\parentbox}%
\setlength{\treeoffset}{\daughteroffset}%
\addtolength{\treeoffset}{\treeoffsetfive}%
\setlength{\treewidth}{\wd\treeboxone}%
\addtolength{\treewidth}{\daughteroffsetone}%
\addtolength{\treewidth}{\treewidthtwo}%
\addtolength{\treewidth}{\daughteroffsettwo}%
\addtolength{\treewidth}{\treewidththree}%
\addtolength{\treewidth}{\daughteroffsetthree}%
\addtolength{\treewidth}{\treewidthfour}%
\addtolength{\treewidth}{\daughteroffsetfour}%
\addtolength{\treewidth}{\treewidthfive}%
\addtolength{\treewidth}{\daughteroffset}%
\ifdim\treewidth<\parentwidth\setlength{\treewidth}{\parentwidth}\fi%
\sbox{\treebox}{\begin{minipage}{\treewidth}%
\begin{flushleft}%
\hspace*{\parentoffset}\usebox{\parentbox}\\
{\setlength{\unitlength}{0.5\branchwidth}%
\hspace*{\treeoffset}\begin{picture}(8,1)%
\put(0,0){\line(4,1){4}}%
\put(2,0){\line(2,1){2}}%
\put(4,0){\line(0,1){1}}%
\put(6,0){\line(-2,1){2}}%
\put(8,0){\line(-4,1){4}}%
\end{picture}}\\
\vspace{-\baselineskip}
\hspace*{\daughteroffset}%
\makebox[\treewidthfive][l]%
{\raisebox{-\ht\treeboxfour}{\usebox{\treeboxfive}}}%
\hspace*{\daughteroffsetfour}%
\makebox[\treewidthfour][l]%
{\raisebox{-\ht\treeboxfour}{\usebox{\treeboxfour}}}%
\hspace*{\daughteroffsetthree}%
\makebox[\treewidththree][l]%
{\raisebox{-\ht\treeboxthree}{\usebox{\treeboxthree}}}%
\hspace*{\daughteroffsettwo}%
\makebox[\treewidthtwo][l]%
{\raisebox{-\ht\treeboxtwo}{\usebox{\treeboxtwo}}}%
\hspace*{\daughteroffsetone}%
\raisebox{-\ht\treeboxone}{\usebox{\treeboxone}}%
\end{flushleft}%
\end{minipage}}%
\setlength{\treeoffsetone}{\parentoffset}%
\addtolength{\treeoffsetone}{0.5\wd\parentbox}%
\setlength{\treeshiftone}{0pt}%
\setlength{\treewidthone}{\treewidth}%
\sbox{\treeboxone}{\usebox{\treebox}}\poptree\poptree\poptree\poptree%
\else\typeout{QobiTeX warning--- Can't handle #1 branching}\fi\fi\fi\fi\fi}
\newcommand{\tree}{%
\usebox{\treeboxone}
\setlength{\treeoffsetone}{\treeoffsettwo}%
\sbox{\treeboxone}{\usebox{\treeboxtwo}}%
\poptree}

\avmfont{\sc}
\avmvalfont{\rm}
\avmsortfont{\em}
\avmoptions{sorted}

\newcommand{\avmBegin}{\begin{center} \begin{footnotesize} \begin{avm}}
\newcommand{\avmEnd}{\end{avm} \end{footnotesize} \end{center}}
\newcommand{\fnsCBegin}{\begin{footnotesize} \begin{center}}
\newcommand{\fnsCEnd}{\end{center} \end{footnotesize}}
\newcommand{\vspGloss}{\vspace{0.15cm}}
\newcommand{\tabSpc}{\qquad\qquad}

\title{
  DESIGN AND IMPLEMENTATION\\
               OF\\
    A COMPUTATIONAL LEXICON\\
          FOR  TURKISH\\}

\author{Abdullah Kurtulu\c{s} Yorulmaz}
\dept{Computer Engineering and Information Science}
\bolum{Bilgisayar ve Enformatik M\"{u}hendisli\u{g}i B\"{o}l\"{u}m\"{u}}
\principaladvisor{Asst. Prof. Kemal Oflazer}
\tezyoneticisi{Yrd. Do\c{c}. Dr. Kemal Oflazer}
\firstreader{Assoc. Prof. Halil Altay G\"{u}venir}
\secondreader{Asst. Prof. \.{I}lyas \c{C}i\c{c}ekli}
\copyrightyear{February, 1997}
\submitdate{February, 1997}
\tarih{\c{S}ubat, 1997}

\begin{document}
\pagenumbering{roman}
\titlep
\signaturepage

\newpage
\setcounter{page}{3}

\begin{center}
{\Large \bf ABSTRACT}

\vspace{0.5cm}

  DESIGN AND IMPLEMENTATION\\
               OF\\
    A COMPUTATIONAL LEXICON\\
          FOR  TURKISH\\

\vspace{0.5cm}
Abdullah Kurtulu\c{s} Yorulmaz \\
M.S. in Computer Engineering and Information Science\\
Supervisor: Asst. Prof. Kemal Oflazer\\
February, 1997\\

\end{center}

All natural language processing systems (such as parsers, generators,
taggers) need to have access to a lexicon about the words in the
language. This thesis presents a lexicon architecture for natural
language processing in Turkish. Given a query form consisting of a
surface form and other features acting as restrictions, the lexicon
produces feature structures containing morphosyntactic, syntactic, and
semantic information for all possible interpretations of the surface
form satisfying those restrictions. The lexicon is based on
contemporary approaches like feature-based representation,
inheritance, and unification. It makes use of two information
sources: a morphological processor and a lexical database containing
all the open and closed-class words of Turkish. The system has been
implemented in {\em SICStus Prolog} as a standalone module for use in
natural language processing applications.
   
\vfill{\em Key words\/}: Natural Language Processing, Lexicon

\newpage
\setcounter{page}{4}

\begin{center}
{\Large \bf \"{O}ZET}

\vspace{0.5cm}

T\"{U}RK\c{C}E \.{I}\c{C}\.{I}N\\ 
B\.{I}R HESAPSAL S\"{O}ZL\"{U}\v{G}\"{U}N TASARIMI VE\\ 
GER\c{C}EKLE\c{S}T\.{I}R\.{I}LMES\.{I}\\ 

\vspace{0.5cm}
Abdullah Kurtulu\c{s} Yorulmaz \\
Bilgisayar ve Enformatik M\"{u}hendisli\u{g}i, Y\"{u}ksek Lisans\\
Tez Y\"{o}neticisi: Yrd. Do\c{c}. Dr. Kemal Oflazer\\
\c{S}ubat, 1997\\

\end{center}

B\"{u}t\"{u}n do\v{g}al dil i\c{s}leme sistemleri (\"{o}rne\v{g}in
\c{c}\"{o}z\"{u}mleyiciler, \"{u}reticiler, metin
i\-\c{s}a\-ret\-le\-yi\-ci\-ler) dildeki kelimeler hakk{\i}nda, bir
s\"{o}zl\"{u}\v{g}e eri\c{s}meye ihtiya\c{c} duyarlar. Bu tezde,
T\"{u}rk\c{c}e'de do\v{g}al dil i\c{s}leme i\c{c}in bir s\"{o}zl\"{u}k
mimarisi sunulmu\c{s}tur. Bir kelimenin y\"{u}zeysel hali ve
k{\i}s{\i}tlay{\i}c{\i} di\v{g}er \"{o}zellikler i\c{c}eren sorguya
kar\c{s}{\i}l{\i}k, s\"{o}zl\"{u}k, verilen kelimenin y\"{u}zeysel
halinin, bu k{\i}s{\i}tlay{\i}c{\i} \"{o}zellikleri sa\v{g}layan her
\c{c}\"{o}z\"{u}m\"{u} i\c{c}in
bi\c{c}imbirimsel/s\"{o}z\-di\-zin\-sel, \c{s}ekilsel ve anlamsal
\"{o}zellikler i\c{c}eren bir \"{o}zellik yap{\i}s{\i} \"{u}retir.
S\"{o}zl\"{u}k, \"{o}zellik temelli temsil, kal{\i}t{\i}m ve
birle\c{s}tirme gibi \c{c}a\v{g}da\c{s} yakla\c{s}{\i}mlara
dayan{\i}r.  \.{I}ki bilgi kayna\v{g}{\i} kullan{\i}r: bir
s\"{o}zc\"{u}kyap{\i}sal i\c{s}leyici ve T\"{u}rk\c{c}e'nin
b\"{u}t\"{u}n a\c{c}{\i}k ve kapal{\i} kelime gruplar{\i}n{\i}
i\c{c}eren bir kelime veritaban{\i}.  Sistem, {\em SICStus Prolog}'da
kendi ba\c{s}{\i}na \c{c}al{\i}\c{s}abilecek ve do\v{g}al dil
i\c{s}leme uygulamalar{\i}nda kullan{\i}labilecek \c{s}ekilde
ger\c{c}ekle\c{s}tirilmi\c{s}tir.

\vfill{\em Anahtar s\"{o}zc\"{u}kler\/}: Do\u{g}al Dil \.{I}\c{s}leme,
S\"{o}zl\"{u}k

\newpage

\newpage
\setcounter{page}{5}

\vspace{5cm}

\begin{center}
{\Large \bf ACKNOWLEDGMENTS}
\end{center} 

I am very grateful to my supervisor, Asst. Prof. Kemal Oflazer, for
his invaluable guidance, motivation, and patience during the
development of this thesis. It was a real pleasure to work with him.

I would like to thank Assoc. Prof. Halil Altay G\"{u}venir and
Asst. Prof. \.{I}lyas \c{C}i\c{c}ekli for reading and commenting on
the thesis. 

I owe special thanks to my colleagues Dilek Z. Hakkani and G\"{o}khan
T\"{u}r and other friends Y\"{u}cel Sayg{\i}n, U\v{g}ur
\c{C}etintemel, Gamze D. Tunal{\i} and Murat Bayraktar for their
endless intellectual and moral support during my graduate study.

Finally, I would like to express my deepest gratitude to my parents
for their infinite moral support and patience to me. I dedicate this
thesis to them.

\tableofcontents

\listoffigures

\listoftables

\newpage
\pagenumbering{arabic}
\setcounter{page}{1}
 
\newpage

\chapter{Introduction}
\label{chapter:introduction}

Natural language processing (NLP) is a research area, under which the
aim is to design and develop systems to process, understand, and
interpret natural language. It employs knowledge from various
fields like artificial intelligence (in knowledge representation,
reasoning), formal language theory (in language analysis, parsing),
and theoretical and computational linguistics (in models of language
structure).

There are many applications of NLP such as translation of natural
language text from one language to another, interfacing machines
with speech or speech-to-speech translation, natural language
interfaces to databases, text summarization, text preparation aids such
as spelling and grammar checking/correction, etc.

One of the first applications of NLP is machine translation (MT).
The research was funded by military and intelligence
communities. These systems, what we call {\em first generation},
translate text almost word by word; the result was a failure. But
considering the lack of theories, methods, and resources with
semantics and ambiguities in natural language text, the result is not
surprising~\cite{Gazdar-Mellish}.~\footnotemark\, Today with the
advance of theories, resources, etc., MT is not a dream; even there
are MT systems available in the market.

\footnotetext{
  Consider the following well-known utterance:
  
  \eenumsentence{
  \item Time flies like an arrow.
  \item Fruit flies like a banana.
    }
  The ambiguity in the sentences above  can be resolved by utilizing
  the knowledge: {\em fruit flies} is a meaningful phrase but {\em
    time flies} is not. However, even today, most systems cannot
  access this kind of information.
  }

Many components of NLP systems, like syntactic analyzers, text
generators, taggers, and semantic disambiguators, need knowledge
about words in the language. This information is stored in the {\em
  lexicon}, which is becoming one of the central components of all NLP
systems. 

\begin{figure}[p]
  \centerline{\psfig{figure=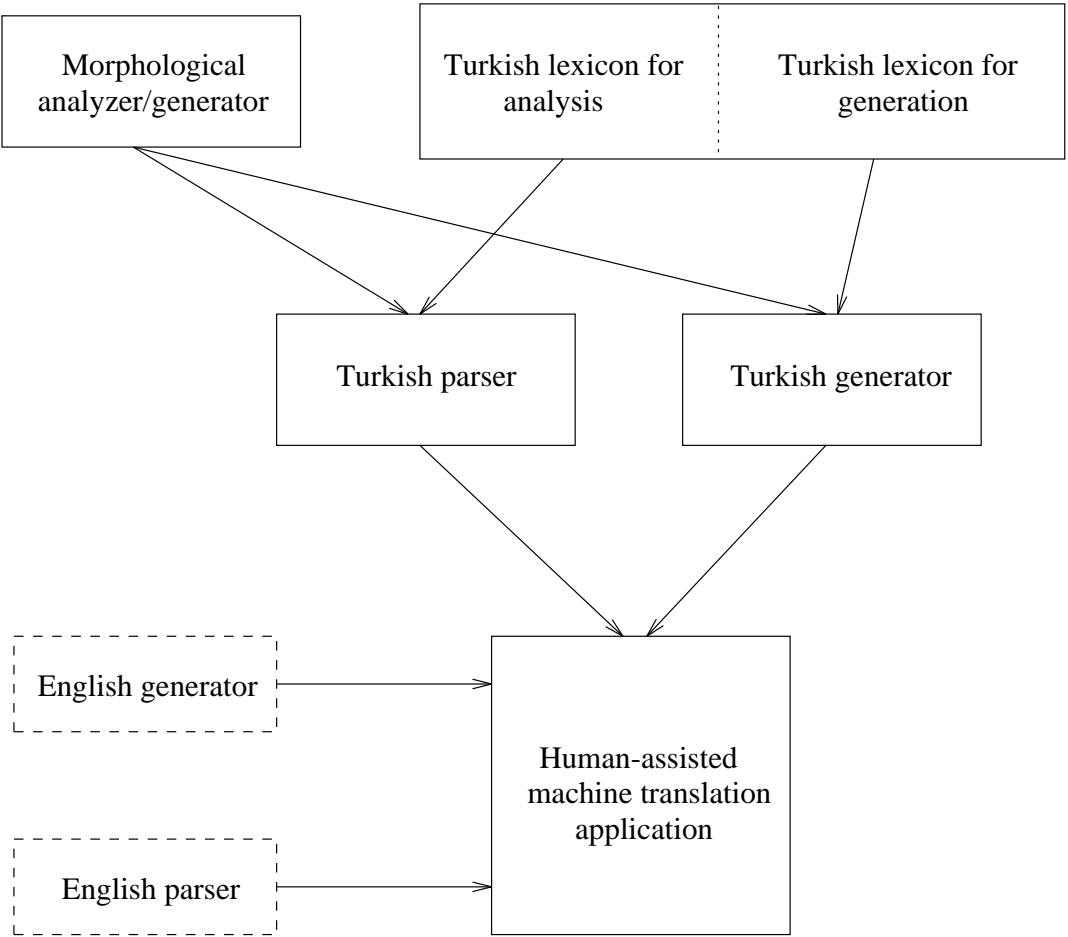}}
  \caption{Simplified architecture of the MT system that would use our
    lexicon.}
  \label{figure-1:TULANGUAGE-architecture}
\end{figure}

In this thesis, we designed and implemented a computational lexicon
for Turkish to be employed in an MT project, which aims to develop
scientific background and tools to translate computer manuals from
Turkish to English and vice versa (see
Figure~\ref{figure-1:TULANGUAGE-architecture} for a simplified
architecture of this system).

A similar work for this project is the design and implementation of a
verb lexicon for Turkish by Y{\i}lmaz~\cite{Yilmaz-Thesis}. This
lexicon contains only verb entries to be utilized in syntactic
analysis and verb sense disambiguation.

Our work aims to develop a generic lexicon for Turkish, which can
provide morphosyntactic, syntactic, and semantic information about
words to NLP systems. The lexicon contains entries for all lexical
categories of Turkish with the information content also covering the
Y{\i}lmaz's work. The morphosyntactic information is not directly
encoded in the lexicon, rather obtained through a morphological
analyzer integrated into the system.

The development of our work is carried out in two steps:

\begin{enumerate}
\item determining the lexical specification for each of the
  lexical categories of Turkish, that is morphosyntactic, syntactic
  and semantic phenomena to be encoded in the lexicon,
\item developing a standalone system that will provide the encoded
  information to NLP systems for a given input.
\end{enumerate}

In this thesis, we present design and implementation of such a
lexicon. 

The outline of the thesis is as follows: In
Chapter~\ref{chapter:lexicon}, we introduce the concept of lexicon
with examples from related work. In Chapter~\ref{chapter:design}, we
present a comprehensive categorization for Turkish lexical types and
associated lexical specification. Next chapter gives the operational
aspects of our lexicon, that is the interface of the system and
algorithms used in producing the result. In
Chapter~\ref{chapter:implementation}, we go through the implementation
of the system and give sample runs.  Chapter~\ref{chapter:conclusion}
concludes and gives suggestions.


\chapter{The Lexicon}
\label{chapter:lexicon}

Lexicon is the collection of morphological/morphosyntactic, syntactic
and semantic information about words in the language. It has been a
critical component of all NLP systems as they move from toy system
operating in demonstration mode to real world applications requiring
wider vocabulary coverage and richer information content.

In this chapter, we will first briefly introduce the concept of
lexicon and the need for it. Then, we will give the role of lexicon
in NLP with specific examples from syntactic analysis and verb sense
disambiguation. Finally, we will present an example work, which is on
reaching a common lexical specification in the lexicon among European
languages.


\section{Lexicon}
\label{sec-2:lexicon}

For a long time the lexicon was seen as a collection of idiosyncratic
information about words in the language. As the requirements of NLP
systems, which perform various tasks ranging from speech recognition
to machine translation (MT) in wide subject domains, grow, those systems
need larger lexicons. Even simple applications such as spelling
checkers may require morphological, orthographic, phonological,
syntactic, and semantic information (for disambiguation) with realistic
vocabulary coverage~\cite{Briscoe}. For instance, The Core Language
Engine, which is a unification-based parsing and generation system for
English, has a lexicon containing 1800 senses of 1200 words and
phrases~\cite{Carter}. Thus, the lexicon design and development has
become the one of the central issues for all NLP systems.

There are two ways to develop the information content of a lexicon:
hand-crafting and use of machine-readable resources. The first is the
classical and costly way of developing the content. However, there is a
growing trend to use existing machine-readable resources, such as
electronic dictionaries and text corpora, to derive useful
information. Research in this area has yielded significant results in
extracting morphosyntactic and syntactic information, but the results
in semantic information side are not yet satisfactory~\cite{Nirenburg}.


\section{The Role of Lexicon in NLP}
\label{sec-2:lexicon-in-NLP}

NLP systems need to access lexical knowledge about words in the
language. This information can be morphosyntactic, such as stem,
inflectional and derivational suffixes (by means of listing them
explicitly or generation), syntactic, such as grammatical category and
complement structures, and semantic, such as multiple senses and
thematic roles.  Depending on the NLP task being performed, other
information can be utilized such as mapping between lexical units and
ontological concepts for transfer tasks in MT, text planning
information for generation, orthographic and phonological information
for speech processing applications.

In the following two sections, we will describe the role of lexicon
in syntactic analysis and verb sense disambiguation.


\subsection{The Role of Lexicon in Syntactic Analysis}
\label{sec-2:syntactic_analysis}

The following paragraph is taken from Zaenen and
Uszkoreit~\cite{Zaenen-Uszkoreit}, which briefly describes text
analysis:

``We understand larger textual units by combining our understanding of
smaller ones. The main aim of linguistic theory is to show how these
units of meaning arise out of the combination of the smaller
ones. This is modeled by means of a grammar. Computational linguistics
then tries to implement this process in an efficient way. It is
traditional to subdivide the task into syntax and semantics, where
syntax describes how the different formal elements of a textual unit,
most often the sentence, can be combined and semantics describes how
the interpretation is calculated.''

The grammar consists of two parts: a set of rules describing how to
combine small textual units into larger ones, and a lexicon containing
information about those small units. In recent theories of grammar,
the first part is reduced to one or two general principles, 
and the rest of the information is encoded in the lexicon. 

Now we will briefly describe the analysis lexicon in KBMT-89
system~\cite{Goodman-Nirenburg}. KBMT-89 is a knowledge-based machine
translation system, in which source language text is analyzed into a
language independent representation (namely {\em interlingua}) and
generated in the target language.

There are two other methods used in MT other than interlingua method:
{\em direct} and {\em transfer method}. In the former one, the source
text is directly translated to target language, almost word by word
with some arrangements, however, in the second one source text is
analyzed into an abstract representation, which is then transfered
into another abstract representation for the target language, and
finally generated as the target language text. Knowledge-based MT
requires more syntactic and semantic information, so a larger and
richer lexicon, than the other methods, such as language independent
knowledge-base for modeling the subworld of translation, etc.

Knowledge acquisition in KBMT-89 is manual, but aided with special
tools so that partial automation is achieved. KBMT-89 uses three types
of lexicon: 

\begin{enumerate}
\item {\em concept lexicon}, which stores semantic information
  for parsing and generation, 
\item {\em generation lexicon}, which contains information for the
  open-class words (e.g., nouns, which accept new words in time), in
  the target language (in that special case, it is Japanese), and
\item {\em analysis lexicon}, which stores morphological and syntactic
  information, word-to-concept mapping rules, and information for the
  mapping case role structures (thematic roles) to subcategorization
  patterns.
\end{enumerate}

Each entry in the analysis lexicon contains the following information:
a word, its syntactic category, inflection, root-word form, syntactic
features, and mappings. Syntactic features and mappings can be
specified locally or through inheritance by properly setting a pointer
to a class in the syntactic feature or structural mapping hierarchy.

Here are two example entries from the English analysis lexicon 
for the verb and noun interpretations of {\em note}: 

\begin{small}
\begin{verbatim}
(``note'' (CAT V)
    (CONJ-FORM INFINITIVE)
    (FEATURES
        (CLASS CAUS-INCHO-VERB-FEAT)
        (all-features
            (*OR*
                ((FORM INF) (VALENCY (*OR* INTRANS TRANS)) (COMP-TYPE NO)
                 (ROOT NOTE))
                ((PERSON (*OR* 1 2 3)) (NUMBER PLURAL) (TENSE PRESENT)
                 (FORM FINITE) (VALENCY INTRANS TRANS)
                 (COMP-TYPE NO) (ROOT NOTE))
                ((PERSON (*OR* 1 2)) (NUMBER SINGULAR) (TENSE PRESENT)
                 (FORM FINITE) (VALENCY INTRANS TRANS))
                 (COMP-TYPE NO) (ROOT NOTE))))
    (MAPPING (local
                 (HEAD (RECORD-INFORMATION)))
             (CLASS AG-TH-VERB-MAP)))
\end{verbatim}
\end{small}

In the frame above, first three slots give the headword, its category
and word form, that is {\em note}, {\em verb} and {\em infinitive},
respectively.  The next slot, {\tt FEATURES}, gives the syntactic
features by inheriting the features of the class {\tt
  CAUS-INCHO-VERB-FEAT}, which are the features of {\em
  causative-inchoative} verb class, and adding other features locally,
such as valence, root word form, and agreement marker in each of the
three cases, as arguments of {\tt *OR*}. The last slot, {\tt MAPPING},
gives word-to-concept mapping, that is the verb {\em note} is mapped
to the ontological concept {\tt RECORD-INFORMATION} in the concept
lexicon, and mapping of case role structures to subcategorization
patterns by inheriting from {\tt AG-TH-VERB-MAP} class in the
structural mapping hierarchy, which is the mapping for {\em
  agent-theme} verbs.

%
%

\begin{small}
\begin{verbatim}
(``note'' (CAT N)
    (CONJ-FORM SINGULAR)
    (FEATURES
        (CLASS DEFAULT-NOUN-FEAT)
        (all-features
            (PERSON 3) (NUMBER SINGULAR) (COUNT YES) (PROPER NO)
            (MEAS-UNIT NO) (ROOT NOTE)))
    (MAPPING
        (local 
            (HEAD (MENTAL-CONTENT)))
        (local 
            (HEAD (TEXT-GROUP (CONVEY (COMMUNICATIVE-CONTENT)))))
        (CLASS OBJECT-MAP)))
\end{verbatim}
\end{small}

The frame above states that the noun {\em note} is {\em singular},
inherits all the syntactic features of the class {\tt
  DEFAULT-NOUN-FEAT} in addition to its local features; for example
its agreement marker is {\em 3sg}, it is countable and not a proper
noun. The {\tt MAPPING} slot gives its mapping to the entries in the
concept lexicon, that is {\em note} describes a mental content or a
text group conveying a communicative content. It also inherits all the
word-to-concept mappings of the class {\tt OBJECT-MAP}.


\subsection{The Role of Lexicon in Verb Sense Disambiguation}
\label{sec-2:verb-sense}

The second specific usage of the lexicon that we will describe is in
verb sense disambiguation specifically for Turkish due to the work by
Y{\i}lmaz~\cite{Yilmaz-Thesis}. 

Verb is the most important component in the sentence; it gives the
predicate. Thus, resolving lexical ambiguities concerning the verb is
very important in syntactic analysis, especially in MT. There are
three kinds of lexical ambiguities:

\begin{enumerate}
\item {\em polysemy}, in which case a lexical item has more than one
  senses close to each other, as in {\em para ye-} ({\em cost a lot
    of money}) and {\em kafay{\i} ye-} ({\em get mentally deranged}).
  For example, {\em T\"{u}rk Dil Kurumu Dictionary} gives 40 senses for
  the verb {\em \c{c}{\i}k} and 32 senses for the verb {\em at}.
\item {\em homonymy}, in which case the words have more than one
  interpretation having no obvious relation among them, e.g., {\em
    vurul-} has two interpretations: {\em fall in love with} and
  {\em be wounded}.
\item {\em categorical ambiguity}, in which case the words have
  interpretations belonging to more than one category, as in {\em ek}
  (noun, {\em appendix/suffix}) and (verb, {\em sow}).
\end{enumerate}

The claim in Y{\i}lmaz's work is that by trying to match the
morphological, syntactic, and semantic information in the sentential
context of a verb (i.e., the information in its complements) with the
corresponding information of the verb entries in the lexicon, the
correct interpretation and sense
of the verb can be determined. 
%
%
For instance, consider the following example:

\eenumsentence{
\item
   \shortex{3}
   {Memur & para & yedi.}
   {{\tt official} & {\tt money} & {\tt accept bribe+PAST+3SG}}
   {`The official accepted bribe.'}
\item
   \shortex{4}
   {Araba  & \c{c}ok & para & yedi.}
   {{\tt car} & {\tt a lot of} & {\tt money} & {\tt cost+PAST+3SG}}
   {`The car costed a lot.'}
}

In the sentences above, the verb {\em ye-} is used in two different
senses as {\em accept bribe} and {\em cost a lot}. The encoding in the
lexicon for the first sense states that the head of the direct
object's noun phrase is {\em para} with no possessive or case marking,
and the subject is human.  For the second sense, the head of the
direct object's noun phrase is {\em para} and the subject is
non-human. By applying those constraints, the correct interpretation
can be determined. In the application of semantic constraints,
however, an ontology (i.e., knowledge-base, which describes the
objects, events, etc. in a subject domain) for nouns should be
utilized, for example, in testing whether {\em memur} is human or not.

The lexicon consists of a list of entries for verbs. Each entry is
identified with its headword, and contains a list of argument
structures, in which there are the labels of the arguments,  
morphological, syntactic, and semantic constraints, and a list of senses 
associated with those argument structures. Each sense has another set of 
constraints specific for that sense and some descriptive information,
such as semantic category, mapping of thematic roles to subcategorization
patterns, concept name, etc. 

Below, we provide the lexicon entry for the verb {\em ilet-}, which has two 
argument structures and three senses (i.e., {\em conduct}, {\em
  convey}, and {\em tell}). In order to save space, we omit the second
argument structure and the last sense associated with it. Here
is the lexicon entry for {\em ilet-}: 

\begin{small}
\begin{verbatim}
((HEAD . "ilet")
 (ENTRY
     (ARG-ST1
        (ARGS
            (SUBJECT 
                (LABEL . S)
                (SEM . T)
                (SYN OCC S OPTIONAL)
                (MORPH . T))
            (DIR-OBJ 
                (LABEL . D)
                (SEM . T)
                (SYN OCC D OBLIGATORY)
                (MORPH 
                    (OR
                        (1 CASE D NOM)
                        (2 CASE D ACC)))))
        (SENSES
            (SENSE1
                (CONST POWER-ENERGY-PHYSICALOBJECT D)
                (V-CAT PROCESS-ACTION)
                (T-ROLE 
                    (1 AGENT S)
                    (2 THEME D))
                (C-NAME . "to conduct")
                (EXAMPLE . "katIlar sesi en iyi iletir."))
            (SENSE2
                (CONST . T)
                (V-CAT PROCESS-ACTION)
                (T-ROLE 
                    (1 AGENT S)
                    (2 THEME D))
                (C-NAME . "to convey")
                (EXAMPLE . "yardImI ilettiler."))))
     (ARG-ST2
         ...))
    (ALIAS-LIST ))
\end{verbatim}
\end{small}

In the first argument structure, there are subject and direct object.
The subject is optional, whereas the object is obligatory, and {\em
  nominative} or {\em accusative} case-marked. These are morphological
and syntactic constraints specified in {\tt MORPH} and {\tt SYN}
slots of the arguments, and no other constraint is posed by this
argument structure. There are two senses associated with this
structure. The first poses a semantic constraint in {\tt CONST} slot,
which requires that the direct object must be an instance of {\tt
  POWER-ENERGY-PHYSICALOBJECT} class, like electricity or sound. Then
it gives verb category, which is {\em process-action}, mapping of
thematic roles to subcategorization patterns, which maps agent to
subject and theme to direct object, and concept name, which is {\em to
  conduct}, with an example sentence. The second sense does not pose
any additional constraint. The verb category and thematic role mapping
of this sense are the same with those of the previous one. Then, the
concept name is given as {\em to convey} with an example sentence.


\section{Example Work}
\label{sec-2:exmaple-work}

Due to the growing needs of NLP systems for larger and richer
lexicons, the cost of designing and developing lexicons with broad
coverage and adequately rich information content is getting high. An
example work, which has developed such large lexical resources, may
be the Electronic Dictionary Research (EDR) project (Japan, 1990),
which run for 9 years, costed 100 million US dollars and intended to
develop bilingual resources for English and Japanese containing
200,000 words, term banks containing 100,000 words, and a concept
dictionary containing 400,000 concepts. Although the development is
aided by special tools, the actual effort is due to the researchers
themselves~\cite{Briscoe}.

In order to avoid such high costs, the research institutions and
companies are trying to combine their efforts in developing
publicly available, large scale language resources, which have
adequate information content, and are generic enough (multifunctional)
to satisfy various requirements of wide range of NLP
applications. Examples of such efforts include ESPRIT BRA (Basic
Research Action) ACQUILEX aiming reuse of information extracted from
machine-readable dictionaries, WordNet Project at Princeton, which
created a large network of word senses related with semantic
relations, and LRE EAGLES (Expert Advisory Group on Language
Engineering Standards) project, which tries to reach a common
lexical specification at some level of linguistic detail among
European languages~\cite{Grishman-Calzolari}.

In the rest of this section, we will concentrate on the EAGLES
project. The information given below is mainly received from Monachini and
Calzolari~\cite{EAGLES}. The objective of this work is to propose a
common set of morphosyntactic features encoded in lexicons and corpora
in European languages, namely Italian, English, German, Dutch, Greek,
French, Danish, Spanish, and Portuguese.

The project has gone through three phases:

\begin{enumerate}
\item to survey previous work on encoding morphosyntactic phenomena in
  lexicons and text corpora, e.g., on MULTILEX and GENELEX models,
  etc.,
\item to work on linguistic annotation of text and lexical description
  in lexicons to reach a compatible set of features,
\item to test the common proposal by applying concretely to European
  languages. 
\end{enumerate}

The common set of features came after the completion of the second
phase, and is described in three main levels corresponding to the
level of {\em obligatoriness}:

\begin{enumerate}
\item {\em Level 0} contains only the part-of-speech category, which
  is the unique obligatory feature.
\item {\em Level 1} gives grammatical features, such as gender,
  number, person, etc. These are generally encoded in lexicons and
  corpora, and called {\em recommended features}, which constitute the
  minimal core set of common features.
\item {\em Level 2} is subdivided into two:
  \begin{itemize}
  \item {\em Level 2a} contains features which are common to
    languages, but either not generally encoded in lexicons and
    corpora or not purely morphosyntactic (e.g., countability for
    nouns). These are considered as {\em optional features}.
  \item {\em Level 2b} gives {\em language-specific features}.
  \end{itemize}
\end{enumerate}

The multilayered description, instead of a flat one, gives more
flexibility in choosing the level detail in specification to match the
requirements of applications. As going down from Level 0 to Level 2,
the description reaches finer granularity, and the information
encoded increases. Additionally, this type of description helps to
extend or update the framework. 

The aim of the common proposal is not to pose a complete specification
ready to implement, but to pose a basic set of features and to leave
the rest to language-specific applications. 

The last phase of the project is the testing of the common proposal
in a multilingual framework, namely the MULTEXT project. The aim of
MULTEXT partners is to design and implement a set of tools for
corpus-based research and a corpus in that multilingual framework.
The tasks involved are developing a common specification for the
MULTEXT lexicon and a tagset for MULTEXT corpus. The partners evaluated
the common proposal at Level 1 (recommended features) by also
considering language-specific issues. The result is that the common
set of features fits well to the description of partners, but needs
further language-specific detail.



\chapter{A Lexicon Design for Turkish}
\label{chapter:design}

All natural language processing  systems, such as parsers, generators, 
taggers, need to access a lexicon of the words in the language. The
information provided by the lexicon includes:

\begin{itemize}
\item morphosyntactic,
\item syntactic, and 
\item semantic information.
\end{itemize}

In this thesis, we have designed a comprehensive lexicon for Turkish,
and integrated it with a morphological processor,
so that the overall system is capable of providing the feature
structures for all interpretations of an input word form (with
multiple senses incorporated). 

For instance, consider the input word form {\em kazma}; first, the
morphological processor receives this input, and provides
its analysis to the static lexicon.  There are three possible
interpretations:

\begin{enumerate}
\item {\em kazma} (noun, {\em pickaxe}),
\item {\em kaz}+NEG (verb, {\em don't dig}), and 
\item {\em kaz}+INF (infinitive, {\em digging}), 
\end{enumerate}

for which the static lexicon produces feature structures
for all senses of the root words involved. 
Moreover, the lexicon allows the interfacing system to constraint the
output. 
For example, the final category feature of the root word in the input
surface form can be restricted to, say, verb. In this case, only
information about the second interpretation, {\em don't dig}, will be
released by the system. Chapter~\ref{chapter:operational-aspects}
describes this process in detail.

By separating the system into two parts, that is a morphological
analyzer and a static lexicon, we make use of the morphological
processor previously implemented and abstract the process of
parsing surface forms. Hence, designing a static lexicon and
interfacing it with the morphological processor is sufficient to
construct a lexicon system.

In this chapter we will present the detailed design of our static
lexicon, that is the associated feature structures with each of the
lexical categories in Turkish. The procedural aspects (i.e., how
feature structures are produced) are described in
Chapter~\ref{chapter:operational-aspects}. 
We will first introduce the main lexical categories, then describe
each one in detail with the associated feature structures.

\section{Lexicon Architecture}\label{sec-3:architecture}

\begin{figure}[tp]
  \centerline{\psfig{figure=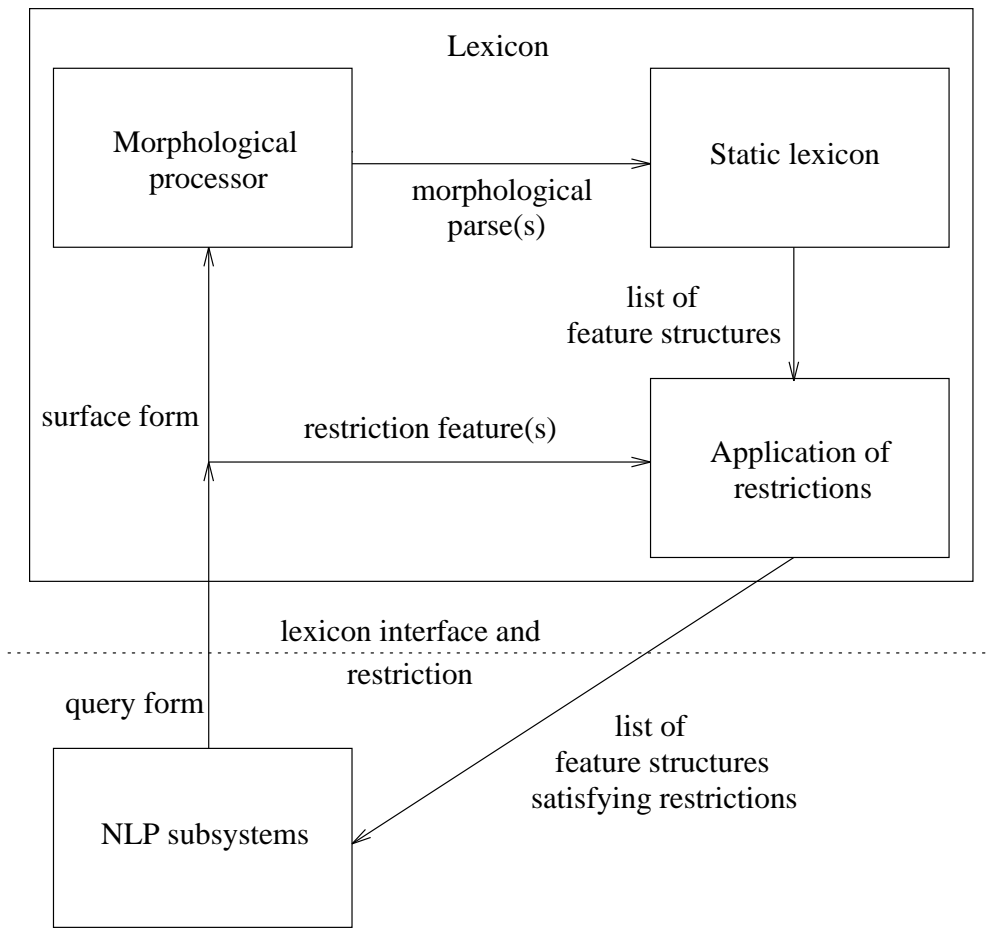}}
  \caption{Architecture of the lexicon.}
  \label{figure-3:architecture}
\end{figure}

The Figure~\ref{figure-3:architecture} briefly describes the
architecture of our lexicon, which consists of a morphological
processor, a static lexicon, and a module applying restrictions.

The input to the system is a query form, which consists of two parts:
a word form and a set of features placing constraints in the output. 
The word form is first received and processed by the morphological
processor, whose output is the possible interpretations of the word
form. 
Then, the static lexicon attaches features to all senses of the root
words of these interpretations, and outputs the feature
structures. But before the result is released, the feature structures
that do not satisfy the restrictions are eliminated, and the rest is
the actual output of the system. The details of this procedure are
given in Chapter~\ref{chapter:operational-aspects}.

\section{Lexical Representation Langugage}\label{sec-3:lexical_r_language}

The lexical representation language that we will use in the rest of
this chapter is feature structures. A feature structures is a list of
$<${\em feature name}:{\em feature value}$>$ pairs, in which at most one
pair with a given feature name can be present. The value of a feature
name may be an atom or a feature structure again. Here are some
examples of feature structures:\footnotemark

\footnotetext{
  See Shieber~\cite{Shieber} for a detailed description of feature
  structures.
  }

\avmBegin
\[{}
  F & a\\
  G & b\\
\]
\avmEnd

\avmBegin
\[{}
  F &
    \[{}
      G & a\\
      H & b\\
    \]\\
  I & c\\
\]
\avmEnd

\section{Lexical Categories}\label{sec-3:lexical_categories}

\begin{figure}[htb]
  \centerline{\psfig{figure=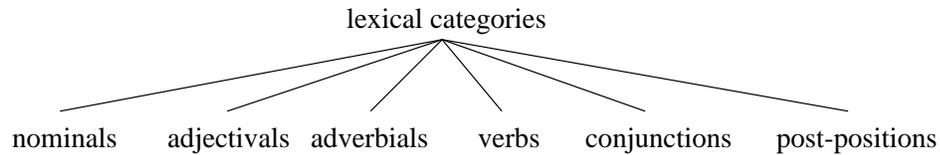}}
  \caption{The main lexical categories of Turkish.}
  \label{figure-3:lexical_types}
\end{figure}

Figure~\ref{figure-3:lexical_types} shows the main lexical categories of
Turkish in our lexicon. All the lexicon categories are depicted in 
Tables~\ref{table-3:lexicon-categories-1}~and~\ref{table-3:lexicon-categories-2}
on page~\pageref{table-3:lexicon-categories-1}.

Each word in the lexicon has the following feature structure:

\avmBegin
\[{word}
  CAT & 
      \[{}
        MAJ & {\em maj}\\
        MIN & {\em min} & (default: none)\\
        SUB & {\em sub} & (default: none)\\
        SSUB & {\em ssub} & (default: none)\\
        SSSUB & {\em sssub} & (default: none)\\        
      \]\\
   MORPH & 
         \[{}
           STEM & {\em stem}\\
           FORM & lexical/derived (default: lexical)\\
         \]\\
   SEM &
       \[{}
         CONCEPT & {\em concept}\\
       \]\\
   PHON & {\em phon}\\
\]
\avmEnd

Thus, each word has category information in CAT feature as a 5-tuple
describing major, minor and subcategories, STEM and FORM as
morphosyntactic features, CONCEPT as semantic fetaure, and phonology.
The major and minor categories and the concept, which uniquely
determine the word with its sense are given in this feature structure.
Additionally, the form, which take {\em lexical} or {\em derived}
values, the stem and the phonology, which is the combination of the
stem and inflections are also present in this structure, e.g., {\em
  kitap} ({\em book}) vs. {\em kitaplar{\i}m} ({\em my books}).

\section{Nominals}\label{sec-3:nominals}

\begin{figure}[htb]
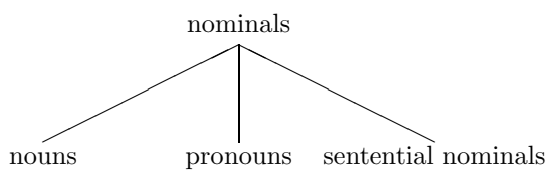

  \fnsCBegin
  \leaf{nouns}
  \leaf{pronouns}
  \leaf{sentential nominals}
  \branch{3}{nominals}
  \tree
  \fnsCEnd
  \caption{Subcategories of nominals.}
  \label{figure-3:nominals}
\end{figure}

This section describes the representation of nominals in our
lexicon. As shown in Figure~\ref{figure-3:nominals}, nominals are divided into
three subcategories: 

\begin{itemize}
\item nouns,
\item pronouns,
\item sentential heads which function as nominals.
\end{itemize}

Figure~\ref{table-3:nominal-categories} gives the detailed
categorization for the nominal category.\footnotemark

\footnotetext{
  The three subcategories of infinitives and the two subcategories of
  participles represent the verbal forms derived using the suffixes
  {\em -mA}, {\em -mAk}, {\em -yH\c{s}}, {\em -dHk}, and {\em
    -yAcAk}. These will be explained later in detail.

  The notation for suffixes follows this convention: {\em A}
  and {\em H} represent unrounded (i.e., \{{\em a}, {\em e}\}) and
  high vowels (i.e., \{{\em {\i}}, {\em i}, {\em u}, {\em
    \"{u}}\}), respectively. The first {\em y} in the suffixes may
  drop.
  }

\begin{figure}[htb]
\begin{center}
\begin{footnotesize}
\begin{tabular}{|l|l|l|l|l|}\hline
{\em maj} & {\em min} & {\em sub} & {\em ssub} & {\em sssub}\\ \hline 
                                                               \hline
nominal & noun     & common         &            & \\ \hline
        &          & proper         &            & \\ \hline
        & pronoun  & personal       &            & \\ \hline
        &          & demonstrative  &            & \\ \hline
        &          & reflexive      &            & \\ \hline
        &          & indefinite     &            & \\ \hline
        &          & quantification &            & \\ \hline
        &          & question       &            & \\ \hline
        & sentential & act          & infinitive & ma         \\ \hline
        &          &                &            & mak        \\ \hline
        &          &                &            & y{\i}\c{s} \\ \hline
        &          & fact           & participle & d{\i}k     \\ \hline
        &          &                &            & yacak      \\ \hline 
\end{tabular}
\end{footnotesize}
\end{center}
\caption{Lexicon categories of nominals.}
\label{table-3:nominal-categories}
\end{figure}

Each nominal has the following additional features, which represent
the inflections of the word:

\avmBegin
\[{nominal}
  MORPH & 
        \[{}
          CASE & {\em case} & (default: none)\\
          AGR  & {\em agr}  & (default: none)\\
          POSS & {\em poss} & (default: none)\\
        \]\\        
\]
\avmEnd

A nominal may be case-marked as 

\begin{itemize}
\item nominative, 
\item accusative, 
\item dative,
\item locative, 
\item ablative,
\item genitive, 
\item instrumental, 
\item equative. 
\end{itemize}

{\em Third person singular} and {\em plural} suffixes are the possible
values  for the agreement marker of nouns and sentential
heads. Pronouns may take {\em first}, {\em second}, and {\em third
  person singular} and {\em plural} agreement markers.
All three types of nominals may take possessive suffix, which is one
of the six person suffixes and {\em none}.

In the following sections we will describe the subcategories of
nominals in detail.

\subsection{Nouns}\label{sec-3:nouns}

\begin{figure}[htb]
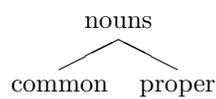

  \fnsCBegin
  \leaf{common}
  \leaf{proper}
  \branch{2}{nouns}
  \tree
  \fnsCEnd
  \caption{Subcategories of nouns.}
  \label{figure-3:nouns}
\end{figure}

Nouns denote the entities in the world, such as objects, events,
concepts, etc.
As shown in Figure~\ref{figure-3:nouns}, nouns can be further divided
into two subcategories as {\em common} and {\em proper nouns}. 
These are described in detail in the next two sections.

\subsubsection{Common Nouns}

\begin{figure}[htb]
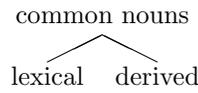

  \fnsCBegin
  \leaf{lexical}
  \leaf{derived}
  \branch{2}{common nouns}
  \tree
  \fnsCEnd
  \caption{Forms of common nouns.}
  \label{figure-3:common-nouns}
\end{figure}

\begin{figure}[t]
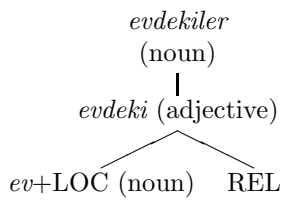

  \fnsCBegin
  \leaf{{\em ev}+LOC (noun)}
  \leaf{REL}
  \branch{2}{{\em evdeki} (adjective)}
  \branch{1}{{\em evdekiler}\\ (noun)}
  \tree
  \caption{Derivation history of {\em evdekiler}.}
  \label{figure-3:evdekiler}
  \fnsCEnd
\end{figure}

Common nouns denote classes of entities.
Figure~\ref{figure-3:common-nouns} depicts the two forms of common
nouns: {\em lexical} and {\em derived}.
Only lexical common nouns are represented in our lexicon as lexical
entries, however, the system can produce feature structures for
derived forms. For example, computation of the feature structure for
{\em evdekiler} ({\em those that are at home}) requires
the retrieval of the feature structure of the noun {\em ev} ({\em
  home}) and the derivation of it to an adjective ({\em evdeki}
({\em that is at home})) and then to the noun {\em evdekiler} (see the
derivation tree for {\em evdekiler} in Figure~\ref{figure-3:evdekiler}).

Common nouns have the following additional features: subcategorization
and a set of semantic properties such as countability and
animateness. 

\avmBegin
\[{common}
  SYN &
      \[{}
        SUBCAT & \{$constraint_1$, \ldots,\\
                   $constraint_i$, \ldots,\\
                   $constraint_n$\}\,(default: none)\\
      \]\\
  SEM & 
      \[{}
        MATERIAL  & +/$-$ & \\
        UNIT      & +/$-$ & \\
        CONTAINER & +/$-$ & \\
        COUNTABLE & +/$-$ & \\
        SPATIAL   & +/$-$ & \\
        TEMPORAL  & +/$-$ & \\
        ANIMATE   & +/$-$ & \\
      \]\\
\]
\avmEnd

\avmBegin
\[{$constraint_i$}
  CAT &
      \[{}
        MAJ & nominal\\
        MIN & {\em min}\\
        SUB & {\em sub}\\
        SSUB & {\em ssub}\\
        SSSUB & {\em sssub}\\
      \]\\
  MORPH &
        \[{}
          CASE & {\em case}\\
          POSS & {\em poss}\\
        \]\\
  SEM &
      \[{}
      \]\\
\]
\avmEnd

The semantic features may only take + or $-$ values. 
This is on the sense basis, since senses may have different semantic
properties; for example, {\em ekin} ({\em culture}) is an abstract
entity, whereas {\em ekin} ({\em crop}) is not.
The default value for the semantic features is $-$.

The subcategorization information consists of a list of constraints on
any complement of the common noun. The application of constraints is
in {\em disjunctive} fashion.
This concept will be extended to cover more than one complement (e.g.,
subject, objects, etc.) in Section~\ref{sec-3:verbs}, when the {\em verb}
category is introduced. Constraints on the complements of common nouns
are of three types: category, case and possessive markings, and 
semantic properties. Note that the constraint structure for common
nouns is simpler than that for verbs. For instance, constraint
structure for the current category does not constrain the stem and
agreement features of the arguments.

In the next sections we will describe the two forms of common nouns in
detail with examples.

\paragraph{Lexical Common Nouns}

As mentioned above, this form of common nouns are present in the
lexicon, and the retrieval does not involve any computation of
features. 
The following are examples of common nouns in lexical form: {\em kum}
({\em sand}), {\em kalem} ({\em pencil}), {\em ihtiya\c{c}} ({\em
  need}), {\em sabah}  ({\em morning}), {\em \c{c}ar\c{s}amba} ({\em
  Wednesday}), {\em ilkbahar} ({\em spring}), {\em a\c{s}a\v{g}{\i}}
({\em bottom}).

As an example, consider the common noun {\em ihtiyac{\i}} ({\em
  his}/{\em her}/{\em its need}), as used
in~(\ref{example:lexical-common-1}):\footnotemark

\footnotetext{
  Note that some of the features are not shown; they take the default
  values specified. 
  }

\eenumsentence{
  \label{example:lexical-common-1}
\item
    \shortexnt{5}
  {Utku'nun & senin   & bu &   i\c{s}i & yapmana}
  {{\tt Utku+GEN} & {\tt you+GEN} & {\tt this} & {\tt job+ACC} & {\tt
      do+INF+P2SG}}
  \newline
  \shortex{2}
  {ihtiyac{\i}   & var.}
  {{\tt need+P3SG} & {\tt existent+PRES+3SG}}
  {`Utku needs you to do this job.'}
\item
  \shortex{5}
  {Bunun & i\c{c}in & sana/Bilge'ye & ihtiyac{\i}m{\i}z & var.}
  {{\tt this+GEN} & {\tt for} & {\tt you/Bilge+DAT} & {\tt need+P1PL} & 
    {\tt existent+PRES+3SG}}
  {`We need you/Bilge for this.'}
}

\avmBegin
\[{lexical common}
  CAT & 
      \[{}
        MAJ & nominal\\
        MIN & noun\\
        SUB & common\\
      \]\\
  MORPH & 
      \[{}
        STEM & ``ihtiya\c{c}''\\
        FORM & lexical\\
        CASE & nom\\
        AGR  & 3sg\\
        POSS & 3sg\\
      \]\\
   SYN &
       \[{}
         SUBCAT & \{$constraint_1, constraint_2$\}\\  
       \]\\
   SEM & 
       \[{}
         CONCEPT & \#ihtiya\c{c}-(need)\\
       \]\\
   PHON & ``ihtiya\c{c}''\\
\]
\avmEnd

\avmBegin
\[{$constraint_1$}
  CAT &
      \[{}
        MAJ & nominal\\
        MIN & \{{\rm noun, pronoun}\}\\ 
      \]\\
  MORPH &
        \[{} 
          CASE & dat\\
        \]\\
\]
\avmEnd

\avmBegin
\[{$constraint_2$}
  CAT &
      \[{}
        MAJ & nominal\\
        MIN & sentential\\
        SUB & act\\
        SSUB & infinitive\\
        SSSUB & ma\\
      \]\\
  MORPH & \[{}
            CASE & dat\\
          \]\\
\]
\avmEnd


The feature structure of {\em ihtiyac{\i}} contains information
stating that {\em ihtiyac{\i}} is a common noun in lexical form,
inflected from {\em ihtiya\c{c}} with {\em 3sg} agreement and
possessive markers. It also specifies that the complement of {\em
  ihtiyac{\i}} should be case-marked as {\em dative} and may be in
one the two forms: noun or pronoun, and infinitive derived with the
suffix {\em -mA}. 
Example sentences in~(\ref{example:lexical-common-1}) depict these
usages.

The following is another example, the common noun {\em geceye} ({\em to
  the night}), as used in~(\ref{example:lexical-common-noun-2}):

\eenumsentence{
\item[] 
  \label{example:lexical-common-noun-2}
  \shortexnt{5}
  {D\"{u}n & geceye & kadar & oraya & gitmek}
  {{\tt yesterday} & {\tt night+DAT} & {\tt until} & {\tt there+DAT} &
      {\tt go+INF}}
  \newline
  \shortex{3}
  {konusunda & karar vermi\c{s} & de\v{g}ildim.}
  {{\tt topic+P3SG+LOC} & {\tt decide+NARR} & {\tt NOT+PAST+1SG}}
  {`I had not decided on going there until last night.'}
  }

\avmBegin
\[{lexical common}
  CAT &
      \[{}
        MAJ & nominal\\
        MIN & noun\\
        SUB & common\\
      \]\\
  MORPH &
        \[{}
          STEM & ``gece''\\
          FORM & lexical\\
          CASE & dat\\
          AGR & 3sg\\
          POSS & none\\
        \]\\
  SYN &
      \[{}
        SUBCAT & none\\
       \]\\
  SEM &
      \[{}
        CONCEPT & \#gece-(night)\\
        COUNTABLE & +\\
        TEMPORAL & +\\
      \]\\
  PHON & ``geceye''\\
\]
\avmEnd

The feature structure above gives the following information: {\em
  geceye} is a common noun in lexical form, inflected from the common
noun {\em gece} with {\em 3sg} agreement and {\em dative} case
markers. It is countable and states temporality.

\paragraph{Derived Common Nouns}

Derived forms of common nouns are not represented directly in the
lexicon. However, in order to produce feature structures, the lexicon
employs the derivation information provided by the morphological
processor. 
This information mainly consists of the target category and the
derivational suffixes. The rest of the information (such as argument
structure, thematic roles,  concept, and stem) are supplied by the lexicon. 
The details of this  process are described in
Chapter~\ref{chapter:operational-aspects}.

Each derived common noun has the following additional features:

\avmBegin
\[{derived common}
  MORPH &
        \[{}
          DERV-SUFFIX & {\em derv-suffix} (default: none)\\
        \]\\
  SEM &
      \[{}
        ROLES & {\em roles} (default: none)
      \]\\
\]
\avmEnd

These give the suffix used in the derivation and the semantic
functions involved. The latter stores the thematic roles of the
lexical verb which is involved somewhere in the derivation process. 
For example, the derived common noun {\em yaz{\i}c{\i}} ({\em
  writer}) has the thematic roles of the verb {\em yaz-} ({\em
  write}), since the derivation process carries the thematic role 
information through categories. The type of this feature's value is
given in Section~\ref{sec-3:verbs}. 

The derivation suffix may take one of the following values: {\em
  -cH}, {\em -cHk}, {\em -lHk}, {\em -yHcH}, {\em -mAzlHk}, {\em
  -yAmAzHk}, {\em -mAcA}, {\em -yAsH} and {\em none}.

However, there is the problem of predicting the semantic properties of
derived common nouns, and this is not an easy task. For example,
consider {\em ak\c{s}amc{\i}} ({\em heavy drinker}) and {\em
  \"{o}\v{g}lenci} ({\em the student attending the afternoon session
  of a school}), which are both derived from common nouns with the
suffix {\em -cH}. 
The semantics is, however, rather unpredictable. The current system
does not attempt to predict those values. 
Instead, the default values are used; but these may not necessarily be
the correct values for the word in consideration. Prediction of these
values is beyond the scope of our work.

There are four types of derivation to derived common nouns:

\begin{itemize}

\item Nominal derivation:
  This type of derivation uses the suffixes {\em -cH}, {\em -cHk},
  \mbox{{\em -lHk}}, as in the examples {\em kap{\i}c{\i}} 
  ({\em doorkeeper}), {\em kitap\c{c}{\i}k} ({\em booklet}), and {\em
    kitapl{\i}k} ({\em bookcase}).
  
  Consider the feature structure for the common noun
  {\em ta\-mir\-cim} ({\em my repairman}), as used in the example sentence
  below:
  
  \eenumsentence{
  \item[]
    \label{example:derived-common-1}
    \shortexnt{4}
    {Her zaman & oldu\v{g}u       & gibi, & tamircim}
    {{\tt always} & {\tt happen+PART+P3SG} & {\tt like } &
      {\tt repairman+P1SG}}
    \newline
    \shortex{4}
    {i\c{s}ini & \c{c}ok & iyi  & yapt{\i}.}
    {{\tt job+P2SG } & {\tt very   } & {\tt well} & {\tt do+PAST+3SG}}
    {`As it is always the case, my repairman did his job very well.'}
    }
  
  \avmBegin
  \[{derived common}
    CAT &
        \[{}
          MAJ & nominal\\
          MIN & noun\\
          SUB & common\\
        \]\\
    MORPH &
          \[{}
            STEM & \@1\\
            FORM & derived\\
            CASE & nom\\
            AGR & 3sg\\
            POSS & 1sg\\
            DERV-SUFFIX & ``c{\i}''\\
          \]\\
     SYN &
         \[{}
           SUBCAT & \@2 none\\
         \]\\
     SEM &
         \[{}
           CONCEPT & f$_{c{\i}}$(\@3)\\
         \]\\
     PHON & ``tamircim''\\
  \]
  \avmEnd

  \avmBegin
  \@1
  \[{lexical common}
    CAT &
        \[{}
          MAJ & nominal\\
          MIN & noun\\
          SUB & common\\
        \]\\
    MORPH &
          \[{}
            STEM & ``tamir''\\
            FORM & lexical\\
          \]\\
    SYN &
        \[{}
          SUBCAT & \@2 none\\
        \]\\
    SEM &
        \[{}
          CONCEPT & \@3 \#tamir-(repair)\\
        \]\\
    PHON & ``tamir''\\
  \]
  \avmEnd

The feature structure for the noun {\em ta\-mir\-cim} is produced first
retrieving the features of {\em tamir} ({\em repair}) and filling a
template for derived common nouns appropriately. Some of the feature
values are obtained from the features of {\em tamir} (e.g.,
subcategorization information), some of them are supplied by the
morphological processor (e.g., inflectional and derivational
suffixes), and the rest is provided by the static lexicon. 

The feature structure above gives the following information: the word
{\em ta\-mir\-cim} is a common noun derived from {\em tamir} with the
suffix {\em cH}, and inflected with {\em 3sg} and {\em 1sg} agreement
and possessive markers, respectively. {\em Tamircim} does not have
subcategorization information. It also includes all the features of
{\em tamir}.

\item Adjectival derivation: 
  Derivation from adjectival uses the suffix {\em -lHk}, e.g., {\em
    iyilik} ({\em goodness}), {\em temizlik} ({\em cleanliness}). But,
  derivation without suffix is also possible as in the following
  examples, though this is not productive:

  \eenumsentence{
  \item[]
    \begin{tabbing}
      -- bor\c{c}lu  \tabSpc \= `that owing debt',\\
      -- ak{\i}ll{\i}   \> `intelligent',\\
      -- geridekine     \> `to the one behind'.\\
    \end{tabbing}
    }

  This is also possible in the case of participles (compare with
  participles in Section~\ref{sec-3:sentential}), such as 

  \eenumsentence{
  \item[]
    \begin{tabbing}
      -- getirdi\v{g}imi \tabSpc \= `the thing that I brought',\\
      -- gelene                  \> `to the one that came/coming'.\\
    \end{tabbing}
    }
  
  As described in the section on qualitative adjectives, this type of
  adjectivals are derived from verbs, and by dropping the head of the
  phrase that they modify and taking their inflectional suffixes, they
  become nominals. An example is given
  in~(\ref{example:derived-common-2}):
  
  \eenumsentence{
    \label{example:derived-common-2}
  \item
    \label{example:derived-common-2-2}
    \shortex{5}
    {Buraya & {\em gelen} & {\em adam{\i}} & g\"{o}rd\"{u}n & m\"{u}?}
    {{\tt here+DAT} & {\tt come+PART} & {\tt man+ACC} & {\tt
        see+PAST+2SG} & {\tt QUES}}
    {`Did you see the man that came here?'}
  \item
    \label{example:derived-common-2-3}
    \shortex{4}
    {Buraya   & {\em geleni} & g\"{o}rd\"{u}n & m\"{u}?}
    {{\tt here+DAT} & {\tt come+PART+ACC} & {\tt see+PAST+2SG  } & {\tt
        QUES}}
    {`Did you see the one that came here?'}  
    }

  In sentence (\ref{example:derived-common-2-2}), the verbal form of
  gapped relative clause, {\em buraya gelen}, acting as the modifier of
  {\em adam} ({\em man}) takes the inflections of {\em adam}, and
  functions as a nominal.

  There are two types of participles (see Underhill
  \cite{Underhill-TG}):

  \begin{itemize}
  \item {\em subject} (such as {\em gelen adam} 
    ({\em the man that came/is coming})),
  \item {\em object} (such as {\em getirdi\v{g}im kitap} 
    ({\em the book that I brought})).
  \end{itemize}
  
  In order for an object participle to be used as a nominal
  (specifically common noun), the verb from which the adjectival is
  derived should take a direct object. 
  Otherwise, the nominal represents a fact. For example, the verb,
  {\em gel-} ({\em come}), may not take a direct 
  object argument, thus the nominal, {\em geldi\v{g}ini} in
  (\ref{example:derived-common-3-1}) represents a fact. 
  In (\ref{example:derived-common-3-2}), however, the nominal, {\em
    getirdi\v{g}ini}, has two readings: a fact and a derived common
  noun.

  \eenumsentence{
  \item
    \label{example:derived-common-3-1}
    \shortex{3}
    {Taner'in  & geldi\v{g}ini  & biliyorum.}
    {{\tt Taner+GEN} & {\tt come+PART+P3SG} & {\tt know+PROG+1SG}}
    {`I know that Taner came.'}
  \item
    \label{example:derived-common-3-2}
    \shortex{3}
    {Taner'in  & getirdi\v{g}ini & biliyorum.}
    {{\tt Taner+GEN} & {\tt bring+PART+P3SG} & {\tt
        know+PROG+1SG}\vspGloss}
    {`I know that Taner brought something.'\\
      `I know the thing that Taner brought.'}
    }

\item Verb derivation:
  This derivation type uses the suffixes {\em -yHcH}, {\em -mAcA},
  {\mbox {\em -mAzlHk}}, {\em -yAmAzlHk}, and {\em -yAsH}, as used in the
  following example nouns:
  {\em yaz{\i}c{\i}} ({\em writer}), {\em ko\c{s}ucu} ({\em
    runner}), {\em ko\c{s}u\c{s}turmaca} ({\em rush}/{\em hurry}), 
  {\em \c{c}ekememezlik} ({\em envy}), {\em kahrolas{\i}} ({\em
    damnable}).

\item Post-position derivation: 
  Derivation from post-positions
  do not use any suffix, e.g., {\em az{\i}n{\i}} ({\em the one that is
    little}),  {\em yukar{\i}s{\i}na} ({\em to the one that is above}).

\end{itemize}

\subsubsection{Proper nouns}

Proper nouns  are used to refer to unique entities in the world.
The only additional feature that proper nouns have states that they
 are always {\em definite}, as in the examples {\em Kurtulu\c{s}},
  {\em Kemal}, {\em Oflazer}, {\em Bilkent}, and {\em Ankara}.

\avmBegin
\[{proper}
  SEM &
      \[{}
        DEFINITE & +\\
      \]\\
\]
\avmEnd

As used in~(\ref{example:proper-noun-1}), the following is the feature
structure of the proper noun {\em Kurtulu\c{s}}:

\eenumsentence{
\item[]
  \label{example:proper-noun-1}
  \shortex{6}
  {Kurtulu\c{s} & yar{\i}m & saat & i\c{c}inde & burada & olacak.}
  {{\tt Kurtulu\c{s}} & {\tt half} & {\tt hour} & {\tt in} & {\tt
      here+LOC} & {\tt be+FUT+3SG}}
  {`Kurtulu\c{s} will be here in half an hour.'}
  }

\avmBegin
\[{proper}
  CAT & 
      \[{}
        MAJ & nominal\\
        MIN & noun\\
        SUB & proper\\
      \]\\
  MORPH & 
      \[{}
        STEM & ``Kurtulu\c{s}''\\
        CASE & nom\\
        AGR & 3sg\\
        POSS & none\\
      \]\\
   SEM & 
       \[{}
         CONCEPT & \#Kurtulu\c{s}-(Kurtulu\c{s})\\
         DEFINITE & +\\
       \]\\
   PHON & ``Kurtulu\c{s}''\\
\]
\avmEnd

\subsection{Pronouns}\label{sec-3:pronouns}

\begin{figure}[htb]
  \centerline{\psfig{figure=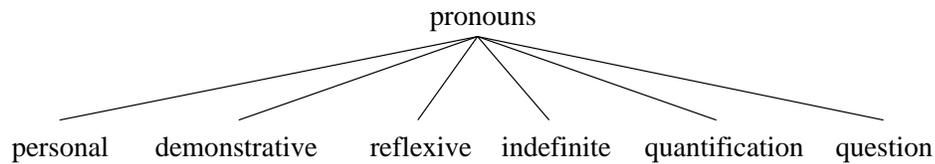}}
  \caption{Subcategories of pronouns.}
  \label{figure-3:pronouns}
\end{figure}

Pronouns are used in place of nouns in sentences, phrases, etc. (see
Ediskun \cite{Ediskun} and Ko\c{c} \cite{Koc})and subdivided into six
categories, as shown in Figure~\ref{figure-3:pronouns}.

Each pronoun also has the following semantic feature, which takes +
value for personal, reflexive and demonstrative pronouns, and  $-$
value for the other subcategories.

\avmBegin
\[{pronoun}
  SEM &
      \[{}
        DEFINITE & +/$-$ (default: $-$)\\
      \]\\
\]
\avmEnd

In the following sections we will give examples for each subcategory
of pronouns.

\subsubsection{Personal pronouns}

Personal pronouns are used to denote the speaker, the one spoken to,
and the one spoken of. 
This category consists of pronouns {\em ben} ({\em I}), {\em sen}
({\em you}), {\em o} ({\em he}/{\em she}/{\em it}), {\em biz}/{\em
  bizler} ({\em we}), {\em siz}/{\em sizler} ({\em you}), and
{\em onlar} ({\em they}). Personal pronouns may take all of the six
person suffixes as the agreement marker, but may not take a possessive
marker.

\subsubsection{Demonstrative pronouns}

Demonstrative pronouns denote the entities by showing them, but
without mentioning their actual names.
The following are examples of demonstrative pronouns: {\em bu} ({\em
 this}), {\em \c{s}u} ({\em that}), {\em bunlar} ({\em these}). Like
personal pronouns, this category of pronouns does not take a possessive
marker. {\em 3sg} and {\em 3pl} suffixes are the possible values for
the agreement marker. 
The following is the feature structure of {\em onlar} ({\em
  they}), as used in~(\ref{example:demonstratve-pronoun-1}):

\eenumsentence{
\item[]
  \label{example:demonstratve-pronoun-1}
  \shortex{5}
  {Bunu     & yapan{\i}n  & onlar & oldu\v{g}undan & eminim.}
  {{\tt this+ACC} & {\tt do+PART+GEN} & {\tt they}  & 
    {\tt be+PART+P3SG+ABL} & {\tt sure+PRES+1SG}}
  {`They, I am sure, did this.'}
  }

\avmBegin
\[{demonstrative pronoun}
  CAT & 
      \[{}
        MAJ & nominal\\
        MIN & pronoun\\
        SUB & demonstrative\\
      \]\\
  MORPH &
        \[{}
          STEM & ``o''\\
          CASE & nom\\
          AGR & 3pl\\
          POSS & none\\
        \]\\
  SEM &
      \[{}
        CONCEPT & \#o-(he/she/it)\\
        DEFINITE & +\\
      \]\\
  PHON & ``onlar''\\
\]
\avmEnd

\subsubsection{Reflexive pronouns}

Reflexive pronouns are words denoting the person or the thing on which
the action in the sentence has an effect.
This category consists of the pronouns {\em kendim} ({\em myself}), 
{\em kendin} ({\em yourself}), {\em kendi}/{\em kendisi} ({\em
  herself}/{\em himself}/{\em itself}), {\em kendimiz} ({\em
  ourselves}), {\em kendiniz} ({\em yourselves}), and {\em kendileri}
({\em themselves}).
The agreement and possessive markers take the same value, which is one
of the six person suffixes, e.g., it is {\em 3pl}
suffix for {\em kendileri}.
The same holds true for the indefinite and quantification pronouns.

\subsubsection{Indefinite pronouns}

Indefinite and quantification pronouns denote entities without
showing them explicitly. The difference between the two is that
quantification pronouns recall the existence of more than one entity.
All indefinite pronouns are inflected forms of the root word {\em
  biri} and {\em kimi}, e.g., {\em biri}/{\em birisi} ({\em someone}),
{\em birimiz} ({\em one of us}), {\em kiminiz} ({\em some of you}),
{\em kimileri} ({\em some of them}).\footnotemark

\footnotetext{
  Note that the inflected forms of {\em iki}, {\em \"{u}\c{c}},
  etc. (such as {\em ikiniz} ({\em two of you})) are classified as
  quantification pronouns. However, this is not productive.
  }

\subsubsection{Quantification pronouns}

There are two forms of quantification pronouns: {\em lexical} and {\em
  derived}.

\paragraph{Lexical}

The following are examples of quantification pronouns in lexical form:
{\em kimisi} ({\em some of them}), {\em kimimiz} ({\em some of us}),
{\em baz{\i}s{\i}} ({\em some of them}), {\em bir\c{c}o\v{g}u} ({\em
  most of them}), {\em \c{c}o\v{g}umuz} ({\em most of us}),
{\em herbirimiz} ({\em each of us}), {\em t\"{u}m\"{u}m\"{u}z}
({\em all of us}), {\em hepsi} ({\em all of them}).

Consider the feature structure of the quantification pronoun {\em
  bir\c{c}o\v{g}u} ({\em most of them}), as used in 
(\ref{example:quantification-pronoun-1}):

\eenumsentence{
\item[]
  \label{example:quantification-pronoun-1}
  \shortexnt{5}
  {K\"{o}t\"{u} & hava & ko\c{s}ullar{\i} & y\"{u}z\"{u}nden, &
    \"{o}\v{g}rencilerin}
  {{\tt bad} & {\tt weather} & {\tt condition+3PL+P3SG} & {\tt due to}
    & {\tt student+3PL+GEN}}
  \newline
  \shortex{2}
  {bir\c{c}o\v{g}u & gelemedi.}
  {{\tt most of them} & {\tt come+NEG+PAST+3SG}}
  {`Due to bad weather conditions, most of the students couldn't come.'}
  }

\avmBegin
\[{lexical quantification pronoun}
  CAT & 
      \[{}
        MAJ & nominal\\
        MIN & pronoun\\
        SUB & quantification\\
      \]\\
  MORPH &
        \[{}
          STEM & ``bir\c{c}ok''\\
          FORM & lexical\\
          CASE & nom\\
          AGR & 3pl\\
          POSS & 3pl\\
        \]\\
  SEM &
      \[{}
        CONCEPT & \#bir\c{c}ok-(most of \ldots)\\
      \]\\
  PHON & ``bir\c{c}o\v{g}u''\\    
\]
\avmEnd

\paragraph{Derived} The derivation to quantification pronouns is
possible only from quantification adjectives, e.g., {\em ikisi} ({\em
  two of them}), {\em \"{u}\c{c}\"{u}n\"{u}z} ({\em you three}). 
The derivation process is not productive: for example, *{\em ikileri}
is not a quantification pronoun. The derivation does not use a
suffix. 

Each derived quantification pronoun has the following additional
feature: 

\avmBegin
\[{derived quantification pronoun}
  MORPH &
        \[{}
          DERV-SUFFIX & {\em none}\\
        \]\\
\]
\avmEnd

\subsubsection{Question pronouns}

This category of pronouns look for entities by asking questions.
The following are examples of question pronouns: {\em kim}/{\em
  kimler} ({\em who}), {\em ne} ({\em what}), {\em hangisi} 
({\em which of them}), {\em hanginiz} ({\em which of you}). For the
agreement and possessive markers, there are two cases:

\begin{itemize}
\item they both take the same value, which is one of the six person
  suffixes, e.g., it is {\em 2pl} for {\em hanginiz},
\item agreement marker takes one of {\em 3sg} and {\em 3pl} suffixes,
  and possessive marker does not take any value, e.g., {\em kim}
  vs. {\em kimler}.
\end{itemize}

\subsection{Sentential Nominals}\label{sec-3:sentential}

\begin{figure}[htb]
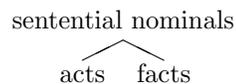

  \fnsCBegin
  \leaf{acts}
  \leaf{facts}
  \branch{2}{sentential nominals}
  \tree
  \fnsCEnd
  \caption{Subcategories of sentential nominals.}
  \label{figure-3:sentential}
\end{figure}

In this section we will describe sentential nominals, which head
sentences and function as nominals in syntax. As shown in
Figure~\ref{figure-3:sentential}, sentential nominals are divided into
two subcategories: {\em acts} and {\em facts}.

Each sentential nominal has the following additional features:

\avmBegin
\[{sentential}
  MORPH &
        \[{}
          DERV-SUFFIX & {\em derv-suffix}\\
        \]\\
  SYN &
      \[{}
        SUBCAT & {\em subcat}\\
      \]\\
  SEM &
      \[{}
        ROLES & {\em roles}\\
      \]\\
\]
\avmEnd

The DERV-SUFFIX feature takes one of the following: {\em -mAk}, {\em
  -mA}, {\em -yH\c{s}}, {\em -dHk}, and {\em
  -yAcAk}. Subcategorization information and thematic roles are also 
present in this feature structure.

\subsubsection{Acts}

The only subcategory of acts is {\em infinitives}, which is described
next.

\paragraph{Infinitives}

\begin{figure}[htb]
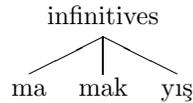

  \fnsCBegin
  \leaf{ma}
  \leaf{mak}
  \leaf{y{\i}\c{s}}
  \branch{3}{infinitives}
  \tree
  \fnsCEnd
  \caption{Subcategories of infinitives.}
  \label{figure-3:infinitives}
\end{figure}

Infinitives may be further divided into three subcategories, which
are derived from verbs with the suffixes {\em -mA}, {\em -mAk}, and {\em
  -yH\c{s}}, respectively, as shown in Figure~\ref{figure-3:infinitives}.
The derivation with {\em -mAk} is indefinite, i.e., the infinitive
does not take a possessive marker, while the other two may or may not
take this inflection.

The following are examples of infinitives: {\em gelmesi} ({\em his
  coming}), {\em geli\c{s}i} ({\em his coming}), {\em ko\c{s}mak}
({\em to run}), {\em \c{c}al{\i}\c{s}maktan} ({\em from working}).
As an example, consider the following feature structure for the
infinitive {\em bilmek} ({\em to know}), as used in
(\ref{example:infinitives-1}):\footnotemark

\footnotetext{
  Sentences~(\ref{example:infinitives-1-2}) and
  (\ref{example:infinitives-1-3}) are given to examplify the argument
  structure of the verb {\em bil-}.
  }

\eenumsentence{
  \label{example:infinitives-1}
\item
  \label{example:infinitives-1-1}
  \shortexnt{5}
  {Tolga'nin & d\"{u}n   & buraya   & neden & geldi\v{g}ini}
  {{\tt Tolga+GEN} & {\tt yesterday} & {\tt here+DAT} & {\tt why} &
    {\tt come+PART+P3SG+ACC}}
  \newline
  \shortex{4}
  {bilmek   & sana    & bir\c{s}ey    & kazand{\i}rmaz.}
  {{\tt to know} & {\tt you+DAT} & {\tt something} & {\tt
      gain+CAUS+NEG+ARST+3SG}}
  {`You will not gain anything by knowing why Tolga came here yesterday.'}
  \item
    \label{example:infinitives-1-2}
    \shortex{4}
    {Araba & kullanmay{\i} & biliyor & musun?}
    {{\tt car} & {\tt drive+INF+ACC} & {\tt know+PRES} & {\tt QUES+2SG}}
    {`Do you know how to drive?'}
  \item
    \label{example:infinitives-1-3}
    \shortex{5}
    {Bu & i\c{s}i & nas{\i}l & bitirece\u{g}imi & biliyorum.}
    {{\tt this} & {\tt job+ACC} & {\tt how} & {\tt end+PART+P1SG+ACC}
      & {\tt know+PRES+1SG}}
    {`I know how to end this thing.'}
  }

\avmBegin
\[{mak}
  CAT & 
      \[{}
        MAJ & nominal\\
        MIN & sentential\\
        SUB & act\\
        SSUB & infinitive\\
        SSSUB & mak
      \]\\
  MORPH &
      \[{}
        STEM & \@1\\
        FORM & derived\\
        DERV-SUFFIX & ``mak''\\
        CASE & nom\\
        AGR & 3sg\\
        POSS & none\\
      \]\\
  SYN &
      \[{}
        SUBCAT & \@2\\
      \]\\
  SEM &
      \[{}
        CONCEPT & f$_{mak}$(\@4)\\
        ROLES & \@3\\
      \]\\
  PHON & ``bilmek''\\
\]
\avmEnd

\avmBegin
\label{avm-3:bil}
\@1
\[{lexical predicative verb}
  CAT &
       \[{}
         MAJ & verb\\
         MIN & predicative\\
       \]\\
  MORPH &
        \[{}
          STEM & ``bil''\\
          FORM & lexical\\
          SENSE & pos\\
        \]\\
  SYN &
      \[{}
        SUBCAT &
               \@2
               \<
                 \@5
                 \[{}
                   SYN-ROLE & subject\\
                   OCCURRENCE & optional\\
                   CONSTRAINTS & \{$constraint_1$\}\\
                 \],\\
                 \@6
                 \[{}
                   SYN-ROLE & dir-obj\\
                   OCCURRENCE & optional\\
                   CONSTRAINTS & \{$constraint_2$,\\
                                   $constraint_3$,\\
                                   $constraint_4$,\\
                                   $constraint_5$\}\\
                 \]
               \>\\
      \]\\
  SEM &
      \[{}
        CONCEPT & \@4 \#bil-(to know)\\
        ROLES & 
              \@3
              \[{}
                AGENT & \@5\\
                THEME & \@6\\
              \]\\
       \]\\
  PHON & ``bil''\\
\]
\avmEnd

\avmBegin
\[{$constraint_1$}
  CAT &
      \[{}
        MAJ & nominal\\
        MIN & \{{\rm noun, pronoun}\}\\
      \]\\
  MORPH &
        \[{}
          CASE & nom\\
        \]\\
\]
\avmEnd

\avmBegin
\[{$constraint_2$}
  CAT &
      \[{}
        MAJ & nominal\\
        MIN & noun\\
      \]\\
  MORPH &
        \[{}
          CASE & \{{\rm acc, nom}\}\\
        \]\\
\]
\avmEnd

\avmBegin
\[{$constraint_3$}
  CAT &
      \[{}
        MAJ & nominal\\
        MIN & pronoun\\
      \]\\
  MORPH &
        \[{}
          CASE & acc\\
        \]\\
\]
\avmEnd

\avmBegin
\[{$constraint_4$}
  CAT &
      \[{}
        MAJ & nominal\\
        MIN & sentential\\
        SUB & act\\
        SSUB & infinitive\\
        SSSUB & ma\\
      \]\\
  MORPH &
        \[{}
          CASE & acc\\
        \]\\
\]
\avmEnd

\avmBegin
\[{$constraint_5$}
  CAT &
      \[{}
        MAJ & nominal\\
        MIN & sentential\\
        SUB & fact\\
        SSUB & participle\\
      \]\\
  MORPH &
        \[{}
          CASE & acc\\
          POSS & $\neg$none\\
        \]\\
\]
\avmEnd

\subsubsection{Facts}

The only subcategory of facts is {\em participles}, which is described
next.

\paragraph{Participles}

\begin{figure}[htb]
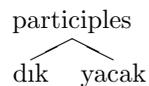

  \fnsCBegin
  \leaf{d{\i}k}
  \leaf{yacak}
  \branch{2}{participles}
  \tree
  \fnsCEnd
  \caption{Subcategories of participles.}
  \label{figure-3:participles}
\end{figure}

Participles may be further divided into two subcategories, which
are derived from verbs with the suffixes {\em -dHk} and {\em -yAcAk},
respectively, as shown in Figure~\ref{figure-3:participles}. 
Both subcategories take possessive markings.

The following are two examples of participles describing facts: 

\eenumsentence{
\item[]
  \begin{tabbing}
    -- geldi\v{g}i  \tabSpc \=`the fact that he came',\\
    -- gelece\v{g}ini       \> `the fact that he is going to come'.
  \end{tabbing}
}

Note that Section~\ref{sec-3:nouns} describes the participles
functioning as common nouns. As an example of participles acting as
sentential nominals and common nouns, consider
(\ref{example:participle-2-1}), which contains a sentence with two
parses. The first mentions about {\em the thing that Gamze
  brought}, and the participle, {\em getirdi\v{g}ini}, used as a
common noun. The latter is about {\em the event that Gamze
  brought something}, and the participle is used to represent this
fact. However, the participle in~(\ref{example:participle-2-2}) can
only be used to describe a fact.

\eenumsentence{
\item
  \label{example:participle-2-1}
  \shortex{4}
  {Gamze'nin & Ankara'dan & getirdi\v{g}ini & g\"{o}rd\"{u}m.}
  {{\tt Gamze+GEN} & {\tt Ankara+ABL} & 
    {\tt bring+PART+P3SG+ACC} & {\tt see+PAST+1SG} \vspGloss}
  {`I saw the thing that Gamze brought from Ankara.'\\
   `I saw that Gamze has brought it from Ankara.'}
\item
  \label{example:participle-2-2}
  \shortex{3}
  {Gamze'nin & geldi\v{g}ini      & g\"{o}rd\"{u}m.}
  {{\tt Gamze+GEN} & {\tt come+PART+P3SG+ACC} & {\tt see+PAST+1SG}}
  {`I saw that Gamze came.'}
  }

\section{Adjectivals}\label{sec-3:adjectivals}

\begin{figure} [hbt]
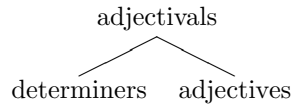

{\footnotesize
\begin{center}
\leaf{determiners}
\leaf{adjectives}
\branch{2}{adjectivals}
\tree
\end{center}
}
\caption{Subcategories of adjectivals.}
\label{figure-3:adjectivals}
\end{figure}

This section describes the representation of adjectivals in our
lexicon. Adjectivals are words that describe the properties of
nominals (specifically common nouns) in a number of ways, e.g.,
quality, quantity, etc. and specify them by differentiating from the
others. 
As shown in Figure~\ref{figure-3:adjectivals},
adjectivals consists of two subcategories: {\em determiners} and
{\em adjectives}. Figure~\ref{table-3:adjectival-categories} shows the
hierarchy under the adjectival category. 

\begin{figure}[htb]
\begin{center}
\begin{footnotesize}
\begin{tabular}{|l|l|l|l|}\hline
{\em maj} & {\em min} & {\em sub} & {\em ssub} \\ \hline \hline
adjectival & determiner  & article       &  \\ \hline
           &             & demonstrative &  \\ \hline
           &             & quantifier    &  \\ \hline         
           & adjective   & quantitative  & cardinal      \\ \hline
           &             &               & ordinal       \\ \hline
           &             &               & fraction      \\ \hline
           &             &               & distributive  \\ \hline
           &             & qualitative   &               \\ \hline
\end{tabular}
\end{footnotesize}
\end{center}
\caption{Lexicon categories of adjectivals.}
\label{table-3:adjectival-categories}
\end{figure}

Each adjectival has the following additional feature structure, which
contains syntactic and semantic information. SYN~$|$~MODIFIES 
specifies constraints on the modified of the adjectival including its
category, agreement marking and countability. 
For example, the cardinal adjective {\em bir} accepts only singular
countable common nouns, e.g., {\em bir kalem} vs. *{\em bir
  kalemler}.\footnotemark

\footnotetext{
  The category information states that adjectivals can only
  modify common nouns, which is not accurate, in fact. Consider the
  following example:
  
  \eenumsentence{
  \item
    \shortex{5}
    {Ankara'ya & {\em bu} & {\em gidi\c{s}imde} & onunla &
      konu\c{s}aca\v{g}{\i}m.}
    {{\tt Ankara+DAT} & {\tt this} & {\tt go+INF+P2SG+LOC} & 
      {\tt him+DAT} & {\tt talk+FUT+1SG}} 
    {`I will talk with him in my next visit to Ankara.'}
    }

  In this sentence, the demonstrative {\em bu} modifies a sentential
  nominal. However, we will omit these and simplify the pattern of
  modified constituent of adjectival phrases.
  }
 
\avmBegin
\[{adjectival}
  SYN &
      \[{}
        MODIFIES &
               \[{}
                 CAT &
                     \[{}
                       MAJ & nominal\\
                       MIN & noun\\
                       SUB & common\\
                     \]\\   
                 MORPH & 
                       \[{}
                         AGR & {\em agr}\\
                       \]\\
                 SEM &
                     \[{}
                       COUNTABLE & +/$-$\\
                     \]\\
               \]\\
       \]\\
  SEM &
      \[{}
        GRADABLE & +/$-$/semi (default: $-$)\\
        QUESTIONAL & +/$-$ (default: $-$)\\
      \]\\
\]
\avmEnd

There are two semantic features. The first one  describes the
gradability of the adjectival in consideration, e.g., the 
article {\em bir} is not gradable, whereas, the adjective {\em
  b\"{u}y\"{u}k} is. The other one is used to describe whether the
adjectival is in questional form, e.g., the following adjectivals are in
this form: {\em ka\c{c}} ({\em how many}), {\em ka\c{c}{\i}nc{\i}}
({\em in what order}), {\em nas{\i}l} ({\em how}), {\em hangi} ({\em
  which}).

In the next sections we will describe the subcategories of adjectivals
in detail.

\subsection{Determiners}\label{sec-3:determiners}

\begin{figure} [hbt]
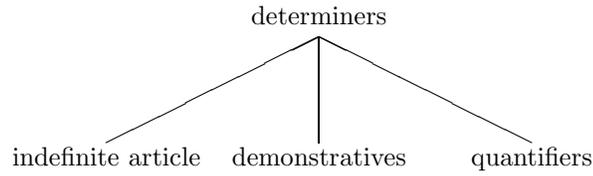

{\footnotesize
\begin{center}
\leaf{indefinite article}
\leaf{demonstratives}
\leaf{quantifiers}
\branch{3} {determiners}
\tree
\end{center}
}
\caption{Subcategories of determiners.}
\label{figure-3:determiners}
\end{figure}

Determiners are limiting adjectivals: they specify entities by showing
them explicitly or indefinitely.
As shown in Figure~\ref{figure-3:determiners}, determiners are 
subdivided into three categories: {\em indefinite article}, {\em
  demonstratives} and {\em quantifiers}, which are described in the
next sections.

\subsubsection{Indefinite Article}

The only article in Turkish is {\em bir}, as used in
(\ref{example:article}). As the name implies, this article, like
quantifiers, does not show entities explicitly.
The feature structure of this article is given below: 

\eenumsentence{
  \label{example:article}
\item
  \shortex{6}
  {Dilek & evinde & b\"{u}y\"{u}k & bir & bal{\i}k & besliyor.}
  {{\tt Dilek} & {\tt home+P3SG+LOC} & {\tt big} & {\tt a} & 
    {\tt fish} & {\tt look after+PROG+3SG}}
  {`Dilek is looking after a big fish at her home.'}
  }

\avmBegin
\[{article}
  CAT &
      \[{}
        MAJ & adjectival\\
        MIN & determiner\\
        SUB & article\\
      \]\\
  MORPH &
        \[{}
          STEM & ``bir''\\
        \]\\
  SYN &
      \[{}
        MODIFIES &
               \[{}
                 CAT &
                     \[{}
                       MAJ & nominal\\
                       MIN & noun\\
                       SUB & common\\
                     \]\\
                 MORPH &
                       \[{}
                         AGR & 3sg\\
                       \]\\
                 SEM &
                     \[{}
                       COUNTABLE & +\\
                    \]\\
               \]\\
       \]\\
  SEM &
      \[{}
        CONCEPT & \#bir-(a)\\
      \]\\
  PHON & ``bir''\\
\]
\avmEnd

\subsubsection{Demonstratives}

Demonstratives specify entities by showing them explicitly.
{\em Bu} ({\em this}), {\em \c{s}u} ({\em that}), {\em hangi}
({\em which}) and {\em di\v{g}er} ({\em other}) are examples of
demonstratives. As a specific example, consider {\em bu} ({\em
  this}), which is used in~(\ref{example:demonstratives}):

\eenumsentence{
\item[]
  \label{example:demonstratives}
  \shortex{6}
  {Buldu\v{g}um & bu  & \"{o}rnek & c\"{u}mle & \c{c}ok & sa\c{c}ma.}
  {{\tt devise+PART+P1SG} & {\tt this} & {\tt example} & 
    {\tt sentence} & {\tt very} & {\tt foolish+PRES+3SG}}
  {`This example sentence I devised is foolish.'}
  }

\avmBegin
\[{demonstrative}
  CAT &
      \[{}
        MAJ & adjectival\\
        MIN & determiner\\
        SUB & demonstrative\\
      \]\\
  MORPH &
        \[{}
          STEM & ``bu''\\
        \]\\
  SYN &
      \[{}
        MODIFIES &
               \[{}
                 CAT &
                     \[{}
                       MAJ & nominal\\
                       MIN & noun\\
                       SUB & common\\
                     \]\\
               \]\\
      \]\\
  SEM &
      \[{}
        CONCEPT & \#bu-(this)\\
      \]\\
  PHON & ``bu''\\
\]
\avmEnd

\subsubsection{Quantifiers}

{\em Her} ({\em each}), {\em baz{\i}}/{\em kimi} ({\em some}), 
{\em biraz} ({\em a little}), {\em bir\c{c}ok} ({\em many}), and
{\em b\"{u}t\"{u}n} ({\em all})  are examples of quantifiers. 
The following is the feature structure of {\em biraz}
({\em a little}), as used in the example sentence below:

\eenumsentence{
\item[]
  \shortex{6}
  {Timu\c{c}in, & bana & biraz & su & getirir & misin?}
  {{\tt Timu\c{c}in} & {\tt me+DAT} & {\tt a little} & {\tt water} &
    {\tt bring+ARST} & {\tt QUES+2SG}}
  {`Timu\c{c}in, could you bring me a little water?'}
  }

\avmBegin
\[{quantifier}
  CAT &
      \[{}
        MAJ & adjectival\\
        MIN & determiner\\
        SUB & quantifier\\
      \]\\
  MORPH &
        \[{}
          STEM & ``biraz''\\
        \]\\
  SYN &
      \[{}
        MODIFIES &
               \[{}
                 CAT &
                     \[{}
                       MAJ & nominal\\
                       MIN & noun\\
                       SUB & common\\
                     \]\\
                  MORPH &
                        \[{}
                          AGR & 3sg\\
                        \]\\
                  SEM &
                      \[{}
                        COUNTABLE & $-$\\
                      \]\\
               \]\\
      \]\\
  SEM &
      \[{}
        CONCEPT & \#biraz-(a little)\\
      \]\\
  PHON & ``biraz''\\
\]
\avmEnd

\subsection{Adjectives}\label{sec-3:adjectives}

Adjectives are used to describe the quantity and quality of
entities.
Figure~\ref{figure-3:adjectives} presents the subcategories of adjectives,
which consists of {\em quantitative} and {\em qualitative
  adjectives}. These subcategories are described in the following
sections.

\begin{figure} [hbt]
  \fnsCBegin
  \leaf{quantitative}
  \leaf{qualitative}
  \branch{2} {adjectives}
  \tree
  \fnsCEnd
  \caption{Subcategories of adjectives.}
  \label{figure-3:adjectives}
\end{figure}

\subsubsection{Quantitative Adjectives}

Quantitative adjectives describe the amount of the entities.
This category is further divided into four subcategories, as shown in
Figure~\ref{figure-3:quantitative-adjectives}.

\begin{figure} [hbt]
  \fnsCBegin
  \leaf{cardinals}
  \leaf{ordinals}
  \leaf{fractions}
  \leaf{distributives}
  \branch{4} {quantitative adjectives}
  \tree
  \fnsCEnd
  \caption{Subcategories of quantitative adjectives.}
  \label{figure-3:quantitative-adjectives}
\end{figure}

\paragraph{Cardinals}

Cardinals specify how many of entities are present. 
The following are examples of cardinals: {\em bir} ({\em one}),
{\em iki} ({\em two}), {\em y\"{u}zlerce} ({\em hundreds of}),
{\em ka\c{c}} ({\em how many}).

\paragraph{Ordinals}

Ordinals specify the rank of an entity.
The following are examples of ordinals: {\em birinci}/{\em ilk} 
({\em first}), {\em ikinci} ({\em second}), {\em sonuncu} ({\em
  last}), {\em ka\c{c}{\i}nc{\i}} ({\em in what order}).

\paragraph{Fractions}

This category of quantitative adjectives specify the relative size of
the parts of an entity.
The following are examples of fractions: {\em b\"{u}t\"{u}n}/{\em
  var}/{\em tam}/{\em t\"um} ({\em whole}), {\em yar{\i}m} ({\em
  half}), \c{c}eyrek ({\em one fourth}).
The following example demonstrates the fraction adjective usage of
{\em var}, which may not be evident at the first glance:

\eenumsentence{
\item[]
  \shortex{5}
  {Kazanmak & i\c{c}in & var & g\"uc\"umle & \c{c}al{\i}\c{s}t{\i}m.}
  {{\tt win+INF } & {\tt for} & {\tt whole} & {\tt power+P1SG+INS} &
    {\tt work+PAST+1SG}}
  {`I word so hard to win.'}
  }

\paragraph{Distributives}

{\em Birer} ({\em one each}) is an example of distributives, which
gives the size of each group that is obtained by dividing an entity
into parts equally. 


\subsubsection{Qualitative Adjectives}

Qualitative adjectives describe the properties of the entities.
There are two forms of qualitative adjectives: {\em lexical} and {\em
  derived}. In the next sections we will describe these forms in
detail with examples. 

Each qualitative adjective has the following additional feature,
which gives the subcategorization information:

\avmBegin
\[{qualitative adj}
  SYN &
      \[{}
        SUBCAT & {\em subcat} (default: none)\\
      \]\\
\]
\avmEnd

\paragraph{Lexical}

The feature structures of this form of adjectives are directly
accessible in the lexicon, i.e., no derivation process is involved.
The subcategorization information for this form consists of a list of
constraints on the only (if any) complement of the adjective (see the
example below). The following are examples of qualitative
adjectives in lexical form: {\em memnun} ({\em pleased}), {\em iyi}
({\em good}), {\em   zeki} ({\em clever}), {\em k\"{u}\c{c}\"{u}k}
({\em small}), {\em ayn{\i}} ({\em same}), {\em ertesi} ({\em
  next}), {\em   \c{c}ok} ({\em many}/{\em much}), {\em sar{\i}}
({\em yellow}), {\em nas{\i}l} ({\em how}).

Consider the feature structure for {\em memnun} ({\em pleased}), as
used in~(\ref{example:lexical-qualitative-adj-1}):\footnotemark

\footnotetext{
  Note that the argument structure of {\em memnun}, when used with the
  auxilary verb {\em ol-}, is different from that of the adjective
  usage. {\em Memnun ol-} ({\em be happy}/{\em satisfied}) is considered
  as a separate compound verb (see Section~\ref{sec-3:verbs}).

  \eenumsentence{
    \item[]
      \shortex{2}
      {{\em Buna} & mennun oldum.}
      {{\tt this+DAT} & {\tt be happy+PAST+1SG}}
      {`I am happy with it.'}
      }
    }

\eenumsentence{
  \label{example:lexical-qualitative-adj-1}
\item
  \label{example:lexical-qualitative-adj-1-1}
    \shortex{7}
    {Ondan   & memnun  & bir & tek    & \c{c}al{\i}\c{s}an & yok & burada.}
    {{\tt him+ABL} & {\tt pleased} & {\tt one} & {\tt unique} & 
      {\tt worker} & {\tt nonexistent+PRES+3SG} & {\tt here+LOC}}
    {`There is no one worker who is pleased from him.'}
\item
  \label{example:lexical-qualitative-adj-1-2}
  \shortexnt{6}
  {Olay{\i}n & bu   & \c{s}ekilde & geli\c{s}mesinden    & memnun}
  {{\tt event+GEN} & {\tt this} & {\tt way+LOC} & {\tt
      develop+INF+P3SG+ABL} & {\tt pleased}}
  \newline
  \shortex{1}
  {de\v{g}iliz.} 
  {{\tt NOT+PRES+1SG}}
  {`We are not pleased from the way it develops.'}
}

\avmBegin
\label{avm-3:memnun}
\[{lexical qualitative adj}
  CAT &
      \[{}
        MAJ & adjectival\\
        MIN & adjective\\
        SUB & qualitative\\
      \]\\
  MORPH &
        \[{}
          STEM & ``memnun''\\
          FORM & lexical\\
        \]\\
  SYN &
      \[{}
        MODIFIES &
               \[{}
                 CAT &
                     \[{}
                       MAJ & nominal\\
                       MIN & noun\\
                       SUB & common\\
                     \]\\
               \]\\
        SUBCAT & \{$constraint_1, constraint_2,$\\
                   $constraint_3, constraint_4$\}\\
      \]\\
  SEM &
      \[{}
        CONCEPT & \#memnun-(pleased)\\
        GRADABLE & +\\
      \]\\
  PHON & ``memnun''\\
\]
\avmEnd

\avmBegin
\[{$constraint_1$}
  CAT & 
      \[{}
        MAJ & nominal\\
        MIN & \{{\rm noun, pronoun}\}\\
      \]\\
  MORPH &
        \[{}
          CASE & abl\\
        \]\\
\]
\avmEnd

\avmBegin
\[{$constraint_2$}
  CAT & 
      \[{}
        MAJ & nominal\\
        MIN & sentential\\
        SUB & act\\
        SSUB & infinitive\\
        SSSUB & mak\\
      \]\\
  MORPH &
        \[{}
          CASE & abl\\
          POSS & none\\
        \]\\
\]
\avmEnd

\avmBegin
\[{$constraint_3$}
  CAT & 
      \[{}
        MAJ & nominal\\
        MIN & sentential\\
        SUB & act\\
        SSUB & infinitive\\
        SSSUB & ma\\
      \]\\
  MORPH &
        \[{}
          CASE & abl\\
          POSS & $\neg$none\\
        \]\\
\]
\avmEnd

\avmBegin
\[{$constraint_4$} 
  CAT & 
      \[{}
        MAJ & nominal\\
        MIN & sentential\\
        SUB & act\\
        SSUB & infinitive\\
        SSSUB & y{\i}\c{s}\\
      \]\\
  MORPH &
        \[{}
          CASE & abl\\
        \]\\
\]
\avmEnd

\paragraph{Derived}

Similar to other categories in derived form, producing feature
structures for derived qualitative adjectives requires computation of
features. 

Each derived qualitative adjective has the following additional
features:

\avmBegin
\[{derived qualitative adj}
  MORPH &
        \[{}
          DERV-SUFFIX & {\em derv-suffix}\\
          POSS & {\em poss} (default: none)\\
        \]\\
  SEM &
      \[{}
        ROLES & {\em roles} (default: none)\\
      \]\\
\]
\avmEnd

The derivation suffix may take one of the following values: 
{\em -lHk}, {\em -lH}, {\em -ki}, {\em -sHz}, {\em -sH},
{\em -yHcH}, {\em -yAn}, {\em -yAcAk}, {\em -dHk}, {\em -yAsH}, and
{\em none}. 
The feature MORPH~$|$~POSS is used  to hold the possessive marking of
adjective derived from verb, as in {\em bildi\v{g}im yemek} ({\em
  bil}+{\em dHk}+P1SG {\em yemek}, {\em dish that I know}). 
Possible values for this feature are the six person suffixes.
The last feature gives the semantic roles of the verb which is
involved in the derivation process.

During the derivation process, since predicting the gradability of the
qualitative adjective is difficult, its default value (i.e., it is
$-$) is used. 
For example, adjective {\em ak{\i}ls{\i}z} ({\em stupid}) is
gradable, while {\em kolsuz} ({\em without arm}) is not, that is
{\em \c{c}ok ak{\i}ls{\i}z} ({\em very stupid}) vs. *{\em \c{c}ok
  kolsuz}. 
However, the following prediction about the constraints on 
the complements of the derived qualitative adjectives is generally
correct: qualitative adjectives are generally modifiers of common
nouns and do not constrain the agreement and countability features of
the modified.

There are two possible derivations to qualitative adjectives:

\begin{itemize}
\item Nominal derivation:
  This derivation uses suffixes {\em -lHk}, {\em -lH}, {\em -ki}, {\em
    -sHz}, {\em -sH}, as in {\em ak{\i}ll{\i}} ({\em intelligent}),
  {\em evdeki} ({\em that is at home}), and {\em \c{c}ocuksu} ({\em
    childish}).

  Consider the feature structure for the derived qualitative adjective,
  {\em ak{\i}ll{\i}} ({\em intelligent}), as used in the following
  sentence: 

  \eenumsentence{
  \item[]
    \label{example:derived-qualitative-adjs-1}
    \shortex{5}
    {Ak{\i}ll{\i} & insanlar & b\"{o}yle & \c{s}eyler & yapmazlar.}
    {{\tt inteligent} & {\tt people} & {\tt such} & {\tt thing+3PL} &
      {\tt do+NEG+ARST+3PL}} 
    {`Intelligent people don't do this kind of things.'}
    }

  \avmBegin
  \label{avm-3:akilli}
  \[{derived qualitative adj}
    CAT &
        \[{}
          MAJ & adjectival\\
          MIN & adjective\\
          SUB & qualitative\\
        \]\\
    MORPH &
          \[{}
            STEM & \@1\\
            FORM & derived\\
            DERV-SUFFIX & ``l{\i}''\\
          \]\\
    SYN &
        \[{}
          SUBCAT & \@2 none\\
          MODIFIES &
                 \[{}
                   CAT &
                       \[{}
                         MAJ & nominal\\
                         MIN & noun\\
                         SUB & common\\
                       \]\\
                 \]\\
        \]\\
    SEM &
        \[{}
          CONCEPT & f$_{l{\i}}$(\@3)\\
          ROLES & none\\
        \]\\
    PHON & ``ak{\i}ll{\i}''\\
  \]
  \avmEnd

  \avmBegin
  \label{avm-3:akil}
  \@1
  \[{lexical common}
    CAT &
        \[{}
          MAJ & nominal\\
          MIN & noun\\
          SUB & common\\
        \]\\
    MORPH &
          \[{}
            FORM & lexical\\
            STEM & ``ak{\i}l''\\
          \]\\
    SYN &
        \[{}
          SUBCAT & \@2 none\\
        \]\\
    SEM &
        \[{}
          CONCEPT & \@3 \#ak{\i}l-(intelligence)\\
        \]\\
    PHON & ``ak{\i}l''
  \]
  \avmEnd

\item Verb derivation:
  This form of derivation uses the following suffixes: {\em -yHcH},
  {\em -yAn}, {\em -yAcAk}, {\em -dHk}, {\em -yAsH}, and {\em none}.
  Verbal form that take suffixes {\em -yAn}, {\em -yAcAk}, {\em -dHk},
  and {\em -yAsH} are, in fact, sentential heads of gapped sentences
  that dropped their subjects, objects, or oblique objects to modify
  these dropped constituents.
  These derivations produce two types of participles according to the
  grammatical function of the dropped constituent: {\em subject} and
  {\em object participles} (see Underhill \cite{Underhill-TG}).

  Derivations with {\em -yAn} and {\em -yAsH} may only produce subject
  participles, as illustrated
  in~(\ref{example:derived-qualitative-adj-2}):
  
  \eenumsentence{
    \label{example:derived-qualitative-adj-2}
  \item
    \shortex{5}
    {K\"{o}\c{s}ede & duran      & adam{\i} & tan{\i}yor & musun?}
    {{\tt corner+LOC} & {\tt stand+PART} & {\tt man+ACC} & 
      {\tt know+PROG} & {\tt QUES+2SG}}
    {`Do you know the man standing at the corner?'}
  \item
    \shortex{2}
    {\"{o}v\"{u}lesi & adam}
    {{\tt praise+PART} & {\tt man}}
    {`man deserving praise'}
  \item
    \shortex{3}
    {elleri & \"{o}p\"{u}lesi & kad{\i}n}
    {{\tt hand+3PL+3SG} & {\tt kiss+PART} & {\tt woman}}
    {`woman whose hands worth kissing'}
    }

  Derivations using {\em -yAcAk} may produce both types of participles,
  whereas the ones with {\em -dHk} may only produce object
  participles. Consider example sentences in 
  (\ref{example:derived-qualitative-adj-3}):
  
  \eenumsentence{
    \label{example:derived-qualitative-adj-3}
  \item
    \label{example:derived-qualitative-adj-3-1}  
    \shortex{5}
    {Paketi     & alacak    & \c{c}ocuk & hen\"{u}z & gelmedi.}
    {{\tt packet+ACC} & {\tt take+PART} & {\tt boy} & {\tt yet} & 
      {\tt come+NEG+PAST+3SG}}
    {`The boy who will take the packet has not come yet.'}
  \item
    \label{example:derived-qualitative-adj-3-2}  
    \shortexnt{6}
    {G\"{o}khan'{\i}n & okudu\v{g}u   & kitab{\i} & ben & daha \"{o}nce}
    {{\tt G\"{o}khan+GEN} & {\tt read+PART+3SG} & {\tt book+ACC} & 
      {\tt I} & {\tt before}}
    \newline
    \shortex{1}
    {okumu\c{s}tum.}
    {{\tt read+NARR+PAST+1SG}}
    {`I read the book that G\"{o}khan is reading before.'}
    }
  
  On the contrast, the qualitative adjectives derived form verbal with
  {\em -yHcH} are not heads of gapped sentences, e.g., {\em yaz{\i}c{\i}}
  ({\em printer}). Note that as used in {\em tan{\i}d{\i}k ki\c{s}i}
  ({\em known person}), {\em bildik biri} ({\em known person}), and
  {\em giyecek elbise} ({\em dress to wear}) not all participles
  derived using {\em -dHk} and {\em yAcAk} are heads of gapped
  sentences.\footnotemark\, These are the idiomatic usages of
  participles.

  \footnotetext{
    Although the form {\em predicative verb}+{\em dHk} is not
    productive (i.e., only some of the verbs may conform to it),
    its negated form is generally applicable to all predicative verbs,
    as used in the following:
  
    \eenumsentence{
    \item
      \shortex{6}
      {O & kitap & i\c{c}in & sormad{\i}k & d\"ukkan & b{\i}rakmad{\i}k.}
      {That & book & for & ask+NEG+PART & shop & leave+NEG+PAST+1PL}
      {`We didn't left any shop that we didn't ask that book.'}
    \item
      \shortex{3}
      {\c{C}almad{\i}k & kap{\i} & kalmad{\i}.}
      {knock+NEG+PART & door & exist+NEG+PAST+3SG}
      {`We consulted everyone.'}
      }
    }
  
  Derivation without using a suffix is also possible, e.g.,
  
  \eenumsentence{
  \item[]
    \begin{tabbing}
      -- bilir\tabSpc\=`that cannot come',\\
      -- okur yazar    \>`that reads and writes',\\
      -- donmu\c{s}    \>`that is frozen'.
    \end{tabbing}
    }
  
  Only object participles derived using {\em -dHk} and {\em -yAcAk}
  take possessive suffix, since the subject may be missing in the
  subordinate clause (see the following example).
  
  Consider the feature structure for {\em bilmedi\v{g}im} ({\em that I
    don't know}), as used
  in~(\ref{example:derived-qualitative-adj-4}):\footnotemark
  
  \footnotetext{
    The constraint structures of subcategorization information for the
    verb {\em bil-} are given on page~\pageref{avm-3:bil}.
    }
  
  \eenumsentence{
  \item[]
    \label{example:derived-qualitative-adj-4}
    \shortex{4}
    {Bilmedi\v{g}im    & yemekleri    & hi\c{c}bir zaman & yemem.}
    {{\tt know+NEG+PART+P1SG} & {\tt dish+3PL+ACC} & {\tt never} 
      & {\tt eat+NEG+ARST+1SG}}
    {`I never eat dishes that I don't know.'}
    }

  \avmBegin
  \[{derived qualitative adj}
    CAT &
        \[{}
          MAJ & adjectival\\
          MIN & adjective\\
          SUB & qualitative\\
        \]\\
    MORPH &
          \[{}
            STEM & \@1\\
            FORM & derived\\
            DERV-SUFFIX  & ``d{\i}k''\\
            POSS & 1sg\\
          \]\\
    SYN &
        \[{}
          SUBCAT & \@2\\
          MODIFIES &
                   \[{}
                     CAT &
                         \[{}
                           MAJ & nominal\\
                           MIN & noun\\
                           SUB & common\\
                         \]\\
                    \]\\
         \]\\
    SEM &
        \[{}
          CONCEPT & f$_{d{\i}k}$(\@4)\\
          ROLES & \@3\\
        \]\\
    PHON & ``bilmedi\v{g}im''\\
  \]
  \avmEnd
  
  \avmBegin
  \@1
  \[{lexical predicative verb}
    CAT &
         \[{}
           MAJ & verb\\
           MIN & predivative\\
         \]\\
    MORPH &
          \[{}
            STEM & ``bil''\\
            FORM & lexical\\
            SENSE & neg\\
          \]\\
    SYN &
        \[{}
          SUBCAT &
                 \@2
                 \<
                   \@5
                   \[{}
                     SYN-ROLE & subject\\
                     OCCURRENCE & optional\\
                     CONSTRAINTS & \{$constraint_1$\}\\
                   \],\\
                   \@6
                   \[{}
                     SYN-ROLE & dir-obj\\
                     OCCURRENCE & optional\\
                     CONSTRAINTS & \{$constraint_2$,\\
                                     $constraint_3$,\\
                                     $constraint_4$,\\
                                     $constraint_5$\}\\
                   \]
                 \>\\
        \]\\
    SEM &
        \[{}
          CONCEPT & \@4 \#bil-(to know something)\\
          ROLES & 
                \@3
                \[{}
                  AGENT & \@5\\
                  THEME & \@6\\
                \]\\
         \]\\
    PHON & ``bil''\\
  \]
  \avmEnd

\end{itemize}

\section{Adverbials}\label{sec-3:adverbials}

\begin{figure}[htb]
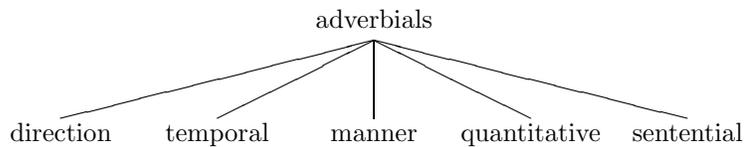

  \fnsCBegin
  \leaf{direction}
  \leaf{temporal}
  \leaf{manner}
  \leaf{quantitative}
  \leaf{sentential}
  \branch{5}{adverbials}
  \tree
  \fnsCEnd
  \caption{Subcategories of adverbials.}
  \label{figure-3:adverbials}
\end{figure}

This section describes the representation of adverbials in our
lexicon. These are words that modify or add to the meaning of
verbs (and verbal forms), adjectives, and adverbials in various ways,
e.g., direction, manner, temporality, etc. (see Ediskun \cite{Ediskun}).
As depicted in Figure~\ref{figure-3:adverbials}, adverbials are divided
into five subcategories, whose details are given in
Figure~\ref{table-3:adverbial-categories}.

\begin{figure}[htb]
\begin{center}
\begin{footnotesize}
\begin{tabular}{|l|l|l|l|}\hline
{\em maj} & {\em min} & {\em sub} & {\em ssub}  \\ \hline \hline
adverbial & direction    &               &          \\ \hline
          & temporal     & point-of-time &          \\ \hline
          &              & time-period   & fuzzy    \\ \hline
          &              &               & day-time \\ \hline
          &              &               & season   \\ \hline
          & manner       & qualitative   &          \\ \hline
          &              & repetition    &          \\ \hline
          & quantitative & approximation &          \\ \hline
          &              & comparative   &          \\ \hline
          &              & superlative   &          \\ \hline
          &              & excessiveness &          \\ \hline
          & sentential   &               &          \\ \hline
\end{tabular}
\end{footnotesize}
\end{center}
\caption{Lexicon categories of adverbials.}
\label{table-3:adverbial-categories}
\end{figure}

Each adverb has the following additional feature, which describes
whether the adverb in consideration is in questional form or not. For
instance, adverbs {\em neden} ({\em why}) and {\em nas{\i}l} ({\em how})
are in questional form.

\avmBegin
\[{adverbial}
  SEM &
      \[{}
        QUESTIONAL & +/$-$ (default: $-$)\\
      \]\\
\]
\avmEnd

\subsection{Direction Adverbs}

As the name implies, direction adverbs modify verbs and verbal forms
by specifying direction.
The following are examples of direction adverbs: {\em
  d{\i}\c{s}ar{\i}} ({\em out}), {\em beri} ({\em here}), {\em
  i\c{c}eri} ({\em in}), {\em geri} ({\em back}), {\em
  kar\c{s}{\i}} ({\em opposite}).

Consider the feature structure of the direction adverb {\em
  d{\i}\c{s}ar{\i}} ({\em out}), as used in~(\ref{example:dir-advs}):

\eenumsentence{
\item[]
  \label{example:dir-advs}
  \shortex{3}
  {D{\i}\c{s}ar{\i} & m{\i} & \c{c}{\i}k{\i}yorsun?}
  {{\tt out} & {\tt QUES} & {\tt get+PROG+1SG}}
  {`Are you getting out?'}
  }

\avmBegin
\[{direction adv}
  CAT &
      \[{}
        MAJ & adverbial\\
        MIN & direction\\
      \]\\
  MORPH &
        \[{}
          STEM & ``d{\i}\c{s}ar{\i}''\\
        \]\\
  SEM &
      \[{}
        CONCEPT & \#d{\i}\c{s}ar{\i}-(out)\\
      \]\\
  PHON & ``d{\i}\c{s}ar{\i}''\\
\]
\avmEnd

\subsection{Temporal Adverbs}

\begin{figure}[htb]
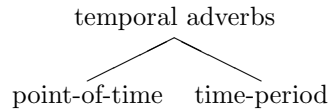

  \fnsCBegin
  \leaf{point-of-time}
  \leaf{time-period}
  \branch{2}{temporal adverbs}
  \tree
  \caption{Subcategories of temporal adverbs.}
  \label{figure-3:temporal-advs}
  \fnsCEnd
\end{figure}

Temporal adverbs specify the point of time and limit the period of
states, actions, and processes. As shown in
Figure~\ref{figure-3:temporal-advs} temporal adverbs comprise
{\em point-of-time} and  {\em time-period adverbs}.

\subsubsection{Point-of-Time Adverbs}

There are two forms of point-of-time adverbs: {\em lexical} and {\em
  derived}. The following two sections describe these with examples.

\paragraph{Lexical}

The following are point-of-time adverbs in lexical form:
{\em d\"{u}n} ({\em yesterday}), {\em bug\"{u}n} ({\em today}),
{\em \c{s}imdi} ({\em now}), {\em demin} ({\em a moment ago}), {\em 
  \"{o}nce} ({\em before}), {\em \"{o}nceden} ({\em beforehand}).

\paragraph{Derived}

This form of adverbs are derived from verbs using suffixes {\em -yHp}
and {\em -yHncA}. The derivation with {\em -yHp} produces adverbs that
state a subordinate action that happens simultaneously or in sequence
with the main action in the sentence. The other type of adverbs state
an action that happens in sequence with the main action. Consider the
following examples:

\eenumsentence{
  \item
    \shortex{6}
    {Bu & soruyu, & konuyu & anlay{\i}p & \c{c}\"ozmek & laz{\i}m.}
    {{\tt this} & {\tt question+ACC} & {\tt topic+ACC} & 
      {\tt understand+ADV} & {\tt solve+INF} & {\tt needed+PRES+3SG}}
    {`It is first needed to understand the topic and then to solve this
      question.'}
  \item
    \shortex{5}
    {Bu & ak\c{s}am & kitap & okuyup & dinlenecektim.\footnotemark}
    {{\tt this} & {\tt evening} & {\tt book} & {\tt read+ADV} 
      & {\tt rest+FUT+PAST+1SG}}
    {`This evening I was going to read a book and rest.'}
    }

\footnotetext{
  This example is due to Underhill \cite{Underhill-TG}.
  }
  
In the first sentence, the adverb, {\em anlay{\i}p}, states a
subordinate action that is performed before the main action. In the
latter one, however, the two actions happen simultaneously.

Each derived point-of-time adverb has the following additional
features, which give the derivation suffix, subcategorization
information and thematic roles of the verb involved in the derivation.

\avmBegin
\[{derived point-of-time adv}
  MORPH &
        \[{}
          DERV-SUFFIX & ``y{\i}nca''/``y{\i}p''\\
        \]\\
  SYN &
      \[{}
        SUBCAT & {\em subcat} \\
      \]\\
  SEM &
      \[{}
        ROLES & {\em roles}\\
      \]\\
\]
\avmEnd

 Consider the feature structure for {\em bitince} ({\em
  when it ends}), as used in~(\ref{example:derived-point-advs-1}):

\eenumsentence{
  \label{example:derived-point-advs-1}
\item
  \shortexnt{5}
  {Toplant{\i} & bitince, & konu\c{s}mac{\i}ya & bu   & konundaki}
  {{\tt meeting} & {\tt end+ADV} & {\tt speaker+DAT} & {\tt this} &
    {\tt subject+LOC+REL}}
  \newline
  \shortex{2}
  {fikrimi          & a\c{c}{\i}klad{\i}m.}
  {{\tt opinion+P1SG+ACC} & {\tt explain+PAST+1SG} \vspGloss}
  {`When the meeting ended, I explained my opinion about this subject
    to the speaker.'}
\item
  \shortexnt{4}
  {Odan{\i}      & toplaman         & bitince    & hemen      }
  {{\tt room+P2SG+ACC} & {\tt tidy up+INF+P2SG} & {\tt finish+ADV} &
    {\tt immediately}}
  \newline
  \shortex{2}
  {yatman{\i}             & istiyorum.}
  {{\tt go to bed+INF+P2SG+ACC} & {\tt want+PROG+1SG}}
  {`I want you to go to bed as soon as you finish tidying up your
    room.'}
}

\avmBegin
\[{derived point-of-time adv}
  CAT &
      \[{}
        MAJ & adverbial\\
        MIN & temporal\\
        SUB & point-of-time\\
      \]\\
  MORPH &
        \[{}
          STEM & \@1\\
          FORM & derived\\
          DERV-SUFFIX & ``y{\i}nca''\\
        \]\\
  SYN &
      \[{}
        SUBCAT & \@2\\
      \]\\
  SEM &
      \[{}
        CONCEPT & f$_{y{\i}nca}$(\@4)\\
        ROLES & \@3\\
      \]\\
  PHON & ``bitince''\\
\]
\avmEnd

\avmBegin
\@1
\[{lexical predicative verb}
  CAT &
      \[{}
        MAJ & verb\\
        MIN & predicative\\
      \]\\
  MORPH &
        \[{}
          STEM & ``bit''\\
          SENSE & pos\\
        \]\\
  SYN &
      \[{}
        SUBCAT &
               \@2
               \<
                 \@4
                 \[{}
                   SYN-ROLE & subject\\
                   OCCURRENCE & optional\\
                   CONSTRAINTS & \{$constraint_1$,\\
                                   $constraint_2$\}\\
                 \]\\
               \>
      \]\\
  SEM &
      \[{}
        CONCEPT & \@4 \#bit-(to end)\\
        ROLES &
              \@3
              \[{}
                AGENT & \@5\\
              \]\\
      \]\\
  PHON & ``bit''\\
\]
\avmEnd

\avmBegin
\[{$constraint_1$}
  CAT &
      \[{}
        MAJ & nominal\\
        MIN & \{{\rm noun, pronoun}\}\\
      \]\\
  MORPH &
        \[{}
          CASE & nom\\
        \]\\
\]
\avmEnd

\avmBegin
\[{$constraint_2$}
  CAT &
      \[{}
        MAJ & nominal\\
        MIN & sentential\\
        SUB & act\\
        SSUB & infinitive\\
        SSSUB & ma\\
      \]\\
  MORPH &
        \[{}
          CASE & nom\\
        \]\\
\]
\avmEnd

\subsubsection{Time-Period Adverbs}

\begin{figure}[htb]
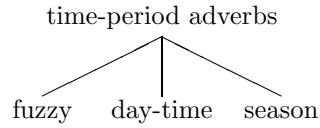

  \fnsCBegin
  \leaf{fuzzy}
  \leaf{day-time}
  \leaf{season}
  \branch{3}{time-period adverbs}
  \tree
  \fnsCEnd
  \caption{Subcategories of time-period adverbs.}
  \label{figure-3:time-period-advs}
\end{figure}

As Figure~\ref{figure-3:time-period-advs} shows, time-period adverbs are
subdivided into three categories: {\em fuzzy}, {\em day-time}, and
{\em season adverbs}.

\paragraph{Fuzzy}

There are two forms of fuzzy time-period adverbs: {\em lexical} and
  {\em derived}.
In the following two sections we will describe these forms with
examples. 

\subparagraph{Lexical} The following are examples of this form of
fuzzy time-period adverbs: {\em dakikalarca} ({\em for minutes}), {\em
  saatlerce}/{\em saatlerdir} ({\em for hours}).

\subparagraph{Derived} This form of adverbs are derived form verbs
using the suffixes, {\em -yAlH} and {\em -ken}, as in 

\eenumsentence{
\item[]
  \begin{tabbing}
    -- sen geleli/gideli\tabSpc\= `since the time you arrived/went',\\
    -- biz gelirken            \> `while we are coming'.
  \end{tabbing}
}

Each derived fuzzy time-period adverb also has the following
features. The derivation suffix is one of {\em -yAlH} and {\em
  -ken}. The other features give subcategorization information and 
semantic roles of the verb which are involved in the derivation
process.

\avmBegin
\[{derived fuzzy time-period adv}
  MORPH &
        \[{}
          DERV-SUFFIX & ``yal{\i}''/``ken''\\
        \]\\
  SYN &
      \[{}
        SUBCAT & {\em subcat} \\
      \]\\
  SEM &
      \[{}
        ROLES & {\em roles}\\
      \]\\
\]
\avmEnd

\paragraph{Day-time}

{\em Sabahleyin} ({\em in the morning}), {\em sabahlar{\i}} ({\em
  in the mornings}), {\em ak\c{s}amlar{\i}} ({\em in the
  evenings}), {\em g\"{u}nd\"{u}z} ({\em in the daytime}) and 
  {\em g\"{u}nd\"{u}zleyin} ({\em in the daytime}) are examples of
  day-time time-period adverbs.

\paragraph{Season}

{\em K{\i}\c{s}{\i}n} ({\em in the winter}) and {\em yaz{\i}n} ({\em
  in the summer}) are two examples of season time-period adverbs.

\subsection{Manner Adverbs}

\begin{figure} [hbt]
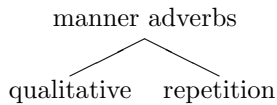

  \fnsCBegin
  \leaf{qualitative}
  \leaf{repetition}
  \branch{2}{manner adverbs}
  \tree
  \fnsCEnd
  \caption{Subcategories of manner adverbs.}
  \label{figure-3:manner-advs}
\end{figure}

Manner adverbs describe the way and how actions, processes, and states
develop. 
As depicted in Figure~\ref{figure-3:manner-advs} manner adverbs are divided
into two subcategories as {\em qualitative} and {\em repetition
  adverbs}, which are described next in detail.

\subsubsection{Qualitative Manner Adverbs}

There are two forms of qualitative manner adverbs: {\em lexical} and
{\em derived}. In the next sections, we will describe these forms in
detail with examples. 

\paragraph{Lexical}

The following are examples of qualitative manner adverbs in lexical
form: {\em birden} ({\em suddenly}), {\em \c{c}abuk} ({\em fast}),
{\em \c{c}abucak} ({\em fast}), {\em \c{s}\"{o}yle} ({\em like
  that}), {\em nas{\i}l} ({\em how}).

\paragraph{Derived} 

Each derived qualitative manner adverb has the following additional
features, in which derivation suffix, subcategorization information
and semantic roles are present. Derivation suffix feature may take one
of the following values: {\em -cAsHnA}, {\em -mAksHzHn},
{\em -mAdAn}, {\em -yAmAdAn}, {\em -yArAk}, and {\em -cA}.

\avmBegin
\[{derived qualitative adv}
  MORPH &
        \[{}
          DERV-SUFFIX & {\em derv-suffix}\\
        \]\\
  SYN &
      \[{}
        SUBCAT & {\em subcat} (default: none)\\
      \]\\
  SEM &
      \[{}
        ROLES & {\em roles} (default: none)\\
      \]\\
\]
\avmEnd

There are two types of derivations to this form of adverbs:

\begin{itemize}

\item Adjectival derivation:
  This derivation uses the suffix {\em -cA}, as in {\em
    ak{\i}ll{\i}ca} ({\em intelligently}), {\em h{\i}zl{\i}ca} ({\em
    fast}), and {\em aptalca} ({\em stupidly}). 
  Consider the feature structure for the qualitative adverb {\em
    ak{\i}ll{\i}ca} as used
  in~(\ref{example:derived-qualitative-adv}):\footnotemark
  
  \footnotetext{
    SYN~$|$~SUBCAT feature is co-indexed with that of {\em
      ak{\i}ll{\i}}, which is shown in the section on qualitative
    adjectives on page~\pageref{avm-3:akilli}.
    }
  
  \eenumsentence{
  \item[]
    \label{example:derived-qualitative-adv}
    \shortex{4}
    {Bug\"{u}n, & olduk\c{c}a & ak{\i}ll{\i}ca & davrand{\i}n.}
    {{\tt today} & {\tt rather} & {\tt intelligently} & {\tt behave+PAST+2SG}}
    {`You behaved rather intelligently today.'}
    }
  
  \avmBegin
  \[{derived qualitative adv}
    CAT &
        \[{}
          MAJ & adverbial\\
          MIN & manner\\
          SUB & qualitative\\
        \]\\
    MORPH &
          \[{}
            STEM & \fbox{``ak{\i}ll{\i}''}\\
            FORM & derived\\
            DERV-SUFFIX & ``ca''\\
          \]\\
    SYN &
        \[{}
          SUBCAT & none\\
        \]\\
    SEM &
        \[{}
          CONCEPT & f$_{ca}$(f$_{l{\i}}$(\#ak{\i}l-(intelligence)))\\
          ROLES & none\\
        \]\\
    PHON & ``ak{\i}ll{\i}ca''\\
  \]
  \avmEnd

\item Verb derivation:
  This derivation uses the suffixes {\em -cAsHnA}, {\em -mAksHzHn},
  {\em -mAdAn}, {\em -yAmAdAn}, and {\em -yArAk}, as in the examples
  below: 

  \eenumsentence{
    \item[]
      \begin{tabbing}
        -- ko\c{s}arcas{\i}na\tabSpc\=`as if running',\\
        -- g\"{o}rmeksizin          \>`without seeing',\\
        -- gelmeden                 \>`without coming',\\
        -- g\"{o}remeden            \>`without seeing',\\
        -- gelerek                  \>`by coming'.
      \end{tabbing}
    }

\end{itemize}

\subsubsection{Repetition Manner Adverbs}

As the name implies, this category of manner adverbs add repetition to
the semantics of the verb and verbal forms.
There are two forms of repetition manner adverbs, which are {\em lexical}
and {\em derived}

\paragraph{Lexical}
{\em Tekrar} ({\em again}), {\em gene} ({\em again}), {\em s{\i}k}
({\em frequently}) are some examples of this form.

\paragraph{Derived} The derivation to this form is only from verbs
and uses the suffix {\em -dHk\c{c}A} as in:

\eenumsentence{
\item[]
  \begin{tabbing}
    -- sen geldik\c{c}e\tabSpc\=`as you come',\\
    -- onlar konu\c{s}tuk\c{c}a   \>`as they talk'.\\
  \end{tabbing}
  }

Each derived repetition adverb has the following
additional feature structure, which has the derivation suffix,
subcategorization information and thematic roles.

\avmBegin
\[{derived repetition adv}
  MORPH &
        \[{}
          DERV-SUFFIX & ``d{\i}k\c{c}a''\\
        \]\\
  SYN &
      \[{}
        SUBCAT & {\em subcat} \\
      \]\\
  SEM &
      \[{}
        ROLES & {\em roles}\\
      \]\\
\]
\avmEnd

\subsection{Quantitative Adverbs}

\begin{figure} [htb]
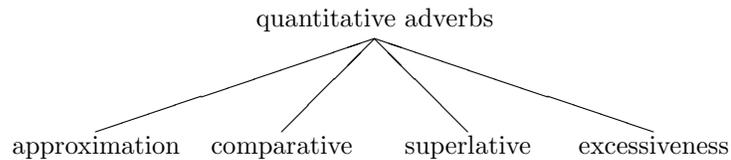

  \fnsCBegin
  \leaf{approximation}
  \leaf{comparative}
  \leaf{superlative}
  \leaf{excessiveness}
  \branch{4}{quantitative adverbs}
  \tree
  \fnsCEnd
  \caption{Subcategories of  quantitative adverbs.}
  \label{figure-3:quantitative-advs}
\end{figure}

Quantitative adverbs modify the semantics of adjectivals, adverbials,
and verbs in quantity.
As shown in Figure~\ref{figure-3:quantitative-advs}, quantitative adverbs
consist of four subcategories, for which many examples are given in
the next sections.

\subsubsection{Approximation}

{\em A\c{s}a\v{g}{\i} yukar{\i}} ({\em approximately}) and
{\em hemen hemen} ({\em approximately}) are two examples of adverbs
that are stating approximation.

\subsubsection{Comparative}

{\em Daha} ({\em more}) is the only member of this category.

\subsubsection{Superlative}

{\em En} ({\em most}) is the unique example of this category.

\subsubsection{Excessiveness}

The following are some examples of quantitative adverbs stating
excessiveness: {\em \c{c}ok} ({\em very}), {\em pek}/{\em gayet} ({\em
  very}), {\em fazla} ({\em too much}), {\em az}/{\em biraz} ({\em
  little}).

\subsection{Sentential Adverbs}

Sentential adverbs can only modify verbs and verbal forms.
The following are some examples of sentential adverbs: {\em evet}
({\em yes}), {\em yok} ({\em no}), {\em \"{o}yle} ({\em so}),
{\em elbette} ({\em certainly}), {\em ger\c{c}ekten} ({\em really}),
{\em daima} ({\em always}), {\em neden} ({\em why}).

\section{Verbs}\label{sec-3:verbs}

\begin{figure}[htb]
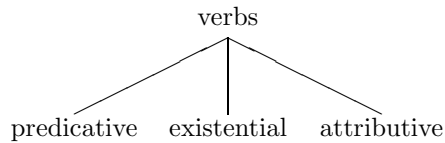

  \fnsCBegin
  \leaf{predicative}
  \leaf{existential}
  \leaf{attributive}
  \branch{3}{verbs}
  \tree
  \fnsCEnd
  \caption{Subcategories of verbs.}
  \label{figure-3:verbs}
\end{figure}

This section describes the representation of verbs in our lexicon with
an emphasis on argument structures and thematic roles. 
Verb is the head of sentence, hence it is the most important
constituent. It describes a state, action, or process
\cite{Yilmaz-Thesis}.
As shown in Figure~\ref{figure-3:verbs}, verbs are divided into three
categories as {\em predicative}, {\em existential}, and {\em
  attributive verbs}.

Each verb in the lexicon has the following additional features, which
represent morhosyntactic, syntactic, and semantic information. {\em
  none} is the default value for all of the features.

\avmBegin
\[{verb}
  MORPH &
        \[{}
          TAM2  & {\em tam2} \\
          COPULA & 1/2       \\
          AGR   & {\em agr}  \\
        \]\\
  SYN &
      \[{}
        SUBCAT & \<$role_1, \ldots, role_i, \ldots, role_n$\>
      \]\\
  SEM &
      \[{}
        ROLES &
              \[{}
                AGENT      \\
                EXPERIENCER\\
                PATIENT    \\
                THEME      \\
                RECIPIENT  \\
                CAUSER     \\
                ACCOMPANIER\\
                SOURCE     \\
                GOAL       \\
                LOCATION   \\
                INSTRUMENT \\
                BENEFICIARY\\
                VALUE-DES  \\
              \]\\
      \]\\
\]
\avmEnd

There are four morphosyntactic features introduced (see
Solak and Oflazer \cite{Solak-Oflazer}). 
The MORPH~$|$~ SENSE feature specifies whether the verb
states a {\em positive} or {\em negative} predicate, attribute, etc. 
There are four possible tenses for attributive and existential verbs,
which are also the possible second tenses for predicative verbs: {\em
  present}, {\em definite past}, {\em narrative past}, and {\em
  conditional forms}. This information is specified in MORPH~$|$~TAM2
feature. The feature MORPH~$|$~COPULA gives the usage of the suffix,
{\em -dHr}, which states probability or definiteness. The last one
represents the person suffix, whose possible values are {\em first},
{\em second}, and {\em third person singular}, and {\em plural
  persons}.  

The subcategorization information, which we will describe later in
detail, gives the valence of the verb for the {\em active}
voice.\footnotemark

\footnotetext{
  There are cases, in which the passive or causative
  voice of the verb gives a different sense than the active voice. In
  those cases, representation is configured accordingly, e.g.,

  \eenumsentence{
  \item
    \shortex{4}
    {Kemal'i & kap{\i}ya & kadar & ge\c{c}irdik.}
    {{\tt Kemal+ACC} & {\tt door+DAT} & {\tt up to} & {\tt see off+PAST+1PL}}
    {`We see Kemal off at the door.'}
  \item
    \shortex{3}
    {\.{I}brahim & Ay\c{s}e'ye  & vuruldu.}
    {{\tt Ibrahim} & {\tt Ay\c{s}e+DAT} & {\tt fall in love+PAST+3SG}}
    {`\.{I}brahim fell in love with Ay\c{s}e.'}
    }
  }

The feature SEM~$|$~ROLES describes the thematic roles of
the arguments of the verb. These role fillers are the following (see
Y{\i}lmaz \cite{Yilmaz-Thesis}):

\begin{itemize}
\item agent,
\item experiencer,
\item theme,
\item patient,
\item causer,
\item accompanier,
\item recipient,
\item goal,
\item source,
\item instrument,
\item value designator,
\item beneficiary,
\item location.
\end{itemize}

The subcategorization information is given as a list of elements, each
one describing an argument of the verb in question. Each such description
consists of three features:

\avmBegin
\[{$role_i$}
  SYN-ROLE & {\em syn-role}\\
  OCCURRENCE & obligatory/optional\\
  CONSTRAINTS & \{$constraint_1, \ldots, constraint_j, \ldots, constraint_m$\}\\
\]
\avmEnd

The feature SYN-ROLE gives the argument type, which is one of the
following:

\begin{itemize}
\item  subject,
\item  direct object,
\item  agentive object,
\item  oblique objects  (dative, ablative, locative)
\item  instrumental object,
\item  beneficiary object,
\item  value designator.
\end{itemize}

The second feature describes whether the occurrence of the argument is
{\em obligatory} or {\em optional}. The last feature gives a list of
constraints on the argument in consideration.

Elements in the subcategorization list are co-indexed with
corresponding thematic role fillers according to the verb in
consideration, i.e., there is a mapping from grammatical functions to
thematic roles. For example, direct object is generally co-indexed
with patient or theme.

The types of constraint structures are different for subject and
(direct, oblique, and agentive) objects, instrumental object, value
designator, and beneficiary object. Each structure will be described
in turn:

\begin{itemize}
\item Constraint structures for subject, direct, oblique and agentive
  objects: 
  The type of constraint structures for subject, direct, oblique, and
  agentive objects is given below. This feature structure gives
  constraints on the category, which is {\em nominal} in the most
  general case, a number of morphosyntactic and semantic properties of
  the argument.
  
  \avmBegin
  \[{$constraint_j$}
    CAT &
        \[{}
          MAJ & nominal\\
          MIN & {\em min}\\
          SUB & {\em sub}\\
          SSUB & {\em ssub}\\
          SSSUB & {\em sssub}\\    
        \]\\
    MORPH &
          \[{}
            STEM & {\em stem}\\
            CASE & {\em case}\\
            POSS & {\em poss}\\
            AGR & {\em agr}\\
          \]\\
    SEM &
        \[{}\]\\
  \]
  \avmEnd

  The subject never takes a case marking, i.e., it is in {\em
    nominative} case. There are cases that morphosyntactic features,
  other than the case, should be constrained, as well, as illustrated
  below: 

  \eenumsentence{
    \label{example:verbs-1}
  \item
    \label{example:verbs-1-1}
    \shortex{2}
    {\.{I}stanbul'u   & sel ald{\i}.}
    {{\tt Istanbul+ACC} & {\tt be flooded+PAST+3SG}}
    {`\.{I}stanbul is flooded.'}
  \item
    \label{example:verbs-1-2}
    \shortex{2}
    {\c{C}ocuk & kafay{\i} yedi.}
    {{\tt boy} & {\tt get mentally deranged+PAST+3SG}}
    {`The boy got mentally deranged.'}
    }

  In~(\ref{example:verbs-1-1}), in addition to the case, the stem and
  the possessive marker are required to be {\em sel} and {\em none},
  respectively. In the second sentence, however, the requirements are
  the following: the stem of the direct object is {\em kafa}; it has
  {\em accusative} case and {\em 3sg} agreement markers, and it is not
  possessive-marked. 
  
  Semantic constraints can also be posed in these structures. For
  example, the verb sense {\em kafay{\i} ye} ({\em to get metally
    deranged}) requires the subject to be human.

  The direct object may be in {\em nominative} or {\em accusative}
  cases, while oblique objects are in {\em dative}, {\em ablative}, and {\em
    locative} cases.

  The agentive object is in {\em ablative} case, and its stem is
  {\em taraf} with a suitable possessive marker. An example sentence
  is given in~(\ref{example:verbs-2}):

  \eenumsentence{
  \item[]
    \label{example:verbs-2}
    \shortex{4}
    {Sorun & bizim  & taraf{\i}m{\i}zdan & \c{c}\"{o}z\"{u}ld\"{u}.}
    {{\tt problem} & {\tt us+GEN} & {\tt by} & {\tt solve+PASS+PAST+3SG}}
    {`The problem is solved by us.'}
    }

\item Constraint structrures for instrumental object:
  The following are the constraint structures for the instrumental
  object. There are two possible types for this argument. The first type
  is  for nominals, which are {\em instrumental} case-marked. The second is
  for post-positional phrases, whose heads are the post-position {\em
    ile}:\footnotemark

  \footnotetext{
    There are two additional forms with the nominals {\em saye}+POSS+LOC
    and {\em arac{\i}l{\i}k}+POSS+INS ({\em arac{\i}l{\i}k}+POSS
    {\em ile}). These can be represented with the structures introduced
    above by imposing proper morphosyntactic constraints, e.g.,
    MORPH~$|$~STEM =  ``saye'', MORPH~$|$~CASE = loc, MORPH~$|$~AGR =
    3sg. But we will omit these forms.
    }

  \avmBegin
  \[{$constraint_j$}
    CAT &
        \[{}
          MAJ & nominal\\
          MIN & {\em min}\\
          SUB & {\em sub}\\
          SSUB & {\em ssub}\\
          SSSUB & {\em sssub}\\    
        \]\\
    MORPH &
          \[{}
            CASE & ins\\
          \]\\
    SEM &
        \[{}\]\\
  \]
  \avmEnd

  \avmBegin
  \[{$constraint_j$}
    CAT &
        \[{}
          MAJ & post-position\\
          MIN & ins-subcat\\
        \]\\
    MORPH &
          \[{}
            STEM & ``ile''\\
          \]\\
    SEM &
        \[{}\]\\
  \]
  \avmEnd

\item Constraint structures for value designator:
  There are two forms in a sentence to describe a value designator. The
  first form uses a nominal, which is {\em dative} case-marked. The
  second uses a post-positional phrase whose head is {\em i\c{c}in},
  as used in~(\ref{example:verbs-3}):\footnotemark

  \footnotetext{
    This example is due to Y{\i}lmaz \cite{Yilmaz-Thesis}.
    }

  \eenumsentence{
  \item[]
    \label{example:verbs-3}
    \shortex{6}
    {Oralarda & 10 & dolar & i\c{c}in & adam & \"old\"ur\"uler.}
    {{\tt there+LOC} & {\tt 10} & {\tt dolar} & {\tt for} & {\tt man} & 
      {\tt kill+ARST+3PL}}
    {They will kill you for 10 dollars there.}
    }

  Thus, the two feature structures that are introduced for instrumental
  object can be used for the value designator by replacing the values
  of case, stem, and the minor category features with {\em dative},
  {\em i\c{c}in}, and {\em nom-subcat} respectively.

\item Constraint structures for beneficiary object:
  The feature structure below is for the beneficiary object, which is a
  post-positional phrase whose head is the post-position, {\em i\c{c}in}:
  
  \avmBegin
  \[{$constraint_j$}
    CAT &
        \[{}
          MAJ & post-position\\
          MIN & nom-subcat\\
        \]\\
    MORPH &
          \[{}
            STEM & ``i\c{c}in''\\
          \]\\
    SEM &
        \[{}\]\\
  \]
  \avmEnd
  
  Furthermore, the oblique object case-marked as {\em dative} can be
  mapped to the beneficiary, as depicted in the following example:

  \eenumsentence{
  \item[]
    \shortex{7}
    {Annesi, & \c{c}ocu\v{g}a & uyumadan & \"{o}nce & kitap & okudu.}
    {{\tt mother+P1SG} & {\tt boy+DAT} & {\tt sleep+INF+ABL} & 
      {\tt before} & {\tt book} & {\tt read+PAST+3SG}}
    {`His mother read book for the boy before he slept.'}
    }
  
\end{itemize}

As mentioned above, the subcategorization information for verbs in
lexical form is given as a list, in which each element gives
constraints on an argument of the verb in consideration. Since
the members of other categories in lexical form, such as common nouns,
qualitative adjectives, and post-positions, cannot have more than one 
argument, just the constraint lists for one complement are given.

In the following sections we will describe the subcategories of verbs
in detail.

\subsection{Predicative Verbs}\label{sec-3:predicative-verbs}

Predicative verb category comprises the verbs that are not existential
or attributive.
There are two forms of predicative verbs, which are {\em lexical} and
{\em derived}.
These forms are described in the next sections.

Each predicative verb has the following additional morphosyntactic
features:

\avmBegin
\[{predicative verb}
  MORPH &
        \[{}
          SENSE      & pos/neg\\
          TAM1       & {\em tam1} & (default: none)\\
          COMP       & {\em comp} & (default: none)\\
          PASSIVE    & $+$/$-$ & (default: $-$)\\
          RECIPROCAL & $+$/$-$ & (default: $-$)\\
          REFLEXIVE  & $+$/$-$ & (default: $-$)\\
          CAUSATIVE  & {\em n}   & (default: 0)\\
        \]\\
\]
\avmEnd

The first tense-aspect-mood marker is specified in MORPH~$|$~TAM1
feature, for which there are ten possible values: {\em present}, {\em
  definite past}, {\em narrative past}, {\em future}, {\em aorist},
{\em progressive}, {\em conditional}, {\em optative}, {\em necessitative},
and {\em imperative}. 
If the verb is a compound one, the compounding suffix is given in
MORPH~$|$~COMP feature, whose value is one of {\em -yAbil}, {\em
  -yHver}, {\em -yAdur}, {\em -yAkoy}, {\em -yAkal}, and {\em
  -yAyaz}. 
The last four features represent the voice of the verb. The value
{\em n} represents a positive integer number, which denotes the level
of causation (see Solak and Oflazer \cite{Solak-Oflazer}).

\subsubsection{Lexical}

This form of predicative verbs are present in the lexicon as lexical
entries mainly consisting of subcategorization information and thematic
roles. The following are example predicative verbs in lexical form: 

\eenumsentence{
\item[]
  \begin{tabbing}
    -- ye-     \tabSpc\=`eat',\\
    -- i\c{c}-        \> `drink',\\
    -- g\"{o}r-       \> `see',\\ 
    -- hediye et-     \> `give present',\\
    -- kafay{\i} ye-  \> `get mentally deranged',\\
    -- r\"{u}\c{s}vet ye- \> `receive bribe'.
  \end{tabbing}
}

Some of the predicative verbs consist of more than one word,
e.g., {\em kafay{\i} ye-} ({\em get mentally deranged}), {\em rezil
  et-} ({\em disgrace}), {\em rezil ol-} ({\em be disgraced}),
{\em kavga et-} ({\em quarrel}), some of which are constructed
with the auxiliary verbs {\em et-} and {\em ol-}. The verbs whose first
constituents are not nominals are taken as separate compound verbs,
whereas there are two cases for the ones whose first constituents are
nominals. In the first case, such constituents are not subject to
inflections as in~(\ref{example:predicative-lexical-1-1}):

\eenumsentence{
\item
  \label{example:predicative-lexical-1-1}
  \shortexnt{4}
  {*Biz & yine de & hediyemizi      & ederiz.}
  {{\tt we} & {\tt anyway} & {\tt present+1PL+ACC} & {\tt do+ARST+1PL}}
\item
  \label{example:predicative-lexical-1-2}
  \shortex{4}
  {Biz & gerekirse & kavgam{\i}z{\i} & ederiz.}
  {{\tt we} & {\tt if needed} & {\tt fight+1PL+ACC} & 
    {\tt do+ARST+1PL}}
  {`If needed, we will fight.'}
}

This type of verbs are taken separately as compound verbs. In the
latter case, as in~(\ref{example:predicative-lexical-1-2}), such
constituents are subject to inflection, which are taken as a different
sense of the main verb, and the first constituent is given as an object
in the argument structure. 
For example, {\em kavga et-} ({\em quarrel}) is represented as a
sense of {\em et-}, and {\em kavga} ({\em quarrel}) is the direct
object of this sense.

We will give feature structures for four senses of the verb, {\em ye-},
which are the following: 

\begin{enumerate}
\item eat something, 
\item eat from something, 
\item get mentally deranged, 
\item be unfair. 
\end{enumerate}

The following is the feature structure for the first sense, {\em eat
  something}, as used in~(\ref{example:predicative-lexical-2}):

\eenumsentence{
\item[]
  \label{example:predicative-lexical-2}
  \shortex{4}
  {Adam & \c{c}atalla & pastay{\i} & yedi.}
  {{\tt man} & {\tt fork+INS} & {\tt pastry+ACC} & {\tt eat+PAST+3SG}}
  {`The man ate the pastry with fork.'}
  }

\avmBegin
\[{lexical predicative verb}
  CAT &
      \[{}
        MAJ & verb\\
        MIN & predicative\\
      \]\\
  MORPH &
        \[{}
          STEM & ``ye''\\
          FORM & lexical\\
          SENSE & pos\\
          TAM1 & past\\
          AGR & 3sg\\
        \]\\
  SYN &
      \[{}
        SUBCAT & \<
                   \@1
                   \[{}
                     SYN-ROLE & subject\\
                     OCCURRENCE & optional\\
                     CONSTRAINTS & \{$constaint_1$\}\\
                   \], \\
                   \@2
                   \[{}
                     SYN-ROLE & dir-obj\\
                     OCCURRENCE & optional\\
                     CONSTRAINTS & \{$constaint_2$,\\
                                     $constaint_3$\}\\
                   \], \\
                   \@3
                   \[{}
                     SYN-ROLE & inst-obj\\
                     OCCURRENCE & optional\\
                     CONSTRAINTS & \{$constaint_4,$\\
                                     $constaint_5$\}\\
                   \]\\
                \>
      \]\\
  SEM &
      \[{}
        CONCEPT & \#ye-(to eat something)\\
        ROLES &
              \[{}
                AGENT & \@1\\
                THEME & \@2\\
                INSTRUMENT & \@3
              \]\\
      \]\\
  PHON & ``yedi''\\
\]
\avmEnd

\avmBegin
\[{$constraint_1$}
  CAT & 
      \[{}
        MAJ & nominal\\
        MIN & \{{\rm noun, pronoun}\}\\
      \]\\
  MORPH &
        \[{}
          CASE & nom\\
        \]\\
  SEM &
      \[{}
        ANIMATE +\\
      \]\\ 
\]
\avmEnd

\avmBegin
\[{$constraint_2$}
  CAT & 
      \[{}
        MAJ & nominal\\
        MIN & noun\\
      \]\\
  MORPH &
        \[{}
          CASE & \{{\rm acc, nom}\}\\
        \]\\
  SEM &
      \[{}
        EDIBLE +\\
      \]\\ 
\]
\avmEnd

\avmBegin
\[{$constraint_3$}
  CAT & 
      \[{}
        MAJ & nominal\\
        MIN & pronoun\\
      \]\\
  MORPH &
        \[{}
          CASE & acc\\
        \]\\
\]
\avmEnd

\avmBegin
\[{$constraint_4$}
  CAT & 
      \[{}
        MAJ & nominal\\
        MIN & \{{\rm noun, pronoun}\}\\
      \]\\
  MORPH &
        \[{}
          CASE & ins\\
        \]\\
  SEM &
      \[{}
        INSTRUMENT & +\\
      \]\\
\]
\avmEnd

\avmBegin
\[{$constraint_5$}
  HEAD &
       \[{}
         CAT & 
             \[{}
               MAJ & post-position\\
               MIN & ins-subcat\\ 
             \]\\
         MORPH &
               \[{}
                 STEM & ``ile''\\
               \]\\
       \]\\
  SEM &
      \[{}
        INSTRUMENT & +\\
      \]\\
\]
\avmEnd

The following is the feature structure for the second sense, {\em eat
  from something}, as used in
(\ref{example:predicative-lexical-3}):\footnotemark 

\footnotetext{
The feature structure for subject and instrumental object are the same
with those of  previous example.
}

\eenumsentence{ 
\item[]
  \label{example:predicative-lexical-3}
  \shortex{4}
  {Adam & \c{c}atalla & pastadan   & yedi.}
  {{\tt man} & {\tt fork+INS} & {\tt pastry+ABL} & {\tt eat+PAST+3SG}}
  {`The man ate from the pastry with fork.'}
}

The difference between the first and the second senses is that the
patient, {\em pasta} ({\em pastry}), is the direct object in the
former one, whereas, it is the oblique object in ablative case in the
latter. Note that the second sense does not subcategorize for a direct
object. 

\avmBegin
\[{lexical predicative verb}
  CAT &
      \[{}
        MAJ & verb\\
        MIN & predicative\\
      \]\\
  MORPH &
        \[{}
          STEM & ``ye''\\
          FORM & lexical\\
          SENSE & pos\\
          TAM1 & past\\
          AGR & 3sg\\
        \]\\
  SYN &
      \[{}
        SUBCAT & \<
                   \@1
                   \[{}
                     SYN-ROLE & subject\\
                     OCCURRENCE & optional\\
                     CONSTRAINTS & \{$constaint_1$\}\\
                   \], \\
                   \@2
                   \[{}
                     SYN-ROLE & obl-abl\\
                     OCCURRENCE & optional\\
                     CONSTRAINTS & \{$constaint_2$\}\\
                   \], \\
                   \@3
                   \[{}
                     SYN-ROLE & inst-obj\\
                     OCCURRENCE & optional\\
                     CONSTRAINTS & \{$constaint_3$,\\
                                     $constaint_4$\}\\
                   \]\\
                \>
      \]\\
  SEM &
      \[{}
        CONCEPT & \#ye-(to eat from something)\\
        ROLES &
              \[{}
                AGENT & \@1\\
                THEME & \@2\\
                INSTRUMENT & \@3\\
              \]\\
      \]\\
  PHON & ``yedi''\\
\]
\avmEnd

\avmBegin
\[{$constraint_2$}
  CAT & 
      \[{}
        MAJ & nominal\\
        MIN & \{{\rm noun, pronoun}\}\\
      \]\\
  MORPH &
        \[{}
          CASE & abl\\
        \]\\
  SEM &
      \[{}
        EDIBLE & +\\
      \]\\ 
\]
\avmEnd

The following is the feature structure for the third sense of {\em
  ye-}, {\em get mentally deranged}, as shown
in~(\ref{example:predicative-lexical-4}):

\eenumsentence{
\item[]
  \label{example:predicative-lexical-4}
  \shortexnt{6}
  {C\"{u}neyt, & okulda   & \c{c}ok  & \c{c}al{\i}\c{s}maktan}
  {{\tt C\"uneyt} & {\tt school+LOC} & {\tt too much} & {\tt working+ABL}}
  \newline
  \shortex{1}
  {kafay{\i} yedi.}
  {{\tt get mentally deranged+PAST+3SG}}
  {`C\"uneyt got mentally deranged from too much working at the school.'}
  }

Note that the direct object has to be {\em kafay{\i}}, and it is
not a semantic role filler.

\avmBegin
\[{lexical predicative verb}
  CAT &
      \[{}
        MAJ & verb\\
        MIN & predicative\\
      \]\\
  MORPH &
        \[{}
          STEM & ``ye''\\
          FORM & lexical\\
          SENSE & pos\\
          TAM1 & past\\
          AGR & 3sg\\
        \]\\
  SYN &
      \[{}
        SUBCAT & \<
                   \@1
                   \[{}
                     SYN-ROLE & subject\\
                     OCCURRENCE & optional\\
                     CONSTRAINTS & \{$constaint_1$\}\\
                   \], \\
                   \[{}
                     SYN-ROLE & dir-obj\\
                     OCCURRENCE & obligatory\\
                     CONSTRAINTS & \{$constaint_2$\}\\
                   \]\\
                \>
      \]\\
  SEM &
      \[{}
        CONCEPT & \#ye-(to get mentally deranged)\\
        ROLES &
              \[{}
                EXPERIENCER & \@1\\
              \]\\
      \]\\
  PHON & ``yedi''\\
\]
\avmEnd

\avmBegin
\[{$constraint_1$}
  CAT & 
      \[{}
        MAJ & nominal\\
        MIN & \{{\rm noun, pronoun}\}\\
      \]\\
  MORPH &
        \[{}
          CASE & nom\\
        \]\\
  SEM &
      \[{}
        HUMAN & +\\
      \]\\ 
\]
\avmEnd

\avmBegin
\[{$constraint_2$}
  CAT & 
      \[{}
        MAJ & nominal\\
        MIN & noun\\
        SUB & common\\
      \]\\
  MORPH & 
        \[{}
          STEM & ``kafa''\\
          CASE & acc\\
          AGR & 3sg\\
          POSS & none\\
        \]\\
\]
\avmEnd

The feature structure for the fourth sense of {\em ye-} is given below,
in which the direct object, {\em hak}, is optionally {\em accusative}
case-marked, as below:

\eenumsentence{
  \label{example:predicative-lexical-5}
\item
  \shortex{3}
  {O\v{g}uz & hep    & hak yiyor.}
  {{\tt O\v{g}uz} & {\tt always} & {\tt be unfair+PROG+3SG}}
  {`O\v{g}uz is always unfair.'}
\item
  \shortex{4}
  {O\v{g}uz & ba\c{s}kalar{\i}n{\i}n & da  & haklar{\i}n{\i} yedi.}
  {{\tt O\v{g}uz} & {\tt others+GEN} & {\tt too} & {\tt be unfair+PAST+3SG}}
  {`O\v{g}uz was unfair to the others, too.'}
  }

\avmBegin
\[{lexical predicative verb}
  CAT &
      \[{}
        MAJ & verb\\
        MIN & predicative\\
      \]\\
  MORPH &
        \[{}
          STEM & ``ye''\\
          FORM & lexical\\
          SENSE & pos\\
          TAM1 & past\\
          AGR & 3sg\\
        \]\\
  SYN &
      \[{}
        SUBCAT & \<
                   \@1
                   \[{}
                     SYN-ROLE & subject\\
                     OCCURRENCE & optional\\
                     CONSTRAINTS & \{$constaint_1$\}\\
                   \], \\
                   \@2
                   \[{}
                     SYN-ROLE & dir-obj\\
                     OCCURRENCE & obligatory\\
                     CONSTRAINTS & \{$constaint_2$\}\\
                   \]\\
                \>
      \]\\
  SEM &
      \[{}
        CONCEPT & \#ye-(to be unfair)\\
        ROLES &
              \[{}
                AGENT & \@1\\
                THEME & \@2\\
              \]\\
      \]\\
  PHON & ``yedi''\\
\]
\avmEnd

\avmBegin
\[{$constraint_1$}
  CAT & 
      \[{}
        MAJ & nominal\\
        MIN & \{{\rm noun, pronoun}\}\\
      \]\\
  MORPH &
        \[{}
          CASE & nom\\
        \]\\
\]
\avmEnd

\avmBegin
\[{$constraint_2$}
  CAT & 
      \[{}
        MAJ & nominal\\
        MIN & noun\\
        SUB & common\\
      \]\\
  MORPH & 
        \[{}
          STEM & ``hak''\\
          CASE & \{{\rm acc, nom}\}\\
        \]\\
\]
\avmEnd

\subsubsection{Derived}

This form of verbs are derived from nominals and adjectivals using the
suffixes {\em -lAn} and {\em -lA\c{s}}. Each derived predicative verb has
the following additional feature, which gives the derivation suffix.

\avmBegin
\[{derived verbal}
  MORPH &
        \[{}
          DERV-SUFFIX & ``lan''/``la\c{s}''\\
        \]\\
\]
\avmEnd

There are two types of derivations to predicative verbs:

\begin{itemize}
\item Nominal derivation:
This derivation uses the suffixes {\em -lAn} and {\em -lA\c{s}}. 
The following are some examples of predicative verbs derived form
nominals: 

\eenumsentence{
\item[]
  \begin{tabbing}
    -- ta\c{s}la\c{s}- \tabSpc\=`turn into stone',\\
    -- a\v{g}a\c{c}land{\i}r- \>`plant trees in an area',\\
    -- sinirlen-              \>`get nervous'.
  \end{tabbing}
}

Consider the feature structure for {\em sinirlen-}, as used in
(\ref{example:predicative-derived-nom-1}):

\eenumsentence{
\item[]
  \label{example:predicative-derived-nom-1}
  \shortex{5}
  {Tembellik & etmen       & beni   & \c{c}ok & sinirlendiriyor!}
  {{\tt laziness} & {\tt do+INF+P2SG} & {\tt me+ACC} & {\tt very} &
    {\tt make angry+PROG+3SG}} 
  {`Your laziness is making me very angry!'}
}

\avmBegin
\[{derived predicative verb}
  CAT &
      \[{}
        MAJ & verb\\
        MIN & predicative\\
      \]\\
  MORPH &
        \[{}
          STEM  & \@1\\
          FORM  & derived\\
          DERV-SUFFIX & ``lan''\\
          SENSE & pos\\
          TAM1  & prog1\\
          CAUSATIVE & 1\\
        \]\\
  SYN &
      \[{}
        SUBCAT & \@2 none\\
      \]\\
  SEM &
      \[{}
        CONCEPT & f$_{lan}$(\@3)\\
        ROLES & none\\
      \]\\
  PHON & ``sinirlendiriyor''\\
\]
\avmEnd

\avmBegin
\@1
\[{lexical common}
  CAT &
      \[{}
        MAJ & nominal\\
        MIN & noun\\
        SUB & common\\
      \]\\
  MORPH &
        \[{}
          STEM & ``sinir''\\
          FORM & lexical\\
        \]\\
  SYN &
      \[{}
        SUBCAT & \@2 none\\
      \]\\
  SEM &
      \[{}
        CONCEPT & \@3 \#sinir-(anger)\\
      \]\\
  PHON & ``sinir''\\
\]
\avmEnd

\item Adjectival derivation:
This derivation uses the same suffixes. The following are some
examples of predicative verbs derived from adjectivals: {\em iyile\c{s}-}
({\em recover from illness}), {\em uzakla\c{s}-}, ({\em go away
  from}), {\em yaralan-} ({\em be hurted}). 

\end{itemize}

\subsection{Existential Verbs}

This category of verbs consists of only {\em var} ({\em existent})
and {\em yok} ({\em nonexistent}), which state existence and
non-existence in sentences, respectively. Two example sentences are
given in (\ref{example:existential}):

\eenumsentence{
  \label{example:existential}
\item
  \shortex{5}
  {Masamda        & ka\v{g}{\i}t & ve  & kalem  & var.}
  {{\tt table+P1SG+LOC} & {\tt paper} & {\tt and} & {\tt pencil} &
    {\tt existent+PRES+3SG}}
  {`There are paper and pencil on my table.'}
\item
  \shortex{5}
  {Bug\"{u}n & yapacak & fazla & i\c{s}im  & yok.}
  {{\tt today} & {\tt do+PART} & {\tt much} & {\tt work+P1SG} & 
    {\tt nonexistent+PRES+3SG}}
  {`I don't have much work to do today.'}
  }

\subsection{Attributive Verbs}

Attributive verbs state properties of entities.
This category consists of verbs in {\em lexical} and {\em derived}
forms, which are described in the next sections. 


\subsubsection{Lexical}

The only attributive verb that is in lexical form is {\em
  de\v{g}il} ({\em not}). This verb makes the sentences negative whose
heads, otherwise, are existential or derived attributive verbs, as
shown in (\ref{example:attributive-lexical}):

\eenumsentence{
  \label{example:attributive-lexical}
\item
  \shortex{3}
  {Onun & bisikleti    & k{\i}rm{\i}z{\i}yd{\i}.}
  {{\tt his} & {\tt bicycle+P3SG} & {\tt red+PAST+3SG}}
  {`His bicycle was red.'}
\item
  \shortex{4}
  {Onun & bisikleti    & k{\i}rm{\i}z{\i} & de\v{g}ildi.}
  {{\tt his} & {\tt bicycle+P3SG} & {\tt red} & {\tt NOT+PAST+3SG}}
  {`His bicycle was not red.'}
}

\subsubsection{Derived}

There are three ways to derive attributive verbs: from nominals,
adjectivals, and post-positions. Attributive verbs in derived form have
the following additional feature giving the derivation suffix, whose
value is {\em none}, since none of the three derivations uses a
suffix:

\avmBegin
\[{derived attributive verb}
  MORPH & 
        \[{}
          DERV-SUFFIX & none\\
        \]\\
\]
\avmEnd

There are three types of derivations to attributive verbs:

\begin{itemize}

\item Nominal derivation:
The sentences below use this type of verb forms:

\eenumsentence{
  \label{examle:attributive-derived-nom}
\item
  \shortex{4}
  {O    & yedi\v{g}in   & benim & elmamd{\i}.}
  {{\tt that} & {\tt eat+PART+P2SG} & {\tt my} & {\tt apple+P1SG+PAST+3SG}}
  {`It was my apple that you ate.'}
\item
  \shortex{6}
  {Bu   & s\"{u}t\"{u}n & son  & kullanma & tarihi & d\"{u}nm\"{u}\c{s}.}
  {{\tt this} & {\tt milk+GEN} & {\tt last} & {\tt usage+P3SG} & 
    {\tt date} & {\tt yesterday+NARR+3SG}}
  {`The expiry date of this milk was yesterday.'}
}

\item Adjectival derivation:
The sentences below give some examples of attributive verbs derived
from adjectivals:

\eenumsentence{
  \label{examle:attributive-derived-mod-1}
\item
  \shortex{4}
  {H{\i}zl{\i} & yazmakta      & olduk\c{c}a & becerikliyim.}
  {{\tt fast} & {\tt write+INF+LOC} & {\tt very} & {\tt skillful+PRES+1SG}}
  {`I am very skillful in writing fast.'}
\item
  \shortex{2}
  {Sen & ka\c{c}{\i}nc{\i}s{\i}n?}
  {{\tt you} & {\tt in what rank+PRES+2SG}}
  {`What is your rank?'}
}

Consider the following feature structure for {\em bor\c{c}luyum}, as
used in~(\ref{example:attributive-derived-mod-2}), which is derived from the
qualitative adjective {\em bor\c{c}lu} ({\em that owing
  debt}). Note that {\em bor\c{c}lu} is also derived from the
common noun, {\em bor\c{c}} ({\em debt}):\footnotemark

\footnotetext{
  This example derivation considers only one sense of {\em bor\c{c}}.
  This process is repeated for all of the senses of this noun
  regardless of the semantics of the derivation with the suffixes
  used. 
  Furthermore, if the morphological processor allows a derivation
  starting from the adjective {\em bor\c{c}lu}, this path is followed,
  as well.
  }

\eenumsentence{
  \label{example:attributive-derived-mod-2}
\item[]
  \shortex{4}
  {Ba\c{s}ar{\i}m{\i} & \c{c}ok   & \c{c}al{\i}\c{s}mama & bor\c{c}luyum.}
  {{\tt success+P1SG+ACC} & {\tt very much} & {\tt work+INF+DAT} & 
    {\tt debtor+PRES+1SG}}
  {`It was my hard working that brought my success.'}
  }

\avmBegin
\[{derived attributive verb}
  CAT &
      \[{}
        MAJ & verb\\
        MIN & attributive\\
      \]\\
  MORPH &
        \[{}
          STEM & \@1\\
          FORM & derived\\
          AGR  & 1sg\\ 
          TAM2 & pres\\
          DERV-SUFFIX & none\\
        \]\\
  SYN &
      \[{}
        SUBCAT & \@2\\
      \]\\
  SEM &
      \[{}
        CONCEPT & f$_{none}$(\@3)\\
      \]\\
  PHON & ``bor\c{c}luyum''\\
\]
\avmEnd

\avmBegin
\@1
\[{derived qualitative adj}
  CAT &
      \[{}
        MAJ & adjectival\\
        MIN & adjective\\
        SUB & qualitative\\
      \]\\
  MORPH &
        \[{}
          STEM & \@4\\
          FORM & derived\\
          DERV-SUFFIX & ``l{\i}''\\
        \]\\
  SYN &
      \[{}
        SUBCAT & \@2\\
        MODIFIES & 
                 \[{}
                   CAT &
                       \[{}
                         MAJ & nominal\\
                         MIN & noun\\
                         SUB & common\\
                       \]\\
                 \]\\
      \]\\
  SEM &
      \[{}
        CONCEPT & \@3 f$_{l{\i}}$(\@5)\\
      \]\\
  PHON & ``bor\c{c}+l{\i}''\\
\]
\avmEnd

\avmBegin
\@4
\[{lexical common}
  CAT &
      \[{}
        MAJ & nominal\\
        MIN & noun\\
        SUB & common\\
      \]\\
  MORPH &
        \[{}
          STEM & ``bor\c{c}''\\
          FORM & lexical\\
        \]\\
  SYN &
      \[{}
        SUBCAT & \@2 \{$constraint_1, constraint_2,  constraint_3$\}\\
      \]\\                         
  SEM & 
      \[{}
        CONCEPT & \@5 \#bor\c{c}-(debt)\\
      \]\\
  PHON & ``bor\c{c}''\\
\]
\avmEnd

\avmBegin
\[{$constraint_1$}
  CAT & 
      \[{}
        MAJ & nominal\\
        MIN & \{{\rm noun, pronoun}\}\\
      \]\\
  MORPH &
        \[{}
          CASE & dat\\
        \]\\
\]
\avmEnd

\avmBegin
\[{$constraint_2$} 
  CAT & 
      \[{}
        MAJ & nominal\\
        MIN & sentential\\
        SUB & act\\
        SSUB & infinitive\\
        SSSUB & ma\\
      \]\\
  MORPH &
        \[{}
          CASE & dat\\
        \]\\
\]
\avmEnd

\avmBegin
\[{$constraint_3$} 
  CAT & 
      \[{}
        MAJ & nominal\\
        MIN & sentential\\
        SUB & act\\
        SSUB & infinitive\\
        SSSUB & y{\i}\c{s}\\
      \]\\
  MORPH & 
        \[{}
          CASE & dat\\
          POSS & $\neg$none\\
        \]\\
\]
\avmEnd


\item Post-position derivation:
The following example demonstrates the derivation from post-position
{\em sonra}:

\eenumsentence{
\item[]
  \shortex{3}
  {Sen & benden & sonras{\i}n.}
  {{\tt you} & {\tt me+ABL} & {\tt after+PRES+2SG}}
  {`You are after me.'}
  }

\end{itemize}

\section{Conjunctions}\label{sec-3:conjunctions}

\begin{figure}[hbt]
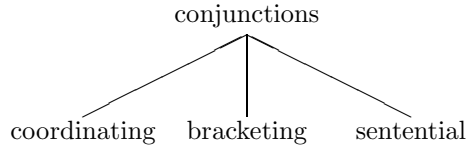

\footnotesize
{
\begin{center}
\leaf{coordinating}
\leaf{bracketing}
\leaf{sentential}
\branch{3} {conjunctions}
\tree
\end{center}
}
\caption{Subcategories of conjunctions.}
\label{figure-3:conjunctions}
\end{figure}

This section describes the representation of conjunctions in our
lexicon. Conjunctions are function words, i.e., they do not convey
meaning when used alone. 
They are used to conjoin words, phrases, and sentences both
syntactically and semantically (see Ediskun \cite{Ediskun}). 
As shown in Figure~\ref{figure-3:conjunctions}, conjunctions are divided
into three subcategories: {\em coordinating}, {\em bracketing} and
{\em sentential conjunctions}. 

The next three sections describe the subcategories of conjunctions
with examples.

\subsection{Coordinating Conjunctions}

The following are examples of coordinating conjunctions: {\em ile}
({\em and}), {\em ve} ({\em and}), {\em veya} ({\em or}), {\em
  ila} ({\em between \ldots{}and}).

Consider the feature structure of the coordinating conjunction {\em
  ve} ({\em and}), as used in the example below:

\eenumsentence{
\item[]
  \label{example:coordinating-cons}
  \shortex{6}
  {Bug\"{u}n & ve & yar{\i}n & hava & bulutlu & olacakm{\i}\c{s}.}
  {{\tt today} & {\tt and} & {\tt tomorrow} & {\tt weather} & {\tt cloudy} & 
    {\tt be+FUT+NARR+3SG}}
  {`They say, today and tomorrow the weather will be cloudy.'}
  }

\avmBegin
\[{coordinating}
   CAT &
      \[{}
        MAJ & conjunction\\
        MIN & coordinating\\
      \]\\
   MORPH &
         \[{}
           STEM & ``ve''\\
         \]\\
   SEM &
       \[{}
         CONCEPT & \#ve-(and)\\
       \]\\
   PHON & ve''\\
\]
\avmEnd

\subsection{Bracketing Conjunctions}

Bracketing conjunctions are used in pairs. These have the following
two semantic features. The first gives the polarity of the
conjunction, e.g., the polarity of {\em ne \ldots{} ne} ({\em neither
\ldots{}nor}) is negative, while it is positive for {\em hem
\ldots{}hem} ({\em both \ldots{}and}). The second specifies how
the two elements bracketed are connected.

\avmBegin
\[{bracketing}
  SEM &
      \[{}
        POLARITY & +/$-$ (default: +)\\
        CONNECTION & and/or (default: and)\\
      \]\\
\]
\avmEnd

The following are some examples of bracketing conjunctions:
{\em gerek \ldots{}gerek}({\em se}) ({\em both \ldots{}and}),
{\em ne \ldots{}ne} ({\em neither \ldots{}nor}), {\em hem \ldots{}hem}
({\em both \ldots{}and}), {\em ya \ldots{}ya} ({\em either
  \ldots{}or}).

The following is the feature structure of the bracketing conjunction,
{\em gerek \ldots{}gerek} ({\em both \ldots{}and}), as used in
(\ref{example:bracketing-cons}):

\eenumsentence{
\item[]
  \label{example:bracketing-cons}
  \shortex{8}
  {Gerek & Y\"{u}cel & gerek & U\v{g}ur & bug\"{u}n & \c{c}ok &
    h{\i}zl{\i} & ko\c{s}tular.} 
  {{\tt both } & {\tt Y\"{u}cel} & {\tt and} & {\tt U\v{g}ur} & 
    {\tt today} & {\tt very} & {\tt fast} & {\tt run+PAST+3PL}}
  {`Both Y\"{u}cel and U\v{g}ur ran very fast today.'}
}

\avmBegin
\[{bracketing}
   CAT &
      \[{}
        MAJ & conjunction\\
        MIN & bracketing\\
      \]\\
   MORPH &
         \[{}
           STEM & ``gerek \ldots{}gerek''\\
         \]\\
   SEM &
       \[{}
         CONCEPT & \#gerek \ldots{}gerek-(both \ldots{}and)\\
       \]\\
   PHON & ``gerek \ldots{}gerek''\\
\]
\avmEnd

\subsection{Sentential Conjunctions}
Sentential conjunctions conjoin sentences.
{\em Ancak} ({\em but}), {\em \c{c}\"{u}nk\"{u}} ({\em because}), {\em
  hatta} ({\em even)}, {\em ama} ({\em but}), {\em nitekim} ({\em
  just as}), {\em e\v{g}er} ({\em if}), {\em yani} ({\em that is to
  say}), and {\em \"{u}stelik} ({\em furthermore}) are some examples
of sentential conjunctions.

\section{Post-positions}\label{sec-3:post_positions}

\begin{figure}[htb]
  \centerline{\psfig{figure=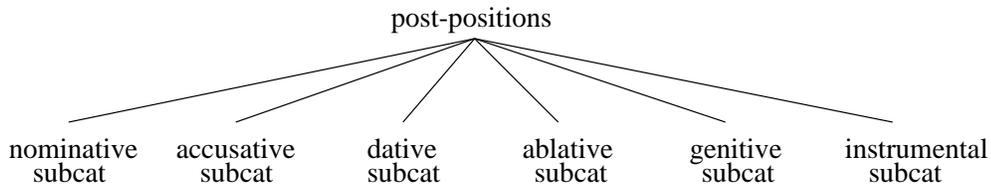}}
  \caption{Subcategories of post-positions.}
  \label{figure-3:post-positions}
\end{figure}

This section describes the representation of post-positions in our
lexicon. Like conjunctions, post-positions are function words, i.e.,
they do not have meaning, unless they are used with nominals in order
to construct post-positional phrases (see Ediskun \cite{Ediskun}).
As shown in Figure~\ref{figure-3:post-positions}, post-positions are
subdivided into six categories according to their subcategorization
types (specifically, the case of the complement).

Each post-position also has the following feature, which gives the
subcategorization information for only one argument, in contrast to
the case in verbs, which accept a number of arguments, such as
subject, direct object, etc. For this reason the subcategorization
information of post-positions  consists of just a list of constraints
for only one argument.

\avmBegin
\[{post-position}
  SYN &
      \[{}
        SUBCAT & {\em subcat}\\
      \]\\
\]
\avmEnd

In the next sections we will describe the subcategories and give
examples for each of them.

\subsection{Post-positions with Nominative Subcategorization}

Post-positions belonging to this subcategory accept nominals in
{\em nominative} case as complements. {\em Boyunca} ({\em along}/{\em
  during}), {\em takdirde} ({\em if}), {\em diye} ({\em named}), {\em
  i\c{c}in} ({\em for}) are examples of post-positions with nominative
subcategorization. 

The feature structure of the post-position, {\em i\c{c}in} ({\em
  for}/{\em because}/{\em in order to}), as used in
(\ref{example:postp-1}), is given below, though the case of the
complement is {\em genitive} for pronouns:

\eenumsentence{
  \label{example:postp-1}
\item
  \shortexnt{5}
  {Almay{\i} & unuttu\v{g}um & kitaplar & i\c{c}in & odama}
  {{\tt take+INF+ACC} & {\tt forget+PART+P1SG} & {\tt book+3PL} & 
    {\tt for} & {\tt room+P1SG+DAT}}
  \newline
  \shortex{2}
  {tekrar & gittim.}    
  {{\tt again } & {\tt go+PAST+1SG}}
  {`I went to my room again for the books that I forgot to take.'}
\item
  \shortexnt{5}
  {Ba\c{s}ar{\i}l{\i} & olabilmesi & i\c{c}in & \c{c}ok &
    \c{c}al{\i}\c{s}mas{\i}}
  {{\tt succesfull} & {\tt be+ABIL+INF+P3SG} & {\tt for} & 
    {\tt much} & {\tt work+INF+P3SG}}
  \newline
  \shortex{1}
  {gerekiyor.}
  {{\tt needed+PROG+3SG}}
  {`In order to be successful, he should work hard.'}
}

\avmBegin
\[{nom-subcat}
  CAT &
      \[{}
        MAJ & post-position\\
        MIN & nom-subcat\\
      \]\\
  MORPH &
        \[{}
          STEM & ``i\c{c}in''\\
        \]\\
  SYN &
      \[{}
        SUBCAT & \{$constraint_1, constraint_2, constraint_3$,\\
                   $constraint_4, constraint_5, constraint_6$\}\\
      \]\\
  SEM &
      \[{}
        CONCEPT & \#i\c{c}in-(for/because/in order to)\\
      \]\\
  PHON & ``i\c{c}in''\\
\]
\avmEnd

\avmBegin
\[{$constraint_1$} 
  CAT & 
      \[{}
        MAJ & nominal\\
        MIN & noun\\
      \]\\
  MORPH &
        \[{}
          CASE & nom\\
        \]\\
\]
\avmEnd

\avmBegin
\[{$constraint_2$} 
  CAT & 
      \[{}
        MAJ & nominal\\
        MIN & pronoun\\
      \]\\
  MORPH &
        \[{}
          CASE & gen\\
        \]\\
\]
\avmEnd

\avmBegin
\[{$constraint_3$} 
  CAT & 
      \[{}
        MAJ & nominal\\
        MIN & sentential\\
        SUB & act\\
        SSUB & infinitive\\
        SSSUB & mak\\
      \]\\
  MORPH & 
        \[{}
          CASE & nom\\
          POSS & none\\
        \]\\
\]
\avmEnd

\avmBegin
\[{$constraint_4$} 
  CAT & 
      \[{}
        MAJ & nominal\\
        MIN & sentential\\
        SUB & act\\
        SSUB & infinitive\\
        SSSUB & ma\\
      \]\\
  MORPH & 
        \[{}
          CASE & nom\\
          POSS & $\neg$none\\
        \]\\
\]
\avmEnd

\avmBegin
\[{$constraint_5$} 
  CAT & 
      \[{}
        MAJ & nominal\\
        MIN & sentential\\
        SUB & act\\
        SSUB & infinitive\\
        SSSUB & y{\i}\c{s}\\
      \]\\
  MORPH &
        \[{}
          CASE & nom\\
        \]\\
\]
\avmEnd

\avmBegin
\[{$constraint_6$} 
  CAT & 
      \[{}
        MAJ & nominal\\
        MIN & sentential\\
        SUB & act\\
        SSUB & participle\\
      \]\\
  MORPH & 
        \[{}
          CASE & nom\\
          POSS & $\neg$none\\
        \]\\
\]
\avmEnd

\subsection{Post-positions with Accusative Subcategorization}

Post-positions belonging to this subcategory accept nominals in
{\em accusative} case as complements. The following examples are
post-positions belonging to this category: {\em a\c{s}k{\i}n} ({\em
  over}), {\em takiben} ({\em following}), {\em m\"{u}teakiben} ({\em
  following}). 

\subsection{Post-positions with Dative Subcategorization}

Post-positions belonging to this subcategory accept nominals in
{\em dative} case as complements. The following examples are
post-positions belonging to this category: {\em ait} ({\em belonging
  to}), {\em g\"{o}re} ({\em according to}), {\em dek} ({\em until}),
{\em kar\c{s}{\i}n} ({\em in spite of}), {\em y\"{o}nelik} ({\em aimed
  at}), {\em do\v{g}ru} ({\em towards}), {\em ili\c{s}kin} ({\em
  related to}). 

\subsection{Post-positions with Ablative Subcategorization}

Post-positions belonging to this subcategory accept nominals in
{\em ablative} case as complements.  {\em Dolay{\i}} ({\em due to}), 
{\em \"{o}t\"{u}r\"{u}} ({\em due to}), {\em itibaren} ({\em starting
  from}), {\em sonra} ({\em after}), and {\em \"{o}nce} ({\em before})
are examples of post-positions with {\em ablative} subcategorization.

\subsection{Post-positions with Genitive Subcategorization}

Post-positions belonging to this subcategory accept nominals
(specifically, pronouns) in {\em genitive} case as complements. 
{\em \.{I}le} ({\em with}) is an example of this type of post-positions.

\subsection{Post-positions with Instrumental Subcategorization}

Post-positions belonging to this subcategory accept nominals in
{\em instrumental} case as complements. The following post-positions 
are examples of this category: {\em birlikte} ({\em together}), 
{\em beraber} ({\em together}).



\chapter{Operational Aspects of the Lexicon}
\label{chapter:operational-aspects}

Our lexicon provides necessary morphosyntactic, syntactic, and
semantic information to NLP subsystems performing syntactic analysis,
tagging, semantic disambiguation, etc.  

The whole system consists of three main parts: 

\begin{enumerate}
\item a morphological processor/analyzer,
\item a static lexicon, and
\item a module filtering the output according to the user's
  restrictions.
\end{enumerate}

\begin{figure}[p]
  \centerline{\psfig{figure=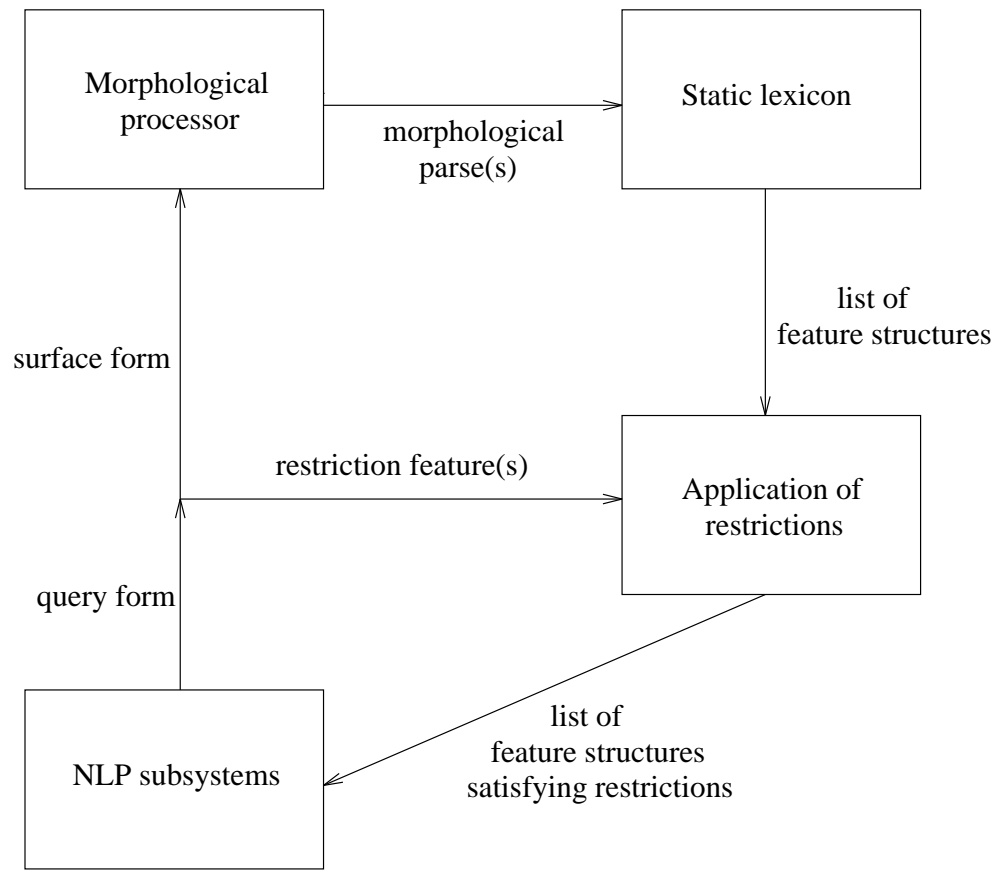}}
  \caption{Data flow in the lexicon.}
  \label{figure-4:architecture}
\end{figure}

As depicted in Figure~\ref{figure-4:architecture}, the system receives
a query form, which includes, at least, a surface form and other
information acting as the restrictions on the output feature
structures. The surface form is first directed to the morphological
processor, which generates all possible interpretations (i.e., parses
or lexical forms) and forwards these to the static lexicon. The static
lexicon accesses feature structure database and retrieves syntactic
and semantic information for the root words involved in the
interpretations. Having unified the morphosyntactic information
provided with corresponding syntactic and semantic information
retrieved, the static lexicon outputs a list of feature structures.
The final step in the process is the elimination of the feature
structures which do not satisfy the user's restrictions.

In this way, the NLP subsystems using the lexicon do not need to
interface with the morphological processor to obtain interpretations,
rather they  just provide the surface form and receive the 
corresponding feature structures containing morphosyntactic,
syntactic, and semantic information.

In this chapter, we will first describe the interface to the
lexicon. Section~\ref{sec-4:producing-fss} describes how the system
produces feature structures step by step by giving examples, and
Section~\ref{sec-4:problems} mentions problems and limitations related
with this task. 

\section{Interfacing with the Lexicon}
\label{sec-4:interface}

We presented many examples of feature structures in
Chapter~\ref{chapter:design} and will describe the method of producing
those feature structures in the next section.
In this section, we will mainly concentrate on how NLP subsystems can
use our lexicon.

\begin{figure}[t]
  \centerline{\psfig{figure=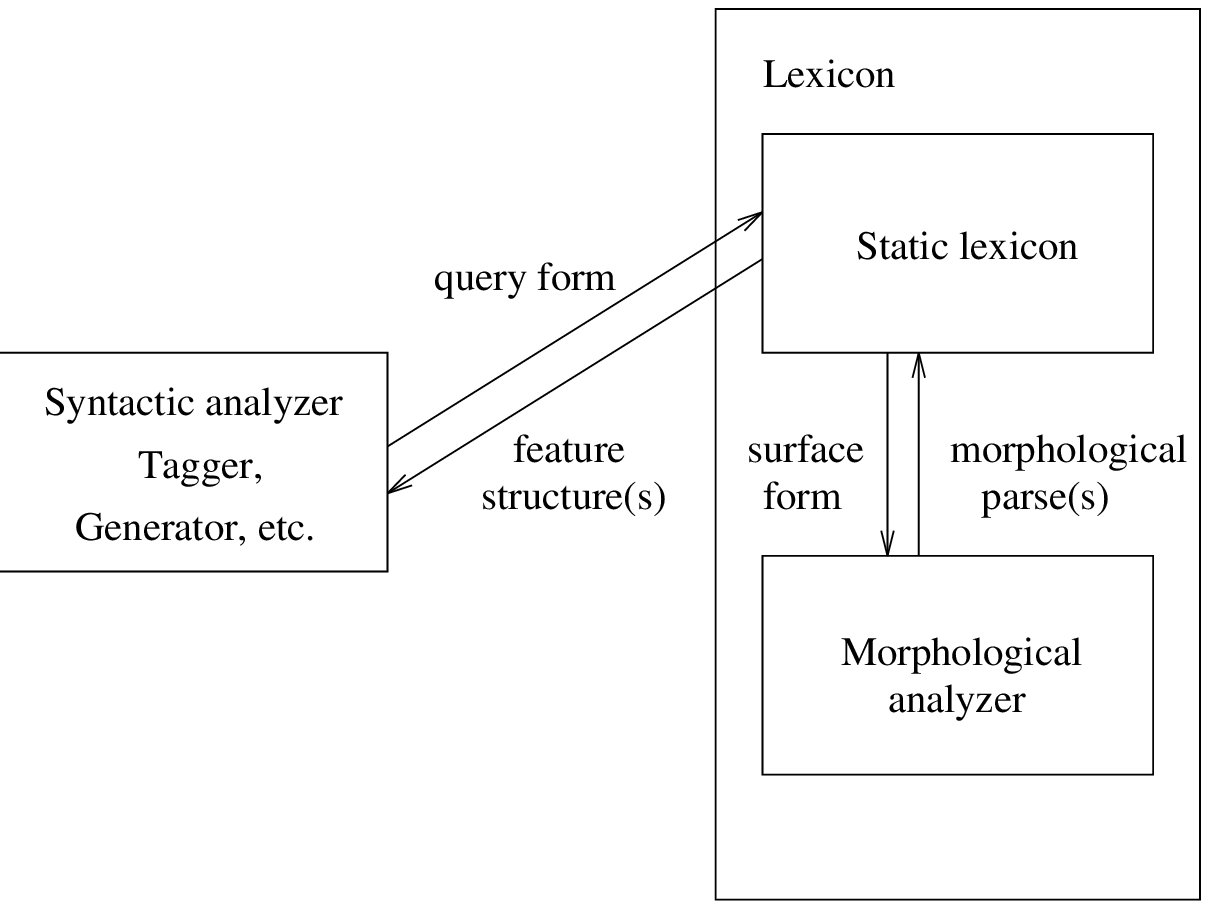}}
  \caption{NLP subsystems interfacing with the lexicon.}
  \label{figure-4:syn_analysis}
\end{figure}

Our lexicon is a front end for a morphological analyzer. Given a
surface form with restriction features, it generates all the
morphosyntactic, syntactic, and semantic information for this surface
form, that is it abstracts morphological analysis and associates
syntactic and semantic information with each interpretation (see
Figure~\ref{figure-4:syn_analysis}). 

The interface described above can be used by a syntactic analyzer for
Turkish.  Additionally, taggers and word sense disambiguators can
employ our lexicon.  Taggers need to set necessary constraints, which
are generally on category and morphosyntactic features, in the query
form.  Consider the following example:

\eenumsentence{
  \item
    \label{example-4:evin_1}
    \shortex{2}
    {evin & kap{\i}s{\i}}
    {{\tt house+GEN} & {\tt door+P3SG}}
    {`door of the house'}
  \item
    \label{example-4:evin_2}
    \shortex{2}
    {senin & evin}
    {{\tt you+GEN} & {\tt house+P2SG}}
    {`your house'}
  }

In the two noun phrases above, the surface form {\em evin} exists with
two different interpretations: in the first one, it is {\em genitive}
case-marked and singular with no possessive marking, whereas in the
second one it is {\em nominative} case-marked with {\em 2sg}
possessive marking.  The ambiguity can be resolved with the help of
morphological features, i.e., case or possessive markings.

Word sense disambiguation is also possible by making use of semantic
features in the feature structures. For example, the two senses of the
root word {\em kazma} ({\em stupid person} and {\em pickaxe}) can be
resolved by setting the SEM~$|$~ANIMATE feature in the query form
properly. Adding semantic features increases the accuracy of
word sense disambiguation process. 
However, rather than adding arbitrary semantic features on demand,
constructing an ontology describing concepts via a semantic network
would be more useful.

Text generators for Turkish or transfer units to Turkish in machine
translation systems can also make use of our lexicon to obtain
information about root words. 
However, the SEM~$|$~CONCEPT feature may not be directly usable by
transfer units, since the English definition in this feature is mostly
human oriented.

The input query form is basically a feature structure, which
contains two types of information: a surface form and a set of other
features. 
The surface form guides the system in producing the feature structures,
that is it is the actual input for the output of the lexicon. 
It is specified as the phonology information (the PHON feature) in the
query form. 
The rest of the features are optional and act as restrictions on the
output structures. 
In fact, the query form {\em subsumes} each of the actual output
feature structures. 
Any set of features can be specified in the query form provided that
they are {\em consistent} and {\em appropriate} for the intended
structure. 

The process of eliminating or filtering the output feature structures
that do not satisfy the restrictions in the query form is the last
step in the whole process. 

Consider the following query form placing  morphosyntactic and
semantic restrictions on the surface form {\em ekimde}, that is the root
word should not be possessive-marked, and its semantics should state
temporality. 

\avmBegin
\[{query form}
  MORPH &
        \[{}
          POSS & none\\
        \]\\
  SEM & 
      \[{}
         TEMPORAL & +\\
      \]\\
  PHON & ``ekimde''\\
\]
\avmEnd

According to the morphological processor, there are two
interpretations of {\em ekimde}:

\begin{enumerate}
\item {\em Ekimde} ({\em in October}):
  The first interpretation is a lexical common noun representing a
  month of the year, as used in the following sentence:

  \eenumsentence{
  \item
    \shortex{4}
    {Bu & i\c{s}i & Ekim'de & bitirmeliydik.}
    {{\tt this} & {\tt job} & {\tt October+LOC} & {\tt finish+NECS+PAST+1PL}}
    {`We should have finished this job in October.'}
    }

  Regarding this interpretation the system produces the following
  feature structure:
    
  \avmBegin
  \[{lexical common}
    CAT &
       \[{}
         MAJ & nominal\\
         MIN & noun\\
         SUB & common\\
       \]\\
   MORPH &
         \[{}
           STEM & ``ekim''\\
           AGR & 3sg\\
           POSS & none\\
           CASE & loc\\
         \]\\
   SYN & 
       \[{}
         \ldots\\
       \]\\
   SEM & 
       \[{}
         TEMPORAL & +\\
         \ldots\\
        \]\\
    PHON & ``ekimde''\\
  \]
  \avmEnd

  The query form subsumes the structure above, hence it satisfies the
  restrictions.
  
\item 
  {\em ekimde} ({\em in my appendix}/{\em suffix}):
  The second interpretation is also a lexical common noun, for which
  there are two senses in the static lexicon: {\em appendix} and {\em
    suffix}. Feature structures for both of the senses are similar, so
  we will consider only the first one, {\em appendix}, which is used
  in the following sentence:

  \eenumsentence{
  \item
    \shortex{5}
    {O & \c{s}ekil & benim & ekimde & olmal{\i}yd{\i}.}
    {{\tt that} & {\tt figure} & {\tt my} & {\tt appendix+P1SG+LOC} 
      & {\tt be+NECS+PAST+3SG}}
    {`That figure should have been in my appendix.'}
    }

  The full feature structure for the second interpretation, {\em in my
    appendix}, is the following: 

  \avmBegin
  \[{lexical common}
    CAT &
        \[{}
          MAJ & nominal\\
          MIN & noun\\
          SUB & common\\
        \]\\
    MORPH &
          \[{}
            STEM & ``ek''\\
            AGR & 3sg\\
            POSS & 1sg\\
            CASE & loc\\
          \]\\
    SYN & 
        \[{}
          \ldots\\
        \]\\
    SEM & 
        \[{}
          TEMPORAL & $-$\\
          \ldots\\
        \]\\
    PHON & ``ekimde''\\
  \]
  \avmEnd

  Due to the $-$ value of SEM~$|$~TEMPORAL and {\em 1sg} value of
  MORPH~$|$~POSS features, the subsumption of the feature structure
  above with the query form will fail, and it will be eliminated.
  Note that both of the restriction features are appropriate for the
  feature structures above.

\end{enumerate}

\section{Producing Feature Structures}
\label{sec-4:producing-fss}

We will describe the processing in the lexicon as consisting of three
main steps: 

\begin{enumerate}
\item morphological analysis,
\item retrieval of syntactic and semantic information and unification
  with morphosyntactic information,
\item application of restrictions.
\end{enumerate}

The first step is external to the system, so we will consider only its
input/output interface. The second step consists of transformation of
morphological parses to feature structure syntax, category mapping,
retrieval from static lexicon, and computing features according to the
morphological parses. The final step is relatively simple; it just
tests the sumbsumtion of input query form with each of the produced
structures.

In the next sections, we will examine each step and provide details
with examples.

\subsection{Morphological Analysis}
\label{sec-4:mp}

Morphological processor provides possible interpretations of a surface
form. Due to the rich set of inflectional and derivational suffixes in
Turkish, it is highly probable that the surface form will have more
than one interpretation.  Consider the possible interpretations of the
surface form {\em kazma}, for which the morphological processor output
is given in Figure~\ref{figure-4:kazma}, as used in the following
examples:

\eenumsentence{
  \label{example-4:kazma}
  \item
    \shortex{6}
    {D\"un & burada & bir & kazma & g\"ord\"un & m\"u?}
    {{\tt yesterday} & {\tt here} & {\tt a} & {\tt pickaxe} & 
      {\tt see+PAST+2SG} & {\tt QUES}}
    {`Did you see a pickaxe here yesterday?'}
  \item
    \shortex{3}
    {Oray{\i} & sak{\i}n & kazma!}
    {{\tt there} & {\tt never} & {\tt dig+NEG+2SG}}
    {`Do not dig there!'}
  \item
    \shortexnt{4}
    {Kazma   & i\c{s}ini & san{\i}r{\i}m & bug\"un}
    {{\tt dig+INF} & {\tt job+P3SG+ACC} & {\tt guess+ARST+1SG} &
      {\tt today}}
    \newline
    \shortex{1}
    {bitiririz.}
    {{\tt finish+ARST+1PL}}
    {`I guess we will finish digging today.'}
  }

\begin{figure}[htb]
  \fnsCBegin
\begin{verbatim}
     1. [[CAT=NOUN][ROOT=kazma][AGR=3SG][POSS=NONE][CASE=NOM]]
     2. [[CAT=VERB][ROOT=kaz][SENSE=NEG][TAM1=IMP][AGR=2SG]]
     3. [[CAT=VERB][ROOT=kaz][SENSE=POS]
         [CONV=NOUN=MA][TYPE=INFINITIVE][AGR=3SG][POSS=NONE][CASE=NOM]]
\end{verbatim}
  \fnsCEnd
  \caption{Interpretations of the surface form {\em kazma}.}
  \label{figure-4:kazma}
\end{figure}

The first interpretation contains the noun reading, {\em  pickaxe}. 
The second and third interpretations consider the verb
{\em kaz-} ({\em dig}). In the second interpretation, the suffix
{\em ma} is an inflectional suffix and negates the predicate, as
opposed to the other one, which is a derivational suffix and used to
derive the infinitive {\em kazma} ({\em digging}).

As seen in the example above, the rich set of inflectional and
derivational suffixes causes many interpretations, which increase in
number when the multiple senses are incorporated. For example, the
predicative verb {\em ye} has at least four senses, which we mentioned
in Section~\ref{sec-3:predicative-verbs}.

The morphological processor output must be transformed to feature
structure syntax, moreover, due to the comprehensive categorization
introduced in Chapter~\ref{chapter:design}, category mapping will take
place.  The following section describes this transformation and
retrieving information in the static lexicon.

\subsection{Retrieving Information in the Static Lexicon}
\label{sec-4:retrieving-info}

The static lexicon follows the interpretations produced by the
morphological processor. Interpretations include category information,
the root words, and a number of inflectional and derivational
suffixes, such as case and possessive markers. 
The retrieval step mainly consists of the following phases:

\begin{itemize}
\item transformation of interpretations into feature structure
  syntax, and correct mapping from the morphological processor
  category to the static lexicon category,
\item accessing the feature structures of the root words involved in
  the morphological parses, and computing features accordingly.
\end{itemize}

During the processing, the system accesses two tables and two
databases. The tables are used to map category information, and the
databases are used to access feature structures of the root words
containing syntactic and semantic information (i.e., lexical
database), and the template structures.

The retrieval process starts with transformation of parses into feature
structure syntax, since the syntactic and semantic information is
stored in the form of feature structures in the static lexicon. As
seen in the interpretations of {\em kazma} in the previous section,
derivations exist in morphological parses and may go to arbitrary
depth, such as {\em
  \c{C}ekoslovakyal{\i}la\c{s}t{\i}ramad{\i}klar{\i}m{\i}zdanm{\i}\c{s}s{\i}n{\i}z}.

As another example for the interpretations containing derivations,
consider the one in Figure~\ref{figure-4:akillica}. It starts with the
noun ak{\i}l ({\em intelligence}), which is used to derive the
adjective {\em ak{\i}ll{\i}} ({\em intelligent}). The derivations end
with the manner adverb {\em ak{\i}ll{\i}ca} ({\em intelligently}).
The derivations in the processor output are highlighted with the CONV
item in the string below, which gives the category and derivational
suffix. Thus, in the following example, there are two derivations and
three categories traversed, that is there are three levels: the first
is the lexical level and the other two are the derivational levels.
Each level is transformed into a feature structure containing category
and morphosyntactic information. So, the interpretation above would be
transformed into a list of levels with three elements.

\begin{figure}[htb]
  \fnsCBegin
  \begin{verbatim}
     [[CAT=NOUN][ROOT=akIl][CONV=ADJ=LI][CONV=ADVERB=CA][TYPE=MANNER]]
  \end{verbatim}
  \fnsCEnd
  \caption{The derivation path to the manner adverb {\em
      ak{\i}ll{\i}ca}.}
  \label{figure-4:akillica}
\end{figure}

While transforming the interpretations, the system maps the category
information in the morphological processor output to correct lexicon
category for all levels, which is due to the finer-grained
categorization of the lexicon. For this purpose, two tables are
maintained for root words and derivations, respectively.  For the
first one, processor category and root word uniquely determine the
lexicon category. For each root word represented in the feature
structure database, an entry in this table must be present. A portion
of such a table for nouns is depicted in~Figure~\ref{figure-4:hash}.
For the second table, processor category and derivational suffix
uniquely determine the lexicon category. This mapping is given in
Table~\ref{table-4:cat-map}.

\begin{figure}[htbp]
  \begin{center}
  \begin{tabular}{|l|l|l|}\hline
    Processor category & Root word & Lexicon category \\ \hline \hline
    \ldots       &  \ldots         & \ldots \\ \hline
    noun         &  kaz{\i}nt{\i}  & common noun \\ \hline
    {\bf noun}   &  {\bf kazma}    & {\bf common noun} \\ \hline
    noun         &  kazmano\v{g}lu & proper noun \\ \hline
    noun         &  ket\c{c}ap     & common noun \\ \hline
    noun         &  \ldots         & \ldots         \\ \hline
    noun         &  kurtulu\c{s}   & proper noun \\ \hline
    \ldots       &  \ldots         & \ldots         \\ \hline
  \end{tabular}
  \end{center}
  \caption{A portion of the table used for category mapping for root
    words.}
  \label{figure-4:hash}
\end{figure}

\begin{table}[t]
  \fnsCBegin
\begin{tabular}{|l|l|l|l|l|l|l|}\hline
\multicolumn{2}{|c|}{Morphological Processor  Output} 
         & \multicolumn{5}{c|}{Lexicon Category} \\ \hline
Category & Suffix 
             & MAJ         & MIN        & SUB    & SSUB & SSSUB \\ 
             \hline \hline
noun     & c{\i}, l{\i}k, c{\i}k, og,
             & nominal     & noun       & common & & \\
         & y{\i}c{\i}, mazl{\i}k,
             &             &            &        & & \\
         & yamazl{\i}k, maca,
             &             &            &        & & \\
         & yas{\i}, {\em none}
             &             &            &        & & \\ \hline
         & mak
             &             & sentential & act  & infinitive  & mak\\ 
             \hline
         & ma
             &             &            &      &             & ma\\ 
             \hline
         & y{\i}\c{s}
             &             &            &      &             & y{\i}\c{s}\\ 
             \hline
         & d{\i}k
             &             &            & fact & participle & d{\i}k\\
             \hline
         & yacak
             &             &            &      &            & yacak\\ 
             \hline \hline
rpronoun & {\em none}
             & noinal      & pronoun    & quantitative &    &      \\
             \hline \hline
adj      & l{\i}k, l{\i}, ki, s{\i}z, s{\i},
             & modifier    & adjective    & qualitative & & \\
         & ik, y{\i}c{\i}, yan, yacak, 
             &             &             &             & & \\
         & d{\i}k, yas{\i}
             &             &             &             & & \\ 
             \hline \hline
adverb   & y{\i}nca, y{\i}p
             & adverbial   & temporal    & point-of-time & & \\ \hline
         & yal{\i}, ken
             &             &             & time-period   & fuzzy & \\ 
             \hline
         & cas{\i}na, maks{\i}z{\i}n,
             &             & manner      & qualitative & & \\
         & madan, yamadan,
             &             &             &             & & \\
         & yerek, ca
             &             &             &             & & \\ \hline
         & d{\i}k\c{c}a
             &             &             & repetition  & & \\ 
             \hline \hline
verb     & lan, la\c{s}
             & verb        & predicative &           & & \\ \hline
         & {\em none}
             & verb        & attributive &           & & \\ \hline
\end{tabular}
\fnsCEnd
\caption{The table used for category mapping for derived words.}
\label{table-4:cat-map}
\end{table}

This step is applied to all of the morphological parses, and at the
end of this step, for each parse there is a list of levels, each of
which  contains the correct lexicon category and a set of features
representing morphosyntactic information of interpretations. 

%

The next phase in the processing is the retrieval of the syntactic and
semantic information and producing feature structures. The syntactic
and semantic information about the root words is stored in the feature
structure database, which is indexed with the category and the root
word information. For the root words in the lexical levels of each
parse, the feature structure database is accessed and matching entries
are retrieved. However, the entries contain only syntactic and
semantic information for the non-derived forms, thus morphosyntactic
information needs to be unified and by following the derivation
information of parses new feature structures should be
constructured. Many examples of this phenomenon are presented in the
Chapter~\ref{chapter:design}. 

Since the morphological parses are previously transformed into feature
structure syntax, unification of morphosyntactic information is
simple. Having unified all the information, the processing for the
lexical level is completed. If the morphological parses do not
contain a derivation to another category, the process above is
sufficient to produce the result.  However, as we have already
mentioned, the cases in which derivations exist are not rare.

For each derivation in the parses, a new feature structure is
constructed. For this purpose, using the category information 
in the derivational levels, the template feature structure database is
accessed and corresponding template feature structures are
retrieved. These structures do not contain feature values, but they 
will be computed by the system.

\begin{figure}[ht]
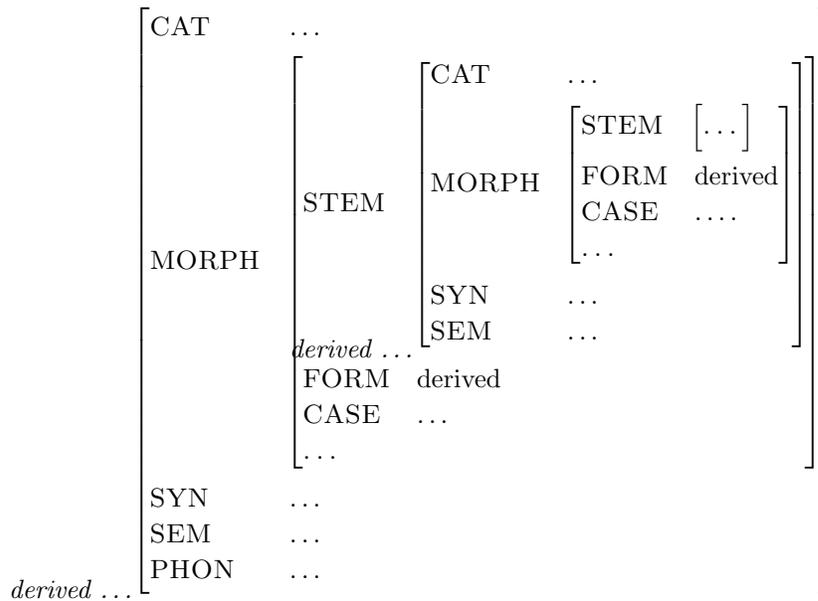

  \avmBegin
  \[{derived \ldots}
    CAT & \ldots\\
    MORPH &
          \[{}
            STEM & 
                 \[{derived \ldots}
                   CAT & \ldots\\
                   MORPH &
                         \[{}
                           STEM &
                                \[{}
                                  \ldots
                                \]\\
                           FORM & derived\\
                           CASE & \ldots.\\
                           \ldots  & \\ 
                         \]\\
                   SYN & \ldots\\
                   SEM & \ldots\\
                 \]\\
            FORM & derived\\
            CASE & \ldots\\
            \ldots  & \\
          \]\\
     SYN & \ldots\\
     SEM & \ldots\\
     PHON & \ldots\\
   \]
  \avmEnd
  \caption{Nested feature structures.}
  \label{figure-4:stem}
\end{figure}

Starting from the leftmost derivational level, the derivation path is
followed: for each derivation a new feature structure is constructed;
feature values are computed. The result is a nested feature
structure, in which the previous structures are stored in
MORPH~$|$~STEM feature as shown in Figure~\ref{figure-4:stem}.

Having retrieved the template feature structure, the feature values
are to be computed by the system. Morphosyntactic information is
already produced by the morphological processor, and unified with the
information in the template structures. A feature structure belonging
to any category should has the following minimum information:
category, phonology, stem, concept, and form. Among them the category
information and the form (i.e., it is {\em derived}) are already
known.  The feature MORPH~$|$~STEM holds the feature structures of the
previous words, as described above. The phonology information is valid
only in the last feature structure in the derivation, whose value is
the surface form given as the input to the morphological
processor.\footnotemark\, The concept feature is computed by means of
a function according to the target derivation category and suffix.

\footnotetext{
  In other structures, this value is undefined, although computation is
  possible by means of  morphological generation.
  }

There are other features to be computed other than the common ones,
among which subcategorization information and thematic roles are the most
important ones. These are co-indexed with the those of the previous
derivational level. Furthermore, a number of features specific to some
categories exist, e.g., semantic properties of common nouns or the
constraints on the modified of qualitative adjectives.  About the
second one, for example, the following prediction can be made:
qualitative adjectives modify the common nouns, and do not constrain
the agreement and countability features. However, predicting the
semantic properties is difficult, and for this reason, the default
values are used, which may not always give the correct description.

In the next section we will clarify the procedure above by giving
examples. 

\subsubsection{Examples}
\label{sec-4:examples}

In summary, the process of producing feature structures follows the
following steps:

\begin{enumerate}
\item[1:] For each parse in the morphological processor output do the
  following: 
  \begin{enumerate}
  \item[1.1:] Find the lexicon category of the initial root word (see
    the table in Figure~\ref{figure-4:hash}),
  \item[1.2:] Find the lexicon entries of all senses of the root word
    by matching the root word information,
  \item[1.3:] Unify morphosyntactic information with the information in
    the lexicon entry/entries, 
  \item[1.4:] While there is derivation in the parse do the following:
    \begin{enumerate}
    \item[1.4.1:] Find the lexicon category and retrieve the
      corresponding template feature structure (see
      Table~\ref{table-4:cat-map}),
    \item[1.4.2:] Compute feature values and unify morphosyntactic
      information, 
    \end{enumerate}
  \item[1.5:] Output the feature structure(s)
  \end{enumerate}
\end{enumerate}

We will describe the process with the input surface form {\em kazma},
which has three interpretations, one of which includes a derivation
(see example~(\ref{example-4:kazma}) and Figure~\ref{figure-4:kazma}
for morphological processor output):

\begin{enumerate}
\item
  {\em Kazma} (common noun):
  This interpretation is due to the common noun {\em kazma} ({\em
    pickaxe}), and  does not contain a derivation, so
  the result can be easily produced by combining morphosyntactic,
  syntactic, and semantic information.

  As we already described, the process starts with determining the lexicon
  category. The morphological processor categorizes {\em kazma} just as
   a noun, however, it is represented as a {\em common noun} in the
  static lexicon. Then, the corresponding feature structure in the 
  lexicon is searched by matching the ROOT information of morphological
  processor with MORPH~$|$~STEM feature of lexicon entries. The matching
  feature structure is given below. Note that there is only one sense
  of {\em kazma} ({\em pickaxe}) in our lexicon.

  \avmBegin
  \[{lexical common}
    CAT &
        \[{}
          MAJ & nominal\\
          MIN & noun\\
          SUB & common\\
        \]\\
    MORPH &
          \[{}
            STEM & ``kazma''\\
            FORM & lexical\\
          \]\\
    SYN &
        \[{}
          SUBCAT & none\\
        \]\\
    SEM &
        \[{}
          CONCEPT & \#kazma-(pickaxe)\\
          COUNTABLE & +\\
      \]\\
  \]
  \avmEnd

  Then, information about inflectional suffixes are unified with the
  lexicon entry, which produces the result:

  \avmBegin
  \[{lexical common}
    CAT &
        \[{}
          MAJ & nominal\\
          MIN & noun\\
          SUB & common\\
        \]\\
    MORPH &
          \[{}
            STEM & ``kazma''\\
            FORM & lexical\\
            CASE & nom\\
            AGR & 3pl\\
            POSS & 1sg\\
          \]\\
    SYN &
        \[{}
          SUBCAT & none\\
        \]\\
    SEM &
        \[{}
          CONCEPT & \#kazma-(pickaxe)\\
          COUNTABLE & +\\
        \]\\
    PHON & ``kazma''\\
  \]
  \avmEnd

  Note that the phonology information is the same as surface form
  given as an input to the system.
  
\item
  {\em Kazma} (verb): This interpretation comes from the verbal root
  {\em kaz-} ({\em dig}). The suffix {\em ma} is an inflectional
  suffix, which negates the meaning (see~Figure~\ref{figure-4:kazma}
  for the parse). Since no derivation step is involved, the
  process is similar to that of the common noun reading. The lexicon
  entry is given below with the morphosyntactic information unified:

  \avmBegin
  \[{lexical predicative verb}
    CAT &
        \[{}
          MAJ & verb\\
          MIN & predicative\\
        \]\\
    MORPH &
          \[{}
            STEM & ``kaz''\\
            FORM & lexical\\
            SENSE & neg\\
            TAM1 & imp\\
            AGR & 2sg\\
          \]\\
    SYN &
        \[{}
          SUBCAT & \ldots\\
        \]\\
    SEM & 
        \[{}
          CONCEPT & \#kaz-(to dig)\\
          ROLES & \ldots\\
        \]\\
    PHON & ``kazma''\\
  \]
  \avmEnd

\item
  {\em Kazma} (infinitive):
  This interpretation involves a derivation from the verb {\em kaz-}
  ({\em dig}) to the infinitive {\em kazma} ({\em digging}). The
  steps up to the derivation is similar to that of the previous two
  examples. 
  The derivation step starts with the determination of the target
  category using the Table~\ref{table-4:cat-map}, and retrieval of the
  template feature structure. 
  The table lookup results in the {\em infinitive} category, and
  corresponding template feature structure is retrieved.

  The next step involves the computation of features, which includes
  subcategorization information, thematic roles, and concept. These
  features, except the concept, are co-indexed with the corresponding
  entries in the lexicon entry of {\em kaz-}.
  The concept feature is computed via a function. The rest of the
  features can be easily found, since category is already known and
  morphosyntactic information is received from the morphological
  processor.
  The phonology feature takes the input surface form, {\em kazma}.

  The feature structure for the infinitive  {\em kazma}  is given
  below, with some of the features co-indexed with those of the
  lexical entry of {\em kaz-}:

  \avmBegin
  \[{ma}
    CAT &
        \[{} 
          MAJ & nominal\\
          MIN & derived\\
          SUB & act\\
          SSUB & infinitive\\
          SSSUB & ma\\
        \]\\
    MORPH &
          \[{}
            STEM & \@1\\
            DERV-SUFFIX & ``ma''\\
            FORM & derived\\
            CASE & nom\\
            AGR & 3sg\\
            POSS & none\\
          \]\\
    SYN &
        \[{}
          SUBCAT & \@2\\
        \]\\
    SEM &
        \[{}
          CONCEPT & f$_{ma}$(\#kaz-(dig))\\
          ROLES & \@3\\
        \]\\
    PHON & ``kazma''\\
  \]
  \avmEnd

  \avmBegin
  \@1
  \[{lexical predicative verb}
    CAT &
        \[{}
          MAJ & verb\\
          MIN & predicative\\
        \]\\
    MORPH &
          \[{}
            STEM & ``kaz''\\
            FORM & lexical\\
            SENSE & pos\\
          \]\\
    SYN &
        \[{}
          SUBCAT & \@2 \ldots\\
        \]\\
    SEM & 
        \[{}
          CONCEPT & \#kaz-(dig)\\
          ROLES & \@3 \ldots\\
        \]\\
    PHON & none\\
  \]
  \avmEnd
  
\end{enumerate}
  

\subsection{Application of Restrictions}
\label{sec-4:application}

The final step in the process is the elimination of the feature
structures that do not satisfy the restrictions.

The input to this phase is a list of feature structures and the user's
query form. Each structure is tested against the query form for
subsumtion relation, that is all of the features in the query form
must be present in the output structures and the feature values must
be the same. The ones that fail to satisfy this relation are
eliminated. 

The process is relatively simple, thus we will not decribe it any
further (see the example in Section~\ref{sec-4:interface}). 


\section{Problems and Limitations}
\label{sec-4:problems}

A limitation with the representation of the entries in the static
lexicon is related with the SEM~$|$~CONCEPT feature, which gives a
brief English description of the object, event, etc. that the root word
represents. 
The description is mostly human-oriented and not directly usable by
NLP subsystems, such as transfer units (from Turkish to English and
vice versa) in machine translation systems. 
For example, this feature may take the value {\em throw a physical
  object} for the verb {\em at-}. 
Using an ontological component in the lexicon eliminates this problem,
in which concepts would be described via a semantic network.

Another problem that the ontological component would eliminate is the
following: the subcategorization information for verbs, common nouns,
etc. may places some semantic constraints on the complements, such as
the agent of the verb {\em ye-} ({\em eat something}) must be animate
(SEM$~|$~ANIMATE is +).  This constraint would be tested with the
semantic feature in the feature structure of the subject during
syntactic analysis. This test, however, may fail due to the absence of
the feature SEM$~|$~ANIMATE, but this structure may describe a human,
such as {\em \"{o}\v{g}renci} {\em student} having SEM~$|$~HUMAN:+, so
satisfying animateness constraint. This syntactic mismatch of the
features would be eliminated easily, since a human object would
inherit animateness property (see Y{\i}lmaz~\cite{Yilmaz-Thesis} for
such a component in a verb lexicon).

One of the problems with producing feature structures, especially with
the derivations involved, is predicting semantic properties of common
nouns and qualitative adjectives. In the other categories either
semantic properties are not introduced or they do not receive
derivation. 

Since the new word generated as a result of the derivation process
does not have a lexicon entry, the process should predict some feature
values. However, the semantics of the object or the quality 
that the derivation process produces is not clear. For example,
consider the derivation that takes a common noun and the suffix {\em
  c{\i}}, and produces a common noun. Both {\em ak\c{s}amc{\i}} and
{\em \"o\v{g}lenci} are produced in this way, however, the semantic
properties of the resultant entities are not predictable. This is the
case in {\em yaz{\i}c{\i}} ({\em yaz-} ({\em write})+{\em c{\i}}),
which has two senses: {\em printer} and {\em the person who writes}. The
two senses have different properties, e.g., animateness.

A similar situation occurs for the qualitative adjectives. For
instance, as we stated previously, the gradability of derived forms
are not quite predictable: {\em \c{c}ok ak{\i}ls{\i}z} vs. *{\em
  \c{c}ok kolsuz}.


\chapter{Implementation}
\label{chapter:implementation}

The processing in the lexicon consists of four main steps each
carried out by a separate module: 

\begin{enumerate}
\item morphological analysis,
\item transformation of morphological processor output to static
  lexicon the syntax (i.e., feature structure syntax), and category
  mapping, 
\item retrieval from feature structure databases and producing
  feature structures,
\item application of restrictions.
\end{enumerate}

Except the morphological processor component,\footnotemark\, which is
previously implemented, all the components are implemented in {\em
  SICStus Prolog release 3 \#5}~\cite{SICStus}. Since we described the
procedural aspects of the lexicon in
Chapter~\ref{chapter:operational-aspects}, we will not go into the
details of this process, however, there is one point to be made here: in
the implementation, the query form can contain features only from CAT
and MORPH, since the lexicon interface does not gain much by adding
the capability of restricting SYN and SEM features, as well. On the
other hand, NLP subsystems using this interface can impose any
restriction externally, because access to all features is allowed.
So, rather than applying restrictions to eliminate unwanted feature
structures as the final step, the system applies restrictions to
{\em parses} right after the transformation phase (i.e., when the CAT
and MORPH features are computed). Thus, unnecessary retrievals and 
computations are avoided.

\footnotetext{
  The morphological processor that our lexicon employs is
  implemented by Oflazer (see Oflazer~\cite{Oflazer} for the two level
  description of Turkish morphology) using a finite-state lexicon
  compiler by Karttunen~\cite{Karttunen}.
}

We provided a procedural interface for the lexicon, rather than
implementing a graphical one, since the interface will be
open to NLP subsystems in practical applications.

In this chapter, we will first describe an important component
of the system, the feature structure database (i.e., the root word
lexicon). Then, we will give outputs from sample runs of the
system.

\section{Feature Structure Database}
\label{sec-5:fsdb}

The feature structure database consists of a list of feature structures
indexed with category and root word. Each word and sense is a separate
entry in the database, so given a category and root word more than one
entry may match, that is the key is not unique. Each entry is a unit
Prolog clause with seven arguments, the first five ones giving the
category, and the other two giving the root word and the corresponding
feature structure (see Figure~\ref{figure5:entry}). In this way, the
database can be stored in the main memory and allows fast access.

\begin{figure}[ht]
  \begin{center}
    \begin{tt}
      fsdb(verb, existential, none, none, none, var,\\
      \quad[cat:[maj:verb, \ldots], syn:[\ldots], \ldots]).
    \end{tt}
  \end{center}
  \caption{The entry for the existential verb {\em var} in the feature
    structure database.}
  \label{figure5:entry}
\end{figure}

Feature structures are represented as a list of
$<${\em feature name}:{\em feature value}$>$ pairs (see Gazdar and
Mellish~\cite{Gazdar-Mellish}). For example, the following feature
structure with abstract representation would be represented in Prolog
as in Figure~\ref{figure5:fs_prolog}:

\avmBegin
\[{}
  MORPH &
        \[{}
          STEM &
               \[{}
                 CAT &
                     \[{}
                       MAJ & nominal\\
                     \]\\
               \]\\
          CASE & dat\\
        \]\\
  SEM &
      \[{}
        ANIMATE & $-$\\
        COUNTABLE & $-$\\
      \]
\]
\avmEnd

\begin{figure}[htp]
  \begin{center}
    \begin{tt}
      [morph:[stem:[cat:[maj:nominal |\_] |\_], case:dat |\_],\\
      sem:[animate:-, countable:- |\_] |\_]
    \end{tt}
  \end{center}  
  \caption{Prolog representation of a feature structure.}
  \label{figure5:fs_prolog}
\end{figure}
 
Currently, our feature structure database contains about 50 entries,
which consists of samples from the closed-class words, such as
post-positions, conjunctions, and from other categories showing some
special property. 
More entries will be added to the system later.
In order to maintain the database, the system provides a number of
predicates to add, delete, and browse entries.

\section{Sample Runs}
\label{sec-5:sample-runs}

In this section we will present three sample runs that will
demonstrate features of our lexicon, and will clarify the algorithms
presented in Chapter~\ref{chapter:operational-aspects}.

The input to the system is a query form in the form of a feature
structure. At least the PHON feature, which holds the surface form,
must be present in the query form. Other features are optional, and if
present they act as restrictions on the final output feature
structures. The user can test presence of a feature or a specific
value for that feature. If the feature restricted is in the output
feature structure, the restriction value, which may be unspecified to
test the presence, is unified with the one in the output structure. If
the unification fails, the output structure is eliminated. If such a
feature is not in the output structure, the restriction feature would
not be appropriate for this structure, so it is again eliminated; for
example MORPH~$|$~TAM1 feature is not appropriate for a conjunction's
feature structure.

As previously mentioned, the process is divided into four phases in
the implementation. All four phases inform the user about the state of
the processing. The final output is a list of feature structures which
satisfy all the constraints.


\subsection{Example 1}
\label{sec-5:example-1}

The first example submits only the surface form {\em at{\i}m} and
does not constrain any other features. According to the morphological
processor, {\em at{\i}m} has three parses, as illustrated by the
following examples:

\eenumsentence{
  \label{exmaple-5:atim}
\item
  \shortex{4}
  {Benim & bir & at{\i}m & var.}
  {{\tt my} & {\tt a} & {\tt horse+P1SG} & {\tt existent}}
  {`I have a horse.'}
\item
  \shortex{5}
  {K\"{u}heylan & ben & bir & at{\i}m & dedi.}
  {{\tt K\"{u}heylan} & {\tt I} & {\tt a} & {\tt horse+PRES+1SG} 
    & {\tt say+PAST+3SG}}
  {`K\"{u}heylan said that it was a horse.'}
\item
  \shortex{4}
  {Tilki & bir & at{\i}m & mesafedeydi.}
  {{\tt fox} & {\tt one} & {\tt shot} & {\tt distance+PAST+3SG}}
  {`The fox was in one shot distance.'}
  }

The category of the surface form {\em at{\i}m} is common noun and
attributive verb, respectively, in the first two parses, and they are
due to the common noun {\em at} ({\em horse}). The third parse comes
from the common noun {\em at{\i}m} ({\em shot}), and does not derive
to another category. Since query form does not place any constraint,
the system will generate output for all of the parses, as far as the
feature structure database contains corresponding entries. 

The user input and the lexicon's output follow:

Input query form:\footnote{ In our system, Turkish words consist of all
  lowercase letters, and {\em {\i}}, {\em \c{c}}, {\em \v{g}}, {\em
    \c{s}}, {\em \"{o}}, and {\em \"{u}} are represented as the
  capital of the nearest letter.  }
\begin{verbatim}
[phon:atIm]
\end{verbatim}

Output:
\begin{footnotesize}
\begin{verbatim}
Parsing surface form started...

     Reading Turkish binary file...
     0%>>>>>>>>>>>>>>>>>>>>>>>>>>>>>>>>>100%
     Read Turkish binary file.

     Parsing: atIm
     Number of parses: 3
     1: [[CAT=NOUN][ROOT=at][AGR=3SG][POSS=1SG][CASE=NOM]]
     2: [[CAT=NOUN][ROOT=at][AGR=3SG][POSS=NONE][CASE=NOM]
         [CONV=VERB=NONE][TAM2=PRES][AGR=1SG]]
     3: [[CAT=NOUN][ROOT=atIm][AGR=3SG][POSS=NONE][CASE=NOM]]

Parsing surface form ended...

Transformation phase started...

     Category mapping from:
          noun, none and at
     to:
          nominal, noun, common, none, none

     Category mapping from:
          noun, none and at
     to:
          nominal, noun, common, none, none

     Category mapping from:
          verb, none and none
     to:
          verb, attributive, none, none, none

     Exception: Entry not found in LCMT: Skipping parse...
          noun
          none
          atIm

Transformation phase ended...

     Transformed parses:
     -------------------
     Parse information:
          Number of parses: 2
          1: 1 level(s)
          2: 2 level(s)

Application of restrictions phase started...

Application of restrictions phase ended...

     Satisfying parses:
     ------------------
     Parse information:
          Number of parses: 2
          1: 1 level(s)
          2: 2 level(s)

Retrieval phase started...

     Access to FSDB with:
          nominal, noun, common, none, none and at
     for:
          1 entry/entries

     Access to FSDB with:
          nominal, noun, common, none, none and at
     for:
          1 entry/entries

     Access to TFSDB with:
          verb, attributive, none, none, none

Retrieval phase ended...

     Final result:
     -------------
     Number of feature structures: 2
     Feature sturucture(s):

        [sem:
              [countable: +
               animate: +
               concept: at-(horse)
               material: -
               unit: -
               container: -
               spatial: -
               temporal: -]
         cat:
              [maj: nominal
               min: noun
               sub: common
               ssub: none
               sssub: none]
         morph:
              [stem: at
               form: lexical
               case: nom
               poss: 1sg
               agr: 3sg]
         syn:
              [subcat: none]
         phon: atIm]
         ,
         [cat:
              [maj: verb
               min: attributive
               sub: none
               ssub: none
               sssub: none]
         morph:
              [stem:
                   [sem:
                        [countable: +
                         animate: +
                         concept: at-(horse)
                         material: -
                         unit: -
                         container: -
                         spatial: -
                         temporal: -]
                    cat:
                        [maj: nominal
                         min: noun
                         sub: common
                         ssub: none
                         sssub: none]
                    morph:
                        [stem: at
                         form: lexical
                         case: nom
                         poss: none
                         agr: 3sg]
                    syn:
                        [subcat: none]
                    phon: none]
               form: derived
               derv_suffix: none
               tam2: pres
               copula: none
               agr: 1sg]
          syn:
              [subcat: none]
          sem:
              [concept: none(at-(horse))
               roles: none]
          phon: atIm]

\end{verbatim}
\end{footnotesize}

The output is a trace of the four phases. The first part is the
morphological parsing, and displays parses. The second part is
the transformation of parses into static lexicon syntax (i.e., feature
structure syntax), and category mapping. The first item in the output
of this phase shows the mapping of the morphological processor
category {\em noun} to the lexicon category {\em common noun} for the
root word {\em at}. The next two output items illustrate category
mapping of the second parse. The last item shows that the category
mapping table for root words does not have an entry for {\em at{\i}m},
that is the system does not have information about {\em at{\i}m}, so
this parse is omitted, and will not be processed in the following
phases.

After the transformation phase, two parses remain, and since no
restriction is imposed by the user, these parses will pass to the next
phase. The retrieval part acknowledges the user that it accessed the
feature structure database entry of the common noun {\em at} two
times, and the template feature structure for attributive verbs, which
is due to the derivation in the second parse.

Each parse produces only one feature structure, because the common
noun {\em at} has only one entry/sense in the database. The final
output is these feature structures. The processing including
interfacing with the morphological processor, producing feature
structures, and pretty-printing takes approximately 30 msec. of
running time for compiled Prolog code, so it is rather fast. 
As we mentioned in Chapter~\ref{chapter:lexicon}, the number of
lexical items in a lexicon of a system with acceptable coverage (e.g.,
The Core Language Engine) will not exceed a few thousand, so whole
database can be stored in the main memory. Thus, as the size of our
lexical database gets larger, the processing time will not exceed
acceptable limits.


\subsection{Example 2}
\label{sec-5:example-2}

This example run submits the surface form {\em memnunum} to the system
and constraints the output to be of category {\em verb}. Given this
surface form, morphological processor gives three parses as used in
the following examples:\footnotemark

\eenumsentence{
  \label{example-5:memnunum}
\item
  \shortex{2}
  {Senden & memnunum.}
  {{\tt you+GEN} & {\tt happy+PRES+1SG}}
  {`I am happy with you.'}
\item
  \shortexnt{2}
  {Memnunum & benim!}
  {{\tt happy one+P1SG} & {\tt my}}
\item
  \shortex{2}
  {Ben & Memnun'um.}
  {{\tt I} & {\tt Memnun+PRES+1SG}}
  {`I am Memnun.'}
}

\footnotetext{
  The usage in the second sentence is like in {\em g\"{u}zelim benim},
  that is the qualitative adjective {\em g\"{u}zel} ({\em beautiful})
  is subject to a derivation to common noun, and becomes {\em the one
    that is beautiful}. This usage of {\em Memnun} is syntacticly
  correct, though semantically it does not make sense.
}

The first two parses are due to the qualitative adjective {\em memnun}
({\em satisfied}/{\em happy}), and contain derivations to attributive
verb and common noun, respectively. The last one is due to the proper
noun {\em Memnun} and contains a derivation to attributive verb. The
only restriction in the query form is that the output feature
structures must be of type {\em verb}, which will cause the second
parse to be eliminated in the third phase.

The input and corresponding output follow:

Input query form:
\begin{verbatim}
[phon:memnunum, cat:[maj:verb]]
\end{verbatim}

Output:
\begin{footnotesize}
\begin{verbatim}
Parsing surface form started...

     Parsing: memnunum
     Number of parses: 3
     1: [[CAT=ADJ][ROOT=memnun][CONV=VERB=NONE][TAM2=PRES][AGR=1SG]]
     2: [[CAT=ADJ][ROOT=memnun][CONV=NOUN=NONE][AGR=3SG][POSS=1SG][CASE=NOM]]
     3: [[CAT=NOUN][ROOT=memnun][TYPE=RPROPER][AGR=3SG][POSS=NONE][CASE=NOM]
         [CONV=VERB=NONE][TAM2=PRES][AGR=1SG]]

Parsing surface form ended...

Transformation phase started...

     Category mapping from:
          adj, none and memnun
     to:
          adjectival, adjective, qualitative, none, none

     Category mapping from:
          verb, none and none
     to:
          verb, attributive, none, none, none

     Category mapping from:
          adj, none and memnun
     to:
          adjectival, adjective, qualitative, none, none

     Category mapping from:
          noun, none and none
     to:
          nominal, noun, common, none, none

     Exception: Entry not found in LCMT: Skipping parse...
          noun
          rproper
          memnun

Transformation phase ended...

     Transformed parses:
     -------------------
     Parse information:
          Number of parses: 2
          1: 2 level(s)
          2: 2 level(s)

Application of restrictions phase started...

     Parse eliminated: Printing only the last level...

         [cat:
              [maj: nominal
               min: noun
               sub: common
               ssub: none
               sssub: none]
          morph:
              [derv_suffix: none
               agr: 3sg
               poss: 1sg
               case: nom]
          phon: memnunum]

Application of restrictions phase ended...

     Satisfying parses:
     ------------------
     Parse information:
          Number of parses: 1
          1: 2 level(s)

Retrieval phase started...

     Access to FSDB with:
          adjectival, adjective, qualitative, none, none and memnun
     for:
          1 entry/entries

     Access to TFSDB with:
          verb, attributive, none, none, none

Retrieval phase ended...

     Final result:
     -------------
     Number of feature structures: 1
     Feature sturucture(s):

         [cat:
              [maj: verb
               min: attributive
               sub: none
               ssub: none
               sssub: none]
          morph:
              [stem:
                   [syn:
                        [subcat: ...
                         modifies: ...]
                    cat:
                        [maj: adjectival
                         min: adjective
                         sub: qualitative
                         ssub: none
                         sssub: none]
                    morph:
                        [stem: memnun
                         form: lexical]
                    sem:
                        [concept: memnun-(satisfied)
                         gradable: -
                         questional: -]
                    phon: none]
               form: derived
               derv_suffix: none
               tam2: pres
               copula: none
               agr: 1sg]
          syn:
              [subcat: ...]
          sem:
              [concept: none(memnun-(satisfied))
               roles: none]
          phon: memnunum]

\end{verbatim}
\end{footnotesize}

In the transformation of parses, no entry regarding the proper noun
{\em Memnun} is found in the category mapping table, so this parse is
eliminated, leaving two parses to the third phase, which discards the
second parse, since it fails to satisfy the restriction, that is the
value of CAT~$|$~MAJ must be {\em verb}. Finally, there is only one
parse left, which is the first one, as an input to the retrieval
phase. As seen in the output, there is only one entry for the
qualitative adjective {\em memnun}, thus only one feature structure is
generated. The processing takes approximately 50 msec. of running
time. The values of SUBCAT and MODIFIES features are omitted to save
space (see the full feature structure of {\em memnun} on
page~\pageref{avm-3:memnun}).


\subsection{Example 3}
\label{sec-5:example-3}

Our last example will demonstrate multiple senses in the database. The
surface form is {\em ekim}, and the restriction is on MORPH~$|$~POSS
feature, whose value must be {\em 1sg}. The interpretations are
similar to those in the previous examples, so we will not give
detailed descriptions.

According to the morphological processor, there are three parses,
which are due to the common noun {\em ek} ({\em appendix}/{\em
  suffix}) and {\em Ekim} ({\em October}). Both root words are in the
database, but the last two parses are eliminated in the third phase.
As a result, there is only one parse as an input to the last step.
There are two entries regarding the common noun {\em ek}, which cause
the system to generate two feature structures for the single parse.
The processing takes about 40 msec.

The input and corresponding output follow:

Input query form:
\begin{verbatim}
[phon:ekim, morph:[poss:'1sg']].
\end{verbatim}

Output:
\begin{footnotesize}
\begin{verbatim}
Parsing surface form started...

     Parsing: ekim
     Number of parses: 3
     1: [[CAT=NOUN][ROOT=eK][AGR=3SG][POSS=1SG][CASE=NOM]]
     2: [[CAT=NOUN][ROOT=eK][AGR=3SG][POSS=NONE][CASE=NOM]
         [CONV=VERB=NONE][TAM2=PRES][AGR=1SG]]
     3: [[CAT=NOUN][ROOT=ekim][TYPE=TEMP1][AGR=3SG][POSS=NONE][CASE=NOM]]

Parsing surface form ended...

Transformation phase started...

     Category mapping from:
          noun, none and ek
     to:
          nominal, noun, common, none, none

     Category mapping from:
          noun, none and ek
     to:
          nominal, noun, common, none, none

     Category mapping from:
          verb, none and none
     to:
          verb, attributive, none, none, none

     Category mapping from:
          noun, temp1 and ekim
     to:
          nominal, noun, common, none, none

Transformation phase ended...

     Transformed parses:
     -------------------
     Parse information:
          Number of parses: 3
          1: 1 level(s)
          2: 2 level(s)
          3: 1 level(s)

Application of restrictions phase started...

     Parse eliminated: Printing only the last level...

         [cat:
              [maj: verb
               min: attributive
               sub: none
               ssub: none
               sssub: none]
          morph:
              [suffix: none
               tam2: pres
               agr: 1sg]
          phon: ekim]

     Parse eliminated: Printing only the last level...

         [cat:
              [maj: nominal
               min: noun
               sub: common
               ssub: none
               sssub: none]
          morph:
              [stem: ekim
               agr: 3sg
               poss: none
               case: nom]
          phon: ekim]

Application of restrictions phase ended...

     Satisfying parses:
     ------------------
     Parse information:
          Number of parses: 1
          1: 1 level(s)

Retrieval phase started...

     Access to FSDB with:
          nominal, noun, common, none, none and ek
     for:
          2 entry/entries

Retrieval phase ended...

     Final result:
     -------------
     Number of feature structures: 2
     Feature sturucture(s):

         [sem:
              [countable: +
               concept: ek-(suffix)
               material: -
               unit: -
               container: -
               spatial: -
               temporal: -
               animate: -]
          cat:
              [maj: nominal
               min: noun
               sub: common
               ssub: none
               sssub: none]
          morph:
              [stem: ek
               form: lexical
               case: nom
               poss: 1sg
               agr: 3sg]
          syn:
              [subcat: none]
          phon: ekim]
         ,
         [sem:
              [countable: +
               concept: ek-(appendix)
               material: -
               unit: -
               container: -
               spatial: -
               temporal: -
               animate: -]
          cat:
              [maj: nominal
               min: noun
               sub: common
               ssub: none
               sssub: none]
          morph:
              [stem: ek
               form: lexical
               case: nom
               poss: 1sg
               agr: 3sg]
          syn:
              [subcat: none]
          phon: ekim]
\end{verbatim}
\end {footnotesize}


\chapter{Conclusions and Suggestions}
\label{chapter:conclusion}

In this thesis, we present a lexicon for Turkish. Our work includes
determination of the lexical specification to be encoded for all
lexical types of Turkish, encoding of this specification, and
constructing a standalone system as an information repository for the
NLP systems. 

The level of lexical specification for morphosyntactic and syntactic
information is adequate, but, as the semantic information is added in
an ad hoc manner, it may not satisfy all the requirements of NLP
systems on semantic information. Including a knowledge-base/ontology
into the system, in which concepts are described through a semantic
network, would be useful. This would solve the problem related with
the satisfying the semantic constraints in the subcategorization
information of lexical entries. For example, the constraint posing
SEM~$|$~ANIMATE:+ will not be unified with SEM~$|$~HUMAN:+, though
this is semantically satisfiable.

In order for our lexical database to be computationally useful, more
entries would be added depending on the requirements of the NLP
systems interfacing with our lexicon. Currently, the database contains
about 50 entries consisting of samples from closed and open-class
words having some special property. We are planning to add more
entries to cover all the closed-class words and enrich the content for
the open-class words of Turkish.  A graphical user interface will be
provided to help insertion, deletetion, and update operations to
lexicon.


\bibliography{main}
\bibliographystyle{abbrv}
           
\begin{appendix}
  \chapter{The Lexicon Categories}
\label{appendix-a}

\begin{table}[htb]
\fnsCBegin
\begin{tabular}{|l|l|l|l|l|}\hline
{\em maj} & {\em min} & {\em sub} & {\em ssub} & {\em sssub}\\ \hline \hline
nominal & noun     & common         &            & \\ \hline
        &          & proper         &            & \\ \hline
        & pronoun  & personal       &            & \\ \hline
        &          & demonstrative  &            & \\ \hline
        &          & reflexive      &            & \\ \hline
        &          & indefinite     &            & \\ \hline
        &          & quantification &            & \\ \hline
        &          & question       &            & \\ \hline
        & sentential & act          & infinitive & ma  \\ \hline
        &          &                &            & mak \\ \hline
        &          &                &            & y{\i}\c{s} \\ \hline
        &          & fact           & participle & d{\i}k \\ \hline
        &          &                &            & yacak  \\ \hline 
                                                             \hline
adjectival & determiner  & article       & & \\ \hline
         &             & demonstrative & & \\ \hline
         &             & quantifier    & & \\ \hline         
         & adjective   & quantitative  & cardinal     & \\ \hline
         &             &               & ordinal      & \\ \hline
         &             &               & fraction     & \\ \hline
         &             &               & distributive & \\ \hline
         &             & qualitative   &         & \\ \hline 
\end{tabular}
\fnsCEnd
\caption{The lexicon categories (nominals and adjectivals)}
\label{table-3:lexicon-categories-1}
\end{table}

\begin{table}[htb]
\fnsCBegin
\begin{tabular}{|l|l|l|l|l|}\hline
{\em maj} & {\em min} & {\em sub} & {\em ssub} & {\em sssub}\\ \hline \hline
adverbial & direction    &               &            & \\ \hline
          & temporal     & point-of-time &            & \\ \hline
          &              & time-period   & fuzzy      & \\ \hline
          &              &               & day-time   & \\ \hline
          &              &               & season     & \\ \hline
          & manner       & qualitative   &            & \\ \hline
          &              & repetition    &            & \\ \hline
          & quantitative & approximation &            & \\ \hline
          &              & comparative   &            & \\ \hline
          &              & superlative   &            & \\ \hline
          &              & excessiveness &            & \\ \hline
                                                           \hline
verb      & predicative  &  & & \\ \hline
          & existential  &  & & \\ \hline
          & attributive  &  & & \\ \hline \hline
conjunction   & coordinating    & & & \\ \hline
              & bracketing      & & & \\ \hline
              & sentential      & & & \\ \hline \hline
post-position & nom-subcat    & & & \\ \hline
              & acc-subcat    & & & \\ \hline
              & dat-subcat    & & & \\ \hline
              & abl-subcat    & & & \\ \hline
              & gen-subcat    & & & \\ \hline
              & ins-subcat    & & & \\ \hline
\end{tabular}
\fnsCEnd
\caption{The lexicon categories (adverbials, verbs, conjunctions, 
  and post-positions)}
\label{table-3:lexicon-categories-2}
\end{table}


\end{appendix}

\end{document}